%% file: main.tex
\documentclass{article}
\usepackage[a4paper, margin=2cm]{geometry}
\usepackage[
backend=bibtex,
style=numeric,
sorting=none
]{biblatex}
\newcommand{\citep}[1]{\cite{#1}}

\usepackage{url,hyperref,lineno,microtype,subcaption,rotating}
\usepackage{wrapfig}
\usepackage[onehalfspacing]{setspace}
\usepackage{lscape}

\usepackage{tikz}
\usetikzlibrary{shapes, positioning}
\usetikzlibrary{calc}
\usepackage{pgfplots}
\pgfplotsset{compat=1.5}
\input{tikzvodoo.tex}
\usepackage{xfrac}
\usepackage{amsmath}
\usepackage{amsthm}
\usepackage{amsfonts}
\usepackage{amssymb}
\usepackage{hyperref}
\usepackage{hhline,array}
\usepackage{nicefrac,xfrac}
\usepackage{longtable}

\usepackage{algorithm}
\usepackage{algpseudocode}

\usepackage{multirow}
\usepackage{pifont}

\newcommand{\N}{\mathbb{N}}
\newcommand{\R}{\mathbb{R}}
\renewcommand{\P}{\mathbb{P}}
\newcommand{\E}{\mathbb{E}}
\newcommand{\X}{\mathcal{X}}
\newcommand{\Y}{\mathcal{Y}}
\newcommand{\T}{\mathcal{T}}
\newcommand{\D}{\mathcal{D}}
\renewcommand{\d}{\textnormal{d}}
\newcommand{\indep}{\perp\!\!\!\perp}
\newcommand{\1}{\mathbf{1}}

\newcommand{\Figname}{Fig.}

\newtheorem{theorem}{Theorem}

\theoremstyle{definition}
\newtheorem{definition}{Definition}

\theoremstyle{remark}

\newcommand{\orcidID}[1]{\href{https://orcid.org/#1}{${}^\text{\includegraphics[width=8pt]{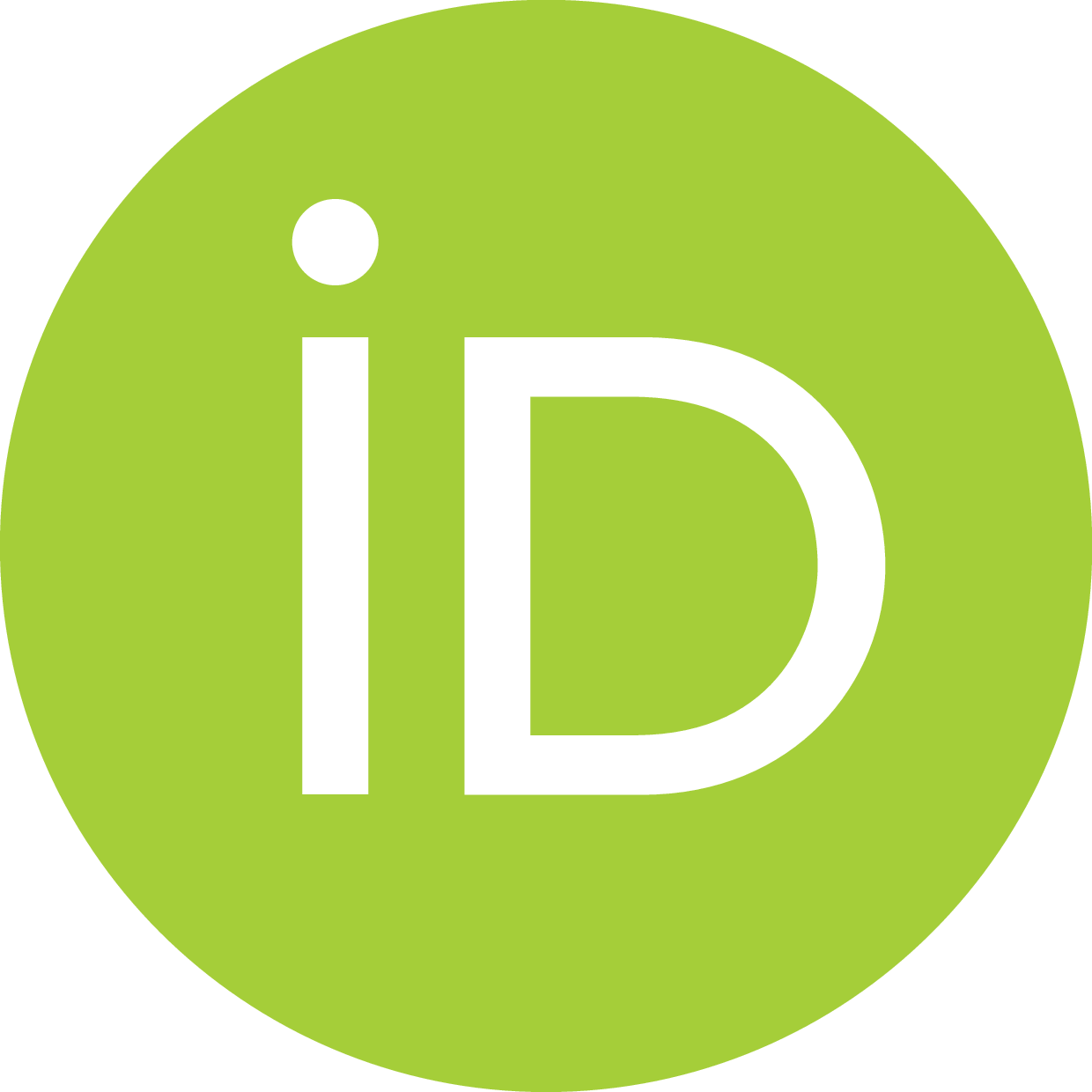}}$}}

\begin{document}
\onecolumn

\title {One or Two Things We know about Concept Drift \\ A Survey on Monitoring Evolving Environments\thanks{Funding in the frame of the ERC Synergy Grant ``Water-Futures'' No. 951424 is gratefully acknowledged.}}

\author{{Fabian Hinder\orcidID{0000-0002-1199-4085}\footnote{Authors contributed equally}\;\footnote{Corresponding Author} \qquad Valerie Vaquet\orcidID{0000-0001-7659-857X}${}^{\dagger}$ \qquad Barbara Hammer\orcidID{0000-0002-0935-5591}} 
\\\;\\ 
\small Bielefeld University - Cognitive Interaction Technology (CITEC)  \\ 
\small Inspiration 1, 33619 Bielefeld - Germany}


\maketitle

\begin{abstract}
The world surrounding us is subject to constant change. These changes, frequently described as concept drift, influence many industrial and technical processes. As they can lead to malfunctions and other anomalous behavior, which may be safety-critical in many scenarios, detecting and analyzing concept drift is crucial.
In this paper, we provide a literature review focusing on concept drift in unsupervised data streams. While many surveys focus on supervised data streams, so far, there is no work reviewing the unsupervised setting. However, this setting is of particular relevance for monitoring and anomaly detection which are directly applicable to many tasks and challenges in engineering.

This survey provides a taxonomy of existing work on drift detection. Besides, it covers the current state of research on drift localization in a systematic way.  In addition to providing a systematic literature review, this work provides precise mathematical definitions of the considered problems and contains standardized experiments on parametric artificial datasets allowing for a direct comparison of different strategies for detection and localization. Thereby, the suitability of different schemes can be analyzed systematically and guidelines for their usage in real-world scenarios can be provided. Finally, there is a section on the emerging topic of explaining concept drift.

\end{abstract}

\section{Introduction}
The environment surrounding us is perpetually changing. While humans are trained to navigate an ever-changing environment, these changes pose challenges to many automated systems~\citep{ditzler_learning_2015}. Considering monitoring and control tasks in critical infrastructure~\citep{vrachimis_battle_2022}, manufacturing~\citep{chen_online_2017}, and quality control~\citep{gabbar_incremental_2023}, in order to work reliably, automatized processes and supervision algorithms need to be able to detect, react, and adapt to changes~\citep{reppa_sensor_2016}. 

Formally, these changes can be described as \emph{concept drift} (or drift for short) -- a change in the data generating distribution~\citep{gama2014survey}. They can be caused by changes in the observed process, the environment, or the sensors acquiring the data. While when monitoring a system, for instance in manufacturing or quality control, it is crucial to detect changes in the observed process as this might indicate faulty productions or general malfunctions, in automated processes, it is important to detect changes in the sensors and the environment to take appropriate actions, e.g. replacing a faulty sensor or adapting the system processing the collected data to a changed scenario~\citep{gonccalves2014comparative,gama2004learning,gama2014survey}. 

Most often, we consider drift in \emph{stream setups}~\citep{ditzler_learning_2015,lu_learning_2018}, where the underlying data distribution changes, requiring models to adapt or to inform a human operator to take appropriate action. This is closely related to concept evolution in \emph{continual learning}~\citep{delange_continual_2021}, commonly discussed in deep learning, where concepts might appear or vanish. Besides, data streams might suffer from temporarily extreme class imbalances or the availability of features might change over time. This can trigger the problem of so-called catastrophic forgetting where a model cannot properly process samples of a class after updating anymore. 
Drift is not limited to data streams but can also occur in \emph{time-series}~\citep{aminikhanghahi2017survey} data where the single observations are highly interdependent~\citep{esling_time-series_2012}. Here, drift mainly occurs in the form of trends. Commonly its absence is referred to as stationarity.

Besides the settings where the data samples are collected over time, considering manufacturing and quality control, frequently data is collected at different locations and processed in the scheme of \emph{federated learning}~\citep{zhang_survey_2021}. Here, instead of gathering all the data at a global server processing is done locally and the results are then combined to get an overarching model at the server. Similar to stream learning, in this setting it is necessary to account for differences or drift in the data collected at different locations to obtain a robust global model~\citep{liu_secure_2020}. Finally, drift must be considered when performing \emph{transfer learning}, a strategy in deep learning~\citep{pan_survey_2010}. The basic idea is to deal with limited data by pre-training a model on a similar task on a more extensive similar dataset and later on fine-tuning to the task and limited dataset at hand. 
In this work, we will focus on data streams only. However, many of the discussed strategies can be directly applied to the previously named tasks.

Considering processing drifting data streams there are two main groups of tasks. While one might be interested in keeping a valid model performing some predictive task on the data, e.g. classifying a product into different categories or estimating a property of interest (\emph{online or stream learning}), another goal is to \emph{monitor a system for anomalous behavior} in order to react appropriately. In this work, we will not consider online learning, as there are many surveys providing a good overview of this task~\citep{ditzler_learning_2015,lu_learning_2018,DBLP:journals/ijon/LosingHW18} and toolboxes~\citep{skmultiflow,montiel2021river}. Instead, we will focus on the monitoring scenario, which is very important in many different settings where drift is expected due to the use of sensor devices or sensitivity to changes in the environment. 
We focus on unsupervised drift detection as this strategy is the most relevant for monitoring settings. Besides, it is highly relevant as a prerequisite for understanding drift phenomena.
Notice that this setup is also different from drift detection for online learning with limited or no label information which is also sometimes referred to as unsupervised drift detection \citep{gemaque2020overview}.
The difference to supervised drift detection will be discussed in detail in Section~\ref{sec:sup-vs-unsup}. 
Furthermore, we also provide a formal mathematical definition of the main problems, concepts, and notions and a survey in how far these are addressed by current technologies. Moreover, we also have a look at the in depth analysis techniques like drift localization in data space and the problem of drift explanation.

The task of monitoring is to observe a system and to provide all the information necessary to enable a human operator or automatized downstream tasks to take actions that ensure that the system runs properly. Which information is required depends on the specific task~\citep{DBLP:journals/kais/GoldenbergW19,laha_machine_2021}. However, generally, it can be summarized by addressing the following questions about the drift~\citep{lu_learning_2018}:

The first question in every setting concerns \emph{whether (and when)} drift occurs. The task of determining whether or not there is drift during a given time period is called \emph{drift detection}~\citep{gama2014survey}. We will consider drift detection in Section~\ref{sec:drift-detection} in detail. In case a drift is detected, additional questions need to be raised to appropriately react to the change in the data distribution.

A second question of interest might concern the severity of the drift, as this might influence which kind of action needs to be taken. Usually \emph{drift quantification}~\citep{lu_learning_2018} can be realized as a precursor of drift detection: As described later, many methods for drift detection estimate the rate of change by some kind of metric and trigger an alarm if those changes exceed a threshold. Thus, we will not focus on this question in great detail.

In order to take appropriate action, it is important to pinpoint the drift more precisely. While drift detection and quantification deal with the when by assigning drift-related information to the time component, i.e., finding change points, determining
the rate of change, \emph{drift localization and segmentation}~\citep{lu_learning_2018} focus on the \emph{where} and assign drift-related information to the data space. Consider for example quality control. There might already be an algorithm in place screening the data for known anomalies. However, in case new anomalies in the product occur it is required to detect those to analyze whether some action is required, e.g. discarding the item. In this case, it is crucial to identify the anomalous items, i.e. the drifting data samples, for further analysis. We will focus on drift localization in Section~\ref{sec:drift-localization}.

In some settings answering the discussed questions is not sufficient. There are systems in which a malfunction, i.e. the drift, causes a change in a number of features in all data points collected after the drift event. For example, this might be the case if a sensor is degrading and thus yielding changed measurements. In this case, only using drift localization does not provide much information about what actually happened. Instead, we need more detailed information of \emph{what} exactly happened and \emph{how} it can be described. Providing these detailed, complete, and human-understandable descriptions of ongoing drift is referred to as \emph{drift explanation}~\citep{neucomp}. Such methods are designed to support human operators by providing relevant information on monitoring and adaptation processes. This is relevant as the complexity of drift can easily surpass the level of information which is provided by change points or the estimate of the rate of change. Indeed, drift can manifest in a change in the correlation of several features alone, making it nearly impossible for humans to observe without machine aid. 
In a sense, drift explanations can be seen as the explainable AI (XAI) counterpart for drift detection: while usual XAI explains why a model makes a decision~\citep{gunning2019xai,molnar2019interpretable}, drift explanations provide an explanation of why a drift detector alerts for drift. This commonly makes use of various techniques, including classical XAI. We provide an overview of the most advanced drift explanation schemes in Section~\ref{sec:drift-explanation}.

This paper is structured as follows. First, we provide a formalization of concept drift (Section~\ref{sec:drift-defs}), and position this paper both in the body of related work in the intersection of the stream setup and supervised and unsupervised approaches (Section~\ref{sec:sup-vs-unsup}). Afterwards, we focus on drift detection (Section~\ref{sec:drift-detection}) and drift localization (Section~\ref{sec:drift-localization}): We first formalize these tasks and provide a general scheme most approaches realize. We then propose a categorization of the methods and finally analyze the strategies with respect to drift and stream-specific criteria. Before concluding this survey (Section \ref{sec:conclusion}), we discuss drift explanation by highlighting some of the most advanced and interesting contributions (Section~\ref{sec:drift-explanation}).

\section{Concept Drift -- defining the setup\label{sec:drift-defs}}
Before we look at drift detection, drift localization, and drift explanations in detail, in this section, we formally define drift and discuss different setups for working with it.
\subsection{A Formal Model of Concept Drift\label{sec:drift-formal}}

In the classical setup of machine learning, one assumes that the distribution at training, testing, and application time is always the same, i.e., we assume that the data generating distribution $\D$ is time-invariant. In this case, a sample of size $n$ is a collection of i.i.d. random variables $X_1,\dots,X_n \sim \D$.

As discussed before, the assumption of time-invariant distributions is violated in many real-world applications, in particular, when learning on data streams. To resolve this issue from a purely formal point of view, we incorporate time into our considerations by allowing every point to follow a potentially different distribution $X_i \sim \D_{t_i}$ that depends on the time point $t_i$ of observation. As it is unlikely to observe two samples at the same time, i.e., $t_i \neq t_j$ for all $i \neq j$, it is common to simply write $\D_i$ instead of $\D_{t_i}$ \citep{gama2014survey}. 

This relates to the classical setup if all $X_i$ actually follow the same distribution, i.e.,  $\D_i = \D_j$ holds for all $i,j$. One speaks of \emph{concept drift} if this assumption is violated, i.e., there exist $i,j$ such that $\D_i \neq \D_j$ \citep{gama2014survey}. 

However, as pointed out by \citep{dawidd} this definition of concept drift depends on the used sample and not on the underlying process. In particular, it might happen that if we take two samples from the same data source over the same period of time using different sampling frequencies, one sample will have concept drift and the other will not. This makes understanding concept drift a hard problem. In order to deal with the issue, it was suggested in \citep{dawidd} to additionally take the statistical properties of time into account. To do so we consider an 
model of time $\T$ rather than a mere index set. We assume that there is a distribution $P_T$ on $\T$ that describes the likelihood of observing a sample at time $t$, and a collection of distributions $\D_t$ for all $t \in \T$ albeit, in practice, only a finite number of time point is observed. 
Together $P_T$ and $\D_t$ form what is referred to as a \emph{drift process}
\begin{definition}
    Let $\T = [0,1]$ and $\X = \R^d$\footnote{All considerations below also work in a very similar way for arbitrary measure spaces. However, as some formal issues may arise from that, we will stick with this far simpler special case for the sake of clarity.}.
    A \emph{drift process} $(P_T,\D_t)$ from the \emph{time domain} $\T$ to the \emph{data space} $\X$ is a probability measure $P_T$ on $\T$ together with a Markov kernel $\D_t$ from $\T$ to $\X$, i.e., for all $t \in \T$ $\D_t$ is a probability measure on $\X$ and for all measurable $A \subset \X$ the map $t \mapsto \D_t(A)$ is measurable. We will just write $\D_t$ instead of $(P_T,\D_t)$ if this does not lead to confusion.
\end{definition}

We can derive two particularly important types of distributions: By adding a time-stamp to every sample from the moment of its arrival the data follows what we will refer to as the holistic distribution $\D$. By collecting all samples observed during a certain time window $W \subset \T$ the data follows the windowed distribution $\D_W$. Formally the distributions are given by the following:

\begin{definition}
    Let $(\D_t,P_T)$ be a drift process from $\T$ to $\X$.
    We refer to the distribution $\D$ on $\X \times \T$ which is uniquely determined by the property $\D(A \times W) = \int_W \D_t(A) \d P_T(t)$ for all $A \subset \X,\; W \subset \T$ as the \emph{holistic distribution} of $\D_t$\footnote{Both existence and uniqueness of $\D$ are assured by the Fubini-Tonelli theorem.}. Furthermore, we call a $P_T$ non-null set $W \subset \T$ a \emph{time-window} and denote by $\D_W(A) = \int_W \D_t(A) \d P_T(t \mid W) = \D(A \times W \mid \X \times W)$ the \emph{mean distribution} during $W$. 
\end{definition}

A benefit of a drift process is that it allows us to sample data from it. This is in stark contrast to the sample-based setup, as there is no reasonable way to create a new sample from an old one. There are two ways to draw new data from a drift process. One option is to draw i.i.d. samples from the holistic distribution $\D$. These samples are dated-data points $(T,X)$ that are often obtained by the following procedure in practice: First draw the time of observing $X$, i.e., $T \sim P_T$, and then draw $X$ according to $\D_t$ assuming $T = t$, i.e., $X \mid [T = t] \sim \D_t$. Another sampling method that is often used in practice is to take i.i.d. samples from $\D_W$ for some time window $W$. Notice that a collection of observations that are collected during a time window $W$ according to $\D$ are exactly distributed according to $\D_W$. Hence both ways to sample are just formal descriptions of practically relevant procedures to obtain data over time.

We derive a definition for drift based on the definition above. As we are now talking about a property of a data-generating process and not just a sample drawn from it, we must add a slight adaptation corresponding to the property that drift can actually be observed: We say that $\D_t$ has drift if the chance of obtaining a sample for which there is drift in the sense of the definition given above occurs is larger zero, i.e., $X_1,X_2,\dots$ is a sample, then there are $i,j$ such that 
\begin{align*}
    \P_{X_i} \overset{\text{def. $X_i$}}{=} \D_{T_i} \neq \D_{T_j} \overset{\text{def. $X_j$}}{=} \P_{X_j}
\end{align*}
with a chance larger than zero. Due to measure theoretical reasons the number of samples actually does not play a role so we can also consider only two samples. Thus, we obtain the following definition.
\begin{definition}
\label{def:drift}
    Let $(P_T,\D_t)$ be a drift process. We say that $\D_t$ has \emph{drift} iff
    \begin{align*}
    \P_{T,S \sim P_T}[\D_T \neq \D_S] = P_T^2(\{(t,s) \in \T^2 \mid \D_t \neq \D_s \}) > 0.
\end{align*}
\end{definition}

One may be wondering why this is different from the existence of $s,t \in \T$ with $\D_t \neq \D_s$. Formally speaking this has to do with $P_T$ null sets. It might happen that the difference only occurs in such a short amount of time, that we will never see only a single sample drawn from the other distribution and thus we will never be able to observe the drift in the data. It is thus a mere artifact of the formal model, rather than the actual process.

In \citep{dawidd} several other, equivalent formalizations which relate to scenarios which have been considered in the literature of concept drift are given: being not equal to a standard distribution, i.e., $P_T[\D_t \neq P] > 0$ for all distributions $P$ on $\X$; being not equal to the mean distribution, i.e., $P_T[\D_t \neq \D_\T] > 0$; different distributions for two time-windows, i.e., $\D_W \neq \D_{W'}$ for some $W,W' \subset \T$. One of the key findings however, which allows the development of new methods, is that drift can equivalently be formulated as data $X$ and time $T$ are dependent, i.e., not statistically independent:

\begin{theorem}
    Let $(\D_t,P_T)$ be a drift process from $\T$ to $\X$ and let $(T,X) \sim \D$ be distributed according to the holistic distribution. Then $\D_t$ has no drift if and only if $T \indep X$ are statistically independent, i.e., there exist $W \subset \T$ and $A \subset \X$ such that $\P[T \in W, X \in A] \neq \P[T \in W]\P[X \in A]$.
\end{theorem}

\subsection{Concept Drift in Supervised and Unsupervised Setups\label{sec:sup-vs-unsup}}
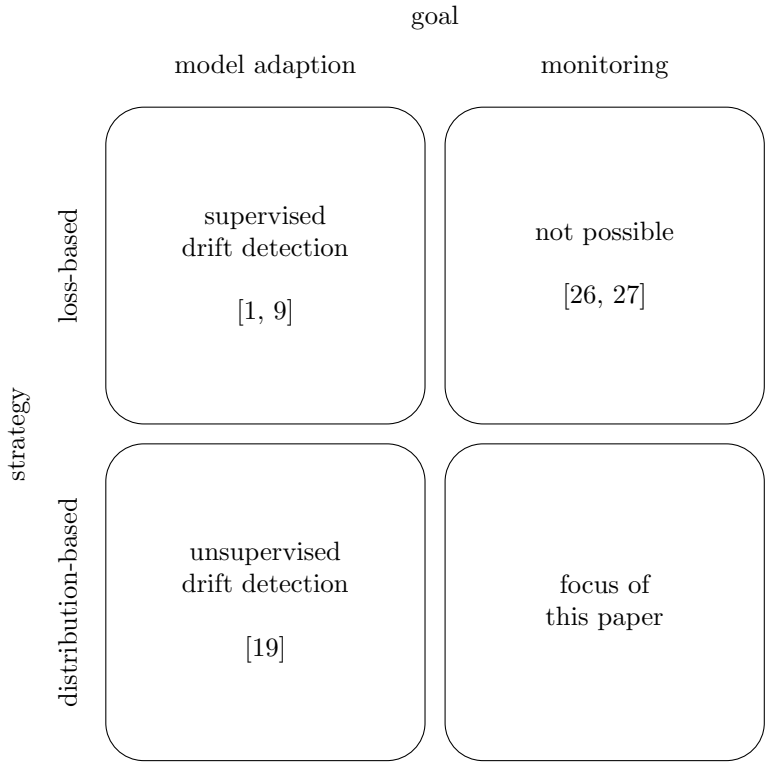
\begin{figure}
    \centering
\tikzset{box/.style={minimum height=4.2cm, minimum width=4.2cm, rounded corners=.5cm,draw}}
\tikzset{label/.style={minimum height=0cm, minimum width=0cm}}
\begin{tikzpicture}[node distance=0.25cm and 0.25cm]
    \node (a11) at (0,0) [box, align=center]        {supervised \\ drift detection \\\,\\ \cite{ditzler_learning_2015,lu_learning_2018}};
    \node (a12) [right = of a11, box, align=center] {not possible \\\,\\ \cite{ida2023,icpram}};
    \node (a21) [below = of a11, box, align=center] {unsupervised \\ drift detection \\\,\\ \cite{gemaque2020overview}};
    \node (a22) [right = of a21, box, align=center] {focus of \\ this paper}; 

    \node (l1) [left = of a11] {\rotatebox[origin=c]{90}{loss-based}};
    \node (l2) [left = of a21] {\rotatebox[origin=c]{90}{distribution-based}};
    \node (l3) [above = of a11] {model adaption};
    \node (l4) [above = of a12] {monitoring};

    \node (l1m) at ($(l1)!0.5!(l2)$) {};
    \node (l5) [left = of l1m] {\rotatebox[origin=c]{90}{strategy}};
    \node (l3m) at ($(l3)!0.5!(l4)$) {};
    \node (l6) [above = of l3m] {goal};
\end{tikzpicture}

    \caption{Display of the drift analysis categorization according to the goal and the applied strategy}
    \label{fig:sup-unsup}
\end{figure}

In the last section, we provided a definition of drift based on the data-generating process. Drift is usually further categorized. A first general distinction is usually drawn based on the way drift manifests itself in time. While the distribution might completely change at time $t$ (\emph{abrupt drift}), there are also slower changes occurring over an interval of time. In \emph{incremental drift}, the distribution changes smoothly over time. In contrast, in \emph{gradual drift}, during the period of change, the samples are drawn from both distributions with different probabilities. Finally, in many real-world applications, we expect old distributions to reoccur, for instance, due to seasonalities. This phenomenon is referred to as \emph{reoccurring drift}. Notice that some authors refer to abrupt drift as concept shift and only call a continuous change as in gradual or incremental drift concept drift. If not further specified we consider all the aforementioned kinds of drift at once. 

Besides, one categorizes drift according to how the distribution changes in the data and label space. 
Assuming that the data stream consists of labeled-data-pairs $(X,Y) \in \X \times \Y$, where $Y$ is the label, in addition to the changes in the joint distribution of $X$ and $Y$, the marginal and conditional distributions are of interest. Usually, a change of the posterior $\D_t(Y\mid X)$ is referred to as \emph{real drift}, and a change of the marginal $\D_t(X)$ is referred to as \emph{virtual drift}. Sometimes virtual drift is also called data drift, while real drift is referred to as concept drift.

As pointed out by \citep{icpram}, from a statistical point of view, drift in the marginal distribution $\D_t(X)$ and the joint distribution $\D_t(X,Y)$ can be modeled in a common mathematical framework, although the interpretations are of course different. In particular, in both cases drift is equivalent to the statistical dependency of time $T$ and (labeled-)data $X$ and $(X,Y)$, respectively, if the time-enriched representation is (holistic distribution) considered.
Real drift $\D_t(Y\mid X)$ on the other hand is equivalent to conditional statistical independence of label $Y$ and time $T$ given data $X$, i.e., $Y \indep T \mid X$.

Analogous to general machine learning tasks, we can consider \emph{supervised} settings, i.e., those that are concerned with conditional distributions, and \emph{unsupervised} tasks, i.e., those that are concerned with the joint or marginal distributions. While in supervised settings both real and virtual drift might be present, in unsupervised settings only virtual drift has to be considered.

As briefly discussed before, there are two main goals when facing drifting data streams. While the goal might be to \emph{keep an accurate learning model} even if the data stream is drifting, it might be of interest to accurately detect and describe the drift in the data distribution (\emph{monitoring}). These two goals intersect with two overarching approaches concerning the discussed settings. While in a supervised setting, one is usually relying on analyzing \emph{model-losses}, i.e., how well a specific model can reconstruct the data or perform a prediction or forecasting task, as a proxy, in unsupervised settings one is immediately considering the \emph{data distribution}. Structuring approaches according to these dimensions, we obtain a categorization shown in \Figname~\ref{fig:sup-unsup}.

Considering this structure considerable work has been conducted on model adaption by using loss-based strategies in the supervised setting. As discussed before, here the main goal is to keep an accurate model. Thus, relying on the model loss (such as the so-called interleaved test-train loss) as an indicator of when to update the model is a reasonable choice. There are many surveys on this particular task~\citep{ditzler_learning_2015,lu_learning_2018,DBLP:journals/ijon/LosingHW18}, so we will not consider it in detail in this work.

In contrast, there are fundamental mathematical obstacles preventing that technologies which are observing the model loss can monitor general drift phenomena other than specific ones which effect the loss.
The connection between model loss, model adaption, and real drift is rather vague and heavily depends on the used model class and the precise setup~\citep{icpram,ida2023}. Thus, using loss-based approaches for drift detection for monitoring setups is not suitable in general.

There exist unsupervised distribution-based approaches for both model adaption and monitoring. As argued before, we are interested in unsupervised drift detection for monitoring tasks and address this topic in the following. To the best of our knowledge, no structured survey has been conducted on drift analysis for this particular task. There exists unsupervised drift detection for model adaption such as surveyed in the work~\citep{gemaque2020overview}. The authors focus on methods designed to trigger model adaption while using only few to none labeled data, essentially aiming at minimizing labeling costs and on maintaining high model performance rather than monitoring. 
Furthermore, the survey by \cite{aminikhanghahi2017survey} also discusses unsupervised change point detection which is a similar problem but in the context of time-series data which we do not consider here.

\section{Drift Detection\label{sec:drift-detection}}
As discussed before, the first important question when monitoring a data stream is \emph{whether (and when)} a drift occurs. In this section, we will focus on how this can be accomplished by means of drift detection.

\subsection{Problem Setup and Challenges}

The task of determining whether or not there is drift during a time period is called \emph{drift detection}. A method designed to perform that task is referred to as \emph{drift detector}. 

Most works on drift detection focus on the supervised scenario with abrupt drift. As discussed in \citep{icpram} this problem is very different from the problem of unsupervised drift detection, i.e., when we are interested in any kind of change and not only in changes in the label distribution given data. We will focus on the unsupervised case. Surprisingly, there does not exist a formal mathematical definition of valid drift detectors in the literature, so we provide a formalization, first.

One can consider drift detectors as a kind of statistical analysis tool that aims to differentiate between the null hypothesis ``for all time points $t$ and $s$ we have $\D_t = \D_s$'' and the alternative ``we may find time points $t$ and $s$ with $\D_t \neq \D_s$''. More formally, a drift detector is a map or algorithm that, when provided with a data sample $S$ drawn from the stream, tells us whether or not there is drift. 

We can formalize that such a drift detection model is valid or accurate, respectively, in the following way: (a) the algorithm will always make the right decision if we just provide enough data, or (b) the algorithm is a valid statistical test. This leads to the following definitions:

\begin{definition}
    A \emph{drift detector} is a decision algorithm on data-time-pairs of any sample size $n$, i.e., a (sequence of) measurable maps $A_n : (\T \times \X)^n \to \{0,1\}$. 

    A drift detector $A$ is \emph{surely drift detecting} iff it raises correct alarms in the asymptotic setting, i.e., for every drift process $\D_t$ and every $\delta > 0$ there exists a number $N$ such that for all $n > N$ we have 
\begin{align*}
    \P_{S \sim \D^n}\left[A_n(S) = \1[\text{$\D_t$ has drift}]\right] > 1-\delta.
\end{align*}
\end{definition}

Notice that the definition is not uniform across multiple streams (or drifts if the method is local in time), i.e., for some streams it suffices to have 100 samples to correctly identify drift, for others 10,000 are not enough. 
This is not a shortcoming of drift detection but a common scheme for all statistical tests. If we for example want to test whether or not the mean values of two normal distributions are the same or not we have to make estimates of the mean values. The smaller the variance and larger the number of samples, the more precise our estimate, still it is never perfect. Thus, in order to derive a reasonable decision from those estimates, the difference of the estimate needs to be smaller than the difference between the estimate and the true value. Therefore, to be able to see a shift of the means that is far smaller than the variance we need more data. By considering smaller and smaller differences we can push the number of samples required ad infinitum. Yet, the statement that we need an infinite amount of data to make any statement is not very helpful. 

To cope with that problem we have to take the two kinds of errors into account: A type I error occurs if there is no drift but we detect one (false alarm), and a type II error occurs if there is drift but we do not detect it. As discussed above, avoiding type II errors is not feasible. Also, as the effect of very mild drifts is usually less severe, missing one might as well be less problematic in practice. Thus, we focus on controlling the type I error.

Controlling the number of false alarms can be stated as follows: Once we provide a certain number of samples, the chance of a false alarm falls below a certain threshold. That number of samples must not depend on the data stream we consider. As this is also fulfilled for the trivial solution that never detects drift, we require the chance that the detector detects drift in case there actually is some to be larger than this threshold provided enough data from the stream is available. Here, the amount of required data is stream-specific as discussed above. If a drift detector fulfills these properties at least for some streams, we say that it is valid. If this holds for all streams, then we call the drift detector universally valid. Formally:
 
\begin{definition}
    A drift detector $A$ is \emph{valid} on a family of drift processes $\mathfrak{D}$, iff it correctly identifies drift in the majority of cases:
    \begin{align*}
        \limsup_{n \to \infty} &\underset{\D_t \in \mathfrak{D}, \D_t \text{ has no drift}}{\sup} \P_{S \sim \D^n}[A_n(S) = 1] \\&< \inf_{\D_t \in \mathfrak{D}, \D_t \text{ has drift}} \liminf_{n \to \infty} \P_{S \sim \D^n}[A_n(S) = 1].
    \end{align*}
    We say that $A$ is \emph{universally valid} if it is valid for all possible streams, i.e., $\mathfrak{D}$ is the set of all drift processes. 
\end{definition}

Notice that validity does not imply that $A$ actually makes the right prediction even if they make use of larger and larger sample sizes. For a concrete case, it makes no statement about the correctness of the output except that it is more likely to predict drift if there actually is drift. This probability however holds across all streams independent of the severity of the drift. Thus, for monitoring, we need a drift detector that is universally valid and surely drift-detecting.

One is frequently additionally interested in the exact time point of the drift. 
This problem is usually addressed indirectly due to the fact that if there is drift observed in a certain time window the algorithm will raise an alarm which is then considered to be the time point of drift. 

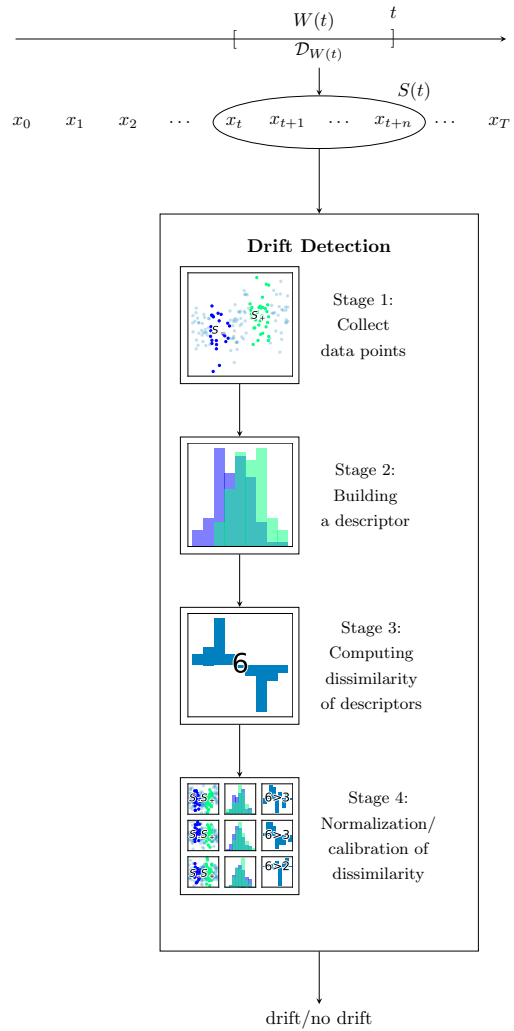
\begin{wrapfigure}{r}{0.4\textwidth}
    \centering
    \resizebox{0.4\textwidth}{!}{
    \begin{tikzpicture}
    \node (x0) at (0,0) {$x_0$};
    \node (x1) [right of = x0] {$x_1$};
    \node (x2) [right of = x1] {$x_2$};
    \node (x3) [right of = x2] {$\dots$};
    \node (x4) [right of = x3] {$x_t$};
    \node (x5) [right of = x4] {$x_{t+1}$};
    \node (x6) [right of = x5] {$\dots$};
    \node (x7) [right of = x6] {$x_{t+n}$};
    \node (x8) [right of = x7] {$\dots$};
    \node (x9) [right of = x8] {$x_T$};

    \node (start) [above =3.5em of x0] {};
    \node (end) [above =3.5em of x9] {};
    \node (windowStart) [above =3em of x4] {[};
    \node (windowEnd) [above =3em of x7] {]};
    \node (t) [above of=windowEnd, yshift=-0.5cm] {$t$};
    \node (window) [left of=t, xshift=-0.5cm, yshift=-0.5em] {$W(t)$};
    \draw [-stealth] (start.west) -- (end.east);

    \node (sample) [ellipse, draw,minimum height=1cm, minimum width=4cm, right of = x4, xshift=0.6cm]  {};
    \node (sampleT) [above right=-0.2em and -0.2em of sample] {$S(t)$};
    \node (dist) [above = 1.5em of sample] {$\D_{W(t)}$};
    
    \draw [-stealth] (dist.south) -- (sample.north);

    \node (dd) [rectangle, draw, minimum width=6cm, minimum height=14cm, below =3.5em of sample.south] {};
    \draw [-stealth] (sample.south) --  (dd.north);
    \node (ddt) [below = 1em of dd.north] {\textbf{Drift Detection}};
    \node (s1) [rectangle, draw, minimum width=2cm, minimum height=2cm, below= of dd.north, xshift=-1.5cm] {\includegraphics[width=2cm]{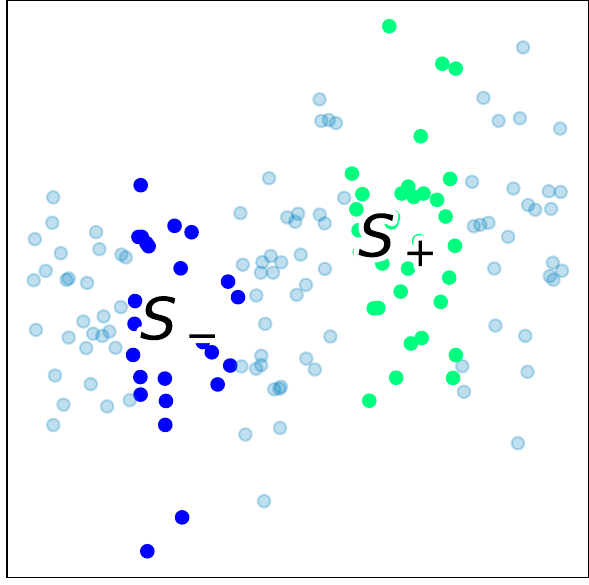}};
    \node (s2) [rectangle, draw, minimum width=2cm, minimum height=2cm, below = of s1] {\includegraphics[width=2cm]{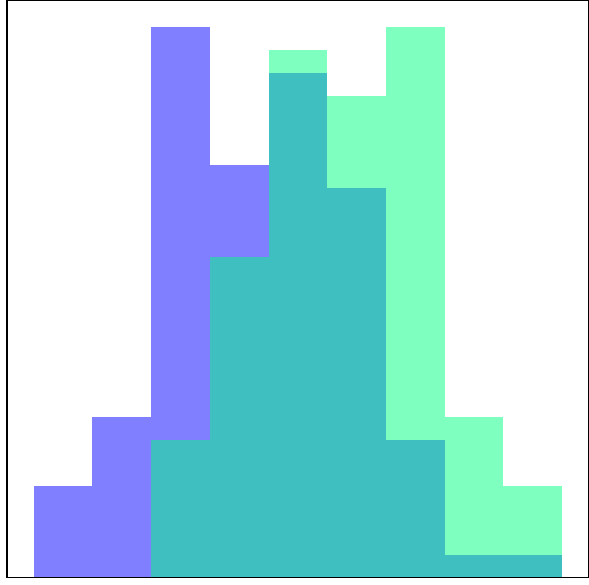}};
    \node (s3) [rectangle, draw, minimum width=2cm, minimum height=2cm, below = of s2] {\includegraphics[width=2cm]{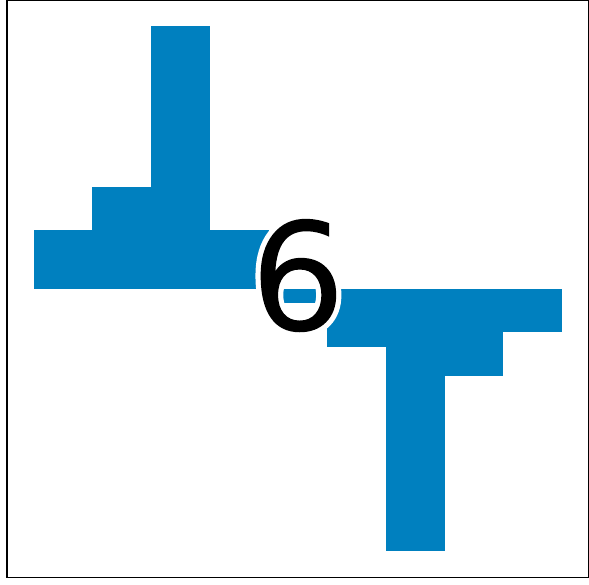}};
    \node (s4) [rectangle, draw, minimum width=2cm, minimum height=2cm, below = of s3] {\includegraphics[width=2cm]{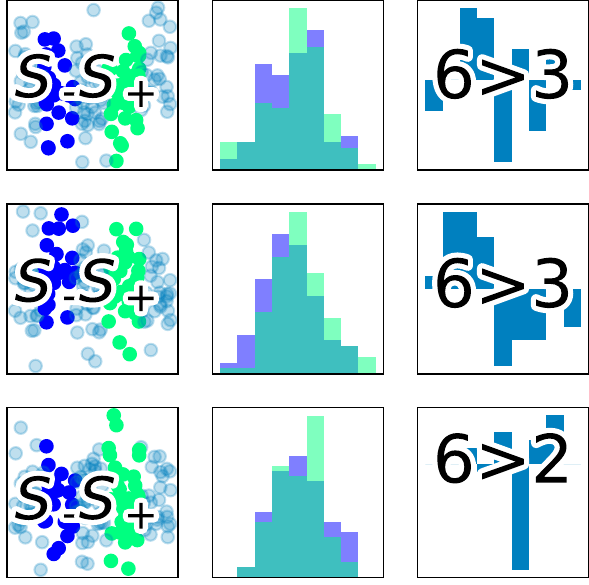}};
    \node (s1t) [right =0.2em of s1.east] {\small \begin{tabular}{c} Stage 1:\\ Collect\\data points   \end{tabular}};
    \node (s2t) [right =0.2em of s2.east] {\small \begin{tabular}{c} Stage 2:\\ Building\\a descriptor   \end{tabular}};
    \node (s3t) [right =0.2em of s3.east] {\small \begin{tabular}{c} Stage 3:\\Computing\\dissimilarity\\of descriptors   \end{tabular}};
    \node (s4t) [right =0.2em of s4.east] {\small \begin{tabular}{c} Stage 4:\\Normalization/\\calibration of\\ dissimilarity   \end{tabular}};
    \node (out) [below = of dd.south] {drift/no drift};
    \draw [-stealth] (dd.south) -- (out.north);
    \draw [-stealth] (s1.south) -- (s2.north);
    \draw [-stealth] (s2.south) -- (s3.north);
    \draw [-stealth] (s3.south) -- (s4.north);
    
\end{tikzpicture}}
    \caption{Visualization of drift detection from a data stream. Given a data stream, for each time window $W(t)$ a distribution $\D_{W(t)}$ generates a sample $S(t)$. A drift detection algorithm estimates whether the $S(t)$ contains drift or not by performing a four-stage detection scheme. Illustrated drift detector uses: two-sliding windows (stage 1), histogram descriptor (stage 2), total variance norm (stage 3), and bootstrap normalization (stage 4).}
    \label{fig:dd-stages}
\end{wrapfigure}

\subsection{A General Scheme for Drift Detection\label{sec:detection-stages}}
As discussed before, the goal of drift detection is to investigate whether or not the underlying distribution changes. As visualized in \Figname~\ref{fig:dd-stages}, drift detection is usually applied in a streaming setting where a stream of data points is arriving over time. At time $t$ a sample $S(t)$ containing some data points which are observed during $W(t)$ and thus are generated by $\D_{W(t)}$ becomes available.
On an algorithmic level, existing drift detection schemes can be described according to the four-staged fashion visualized in the figure.
Before delving into a more detailed discussion of the different existing approaches, we describe this general scheme and briefly summarize the overarching choices for realizing the four stages.

The way these stages are implemented varies depending on the specific algorithm.
In this section, we discuss some of the most prominent choices for the relevant stages 1-4 of this drift detection scheme as described in \citep{lu_learning_2018}. 

\paragraph*{Stage 1: Acquisition of data} 
\;\\
\begin{tabular}{|lm{15cm}}
    input: & data stream\\
    output: & window(s) of data samples, e.g. one reference window \\& and one containing the most recent samples
\end{tabular}
\begin{figure}
\definecolor{colorbefore}{rgb}{0.0,0.0,1.0}
\definecolor{colorafter}{rgb}{0.0,1.0,0.5}
\definecolor{colorboth}{rgb}{0.411764706,0.411764706,0.411764706}
    \centering
    \begin{tikzpicture}
    \node (start) at (0,0) {};
    \node (name) [left =of start, minimum width=2.5cm] {Fixed};
    \node (end) [right =13em of start] {};
    \draw [-stealth] (start.west) -- (end.east);
    \node (windowStart) [color= colorboth,right =2em of start] {[};
    \node (windowEnd) [color= colorboth,right =4em of windowStart] {]};
\end{tikzpicture}\\
\begin{tikzpicture}
    \node (start) at (0,0) {};
    \node (name) [left =of start, minimum width=2.5cm] {Sliding};
    \node (end) [right =13em of start] {};
    \draw [-stealth] (start.west) -- (end.east);
    \node (windowStart) [color= colorbefore, right =2em of start] {[};
    \node (windowEnd) [color= colorbefore, right =4em of windowStart] {]};
    \node (windowStart2) [color= colorafter, right =2em of windowStart] {[};
    \node (windowEnd2) [color= colorafter, right =4em of windowStart2] {]};
\end{tikzpicture}\\
\begin{tikzpicture}
    \node (start) at (0,0) {};
    \node (name) [left =of start, minimum width=2.5cm] {Growing};
    \node (end) [right =13em of start] {};
    \draw [-stealth] (start.west) -- (end.east);
    \node (windowStart) [color= colorboth, right =2em of start] {[};
    \node (windowEnd) [color= colorbefore, right =4em of windowStart] {]};
    \node (windowEnd2) [color= colorafter, right =2em of windowEnd] {]};
\end{tikzpicture}\\
\begin{tikzpicture}
    \node (start) at (0,0) {};
    \node (name) [left =of start, minimum width=2.5cm] {Implicit};
    \node (end) [right =13em of start] {};
    \draw [-stealth] (start.west) -- (end.east);
    \node (windowStart) [color= colorbefore, right =2em of start] {[};
    \node (windowEnd) [color= colorbefore, right =4em of windowStart] {]};
    \node (h) [color= colorboth, below right= 2em and 2.5em of windowStart.west, rectangle, draw] {$h$};
    \node (center) [right= 2.75em of windowStart.west] {};
    \draw [-stealth] (center) -- (h.north);
\end{tikzpicture}\\
    \caption{Illustration of Reference Window Types. Area in brackets refers to reference window $W(t), W(s)$ for time point $t < s$. Border of $W(t)$ is marked in dark blue, border of $W(s)$ in light green, overlapping borders in gray.}
    \label{fig:windows}
\end{figure}
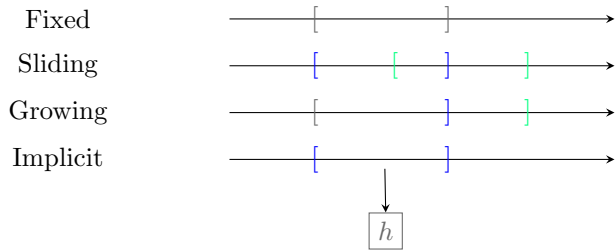

As a first step, a strategy for selecting which data points are used for further analysis needs to be selected. Depending on the strategy used (we will discuss those in Section~\ref{sec:detection-methods}) either one or two windows of the data are selected. 
Most approaches rely on sliding window strategies~\citep{lu_learning_2018}. There are four main categories that are visualized in \Figname~\ref{fig:windows}: While one can rely on a stationary fixed window that keeps the same sample until it is reset, it is also possible to use a dynamic window approach. Options for the latter are so-called growing windows, i.e. the start point stays constant and new samples are added to the window of considered data points, and sliding windows, i.e. the window keeps the same size and is shifted over the stream as new samples are arriving. Finally, the window can be implicitly realized by building a model on the corresponding data. 

Apart from these examples, some approaches use preprocessing such as a deep latent space embedding~\citep{deep-ssci}. We do not explain those possibilities in more detail.

\paragraph*{Stage 2: Building a descriptor} 
\;\\
\begin{tabular}{|ll}
    input: &  window(s) of data samples\\
    output: & possibly smoothed descriptor of window(s)
\end{tabular}

The goal of the second stage is to provide a possibly smoothed descriptor of the data distribution in the window obtained in stage 1. 

Possible descriptors are grid- or tree-based binnings, neighbor-, model-, and kernel-based approaches:
Binning can be considered as one of the simplest strategies. The input space is split into bins and the number of samples per bin is counted. The bins can be obtained as a grid or by using a decision tree.
In order to cope with the exponential growth of the number of bins with dimensions, some authors consider one-dimensional projections~\citep{hdddm}. Also, non-equally spaced grids have proved useful.
Decision trees can be constructed randomly~\citep{kdqtree} or using a criterion that takes temporal structure into account~\citep{momenttrees,ida2022} which can result in better performance. 

Instead of using machine learning techniques to find a binning, we can also use a machine learning model's data compression capabilities. To do so, we train the model and this way implicitly store the training data~\citep{shalev2014understanding,dwork2006differential,haim2022reconstructing}. Later on, the information can be accessed by performing querying. Different options for prominent choices will be discussed in Section~\ref{sec:detection-methods}.

Neighborhood-based approaches offer robust, non-parametric methods for a wide variety of applications~\citep{kNNDKL,LDD}. In this case, the adjacency matrix of the $k$-neighborhood graph is used.

Similarly, kernel methods offer another versatile, non-parametric approach~\citep{mmd,fail}. Here the entire information is contained in the so-called kernel matrix which can be considered as a pair-wise similarity.

\paragraph*{Stage 3: Computing dissimilarity} 
\;\\
\begin{tabular}{|ll}
    input: &  descriptor of window(s)\\
    output: & dissimilarity score
\end{tabular}

In the last stage, two descriptors have been obtained. The goal of this stage is to compute a dissimilarity score to analyze the windows' similarity. 
Although several approaches for building descriptors exist, many 
utilize the same or at least comparable dissimilarity measures. 
Popular choices are the total variation norm~\citep{webb2017understanding}, Hellinger distance~\citep{hdddm,webb2016characterizing}, MMD~\citep{fail,mmd}, Jensen-Shannon~\citep{esann2023,salem2012anomaly}, Kullback-Leibler divergence~\citep{kdqtree}, model loss~\citep{fail,LDD}, neighbourhood intersection\citep{LDD}, Wasserstein metric~\citep{zhao2020feature,ida2022,neucomp}, mean and moment differences~\citep{de2019learning,ida2022,neucomp}. 
Suitable combinations of dissimilarity measures and descriptors are summarized in Table~\ref{tab:summary} and will be discussed later. 
Notice, that the comparison does not have to be based on a two-sample approach but can also use more advanced strategies like independence tests into account.

\paragraph*{Stage 4: Normalization}
\;\\
\begin{tabular}{|ll}
    input: &  dissimilarity score\\
    output: & normalized dissimilarity
\end{tabular}

As the obtained dissimilarities typically depend on both, the method, i.e., stages 1-3, and the concrete distribution at hand, it is necessary to normalize the result to obtain a useful scale. One of the most common ways to do this is to relate the dissimilarity to the statistic of a statistical test. In this case, the $p$-value offers a normalized scale. In the literature, a large variety of approaches are considered. Examples are the kernel two-sample test~\citep{mmd,fail}, the Kolmogorov-Smirnov test~\citep{kswin}, the neighborhood-statistic~\citep{LDD}, or the HSIC-test~\citep{dawidd,hsic}. Also, other specific metrics like accuracy or the ROC-AUC also offer a normalized scale~\citep{d3}.

\paragraph*{Ensemble and hierarchical approaches}
Some authors~\citep{lu_learning_2018} suggest to combine multiple drift detectors. They are usually arranged in an ensemble, e.g. by combining multiple $p$-values after stage 4 into a single one, or hierarchical, e.g. by combining a computationally inexpensive but imprecise detector with a precise but computationally expensive validation. Although those approaches differ on a technical level, they do not from a theoretical perspective, as the suggested framework is sufficiently general.

\begin{landscape}
\newcommand{\symYes}{\ding{51}}%
\newcommand{\symProb}{(\ding{51})}%
\newcommand{\symNo}{\ding{55}}%
    \centering
    \begin{longtable}{p{1.5cm}cp{4cm}|p{2.5cm}p{2.5cm}p{2.5cm}p{2.5cm}|ccccc}
    \hline
    Strategy & Type & Method & Stage 1 (Reference Window) & Stage 2 & Stage 3 & Stage 4 & DD & DP & DL & DE &  \\
    \hline
    Two-Sample 
    &MB& D3~\citep{d3} & Sliding-window & Virtual-Classifier & ROC-AUC & -- & \symYes${}^a$ & \symNo & \symNo & \symNo & \\
    &ST& Window-KS~\citep{kswin,dos2016fast} & Sliding-Window & Feature-wise Empirical CDF & KS-statistic & KS-distribution & \symYes & \symNo & \symNo & \symNo \\
    &ST& MMD~\citep{mmd,fail} & Sliding-Window & Kernel-Matrix & MMD & Bootstrap test & \symYes & \symNo & \symNo & \symNo \\
    &ST& HDDDM~\citep{hdddm} & Histogram & Feature-wise Histogram & Hellinger Distance & Adaptive Threshold & \symYes & \symNo & \symNo & \symNo \\
    &ST& PCA-CD~\citep{PCACD} & Fixated-Window & KDE and Histograms on PCA-protection & maximum symmetrised Kulback-Leibler Divergence & Page-Hinkley test & \symYes & \symNo & \symNo & \symNo \\
    &ST& Drift Magnitude~\citep{webb2017understanding,webb2016characterizing} & Sliding-Window & Gird Histogram & total variation / Hellinger distance & -- & \symYes & \symNo & \symNo & \symYes \\
    &ST& $kdq$-Tree~\citep{kdqtree} & Sliding-Window & $kdq$-Tree bins & sym. Kulback-Leibler Divergence & Scanner Statistic & \symYes & \symNo & \symYes & \symNo \\
    &ST& LDD-DIS~\citep{LDD} & Sliding-window & $k$-Neighbourhood & Neighbourhood Ratio (LDD) & Bootstrap test & \symYes & \symNo & \symYes & \symNo  \\
    &ST& LSDD~\citep{bu2016pdf,bu2017incremental} & Growing-Window & Density Estimator & $L^2$-Distance of Densities & Bootstrap test + FP correction & \symYes & \symNo & \symNo & \symNo \\
    &ST& MB-DL~\citep{neucomp} & Sliding-Window & Random Forest & Kulback-Leibler Divergence to Time Independent Model & Bootstrap test & \symYes & \symNo & \symYes & \symYes${}^b$ \\
    &MB& Neighbour Density Comparison~\citep{ida2022,kNNDKL} & Sliding-Window & $k$-Neighbohood & Kulback-Leibler Divergence & -- & \symYes & \symNo & \symNo & \symNo \\
    &MB& Random Proj. Bin.~\citep{ida2022,fail} & Sliding-Window & Histogram on Random Projection & Total Variation & -- & \symYes & \symNo & \symNo & \symNo \\
    
    Meta-Statistic 
    &LB& Model+AdWin~\citep{deep-ssci} & ML Model & - & Model-loss & AdWin-Statistic & \symProb${}^c$ & \symYes & \symNo & \symNo \\
    &ST& ShapeDD~\citep{shape} & Consecutive Sliding-Windows & Kernel-Matrix & MMD & Shape Match + Bootstrap Test & \symProb${}^d$ & \symYes & \symNo & \symNo \\

    Block Based 
    &ST& DAWIDD\citep{dawidd} & Sliding-Window & Kernel-Matrix & HSIC & Bootstrap test & \symYes & \symNo & \symNo & \symNo \\
    &CL& KCpD\citep{comte2004new,harchaoui2008kernel,arlot2019kernel} & Sliding-Window & Kernel-Matrix & Kernel-Variance & None / Model Selection Heuristic & \symYes & \symYes & \symNo & \symNo & \\
    &MB& Moment Tree Binning~\citep{ida2022} & Sliding-Window & Moment Tree & Total Variation & -- & \symYes & \symNo & \symYes & \symYes${}^b$ \\
    &MB& Drift Segmentation~\citep{segmentation,neucomp} & Sliding-Window & Kolmogorov/Moment Tree & -- & -- & \symNo & \symNo & \symYes & \symYes${}^b$ \\
    \hline
    \caption{{Overview of unspervised drift analysis methods from the literature.  
    Headers stand for: Drift Detection (DD), Drift Pinpointing in time (DP), Drift Localization (DL), Drift Explanation (DE).
    Stage 1 current window is sliding window in all cases. Type refers to the type of normalization strategy used: Statistical Test (ST), model Loss Based (LB), virtual classifier / Model Based (MB), and CLustering heuristic based (CL).
    ${}^a$ depends on model but low requirements \citep{ida2022},
    ${}^b$ used by \citep{neucomp} as basis, 
    ${}^c$ depends on model, dataset and training \citep{ida2023,icpram},
    ${}^d$ for distant abrupt drifts} }
    \label{tab:summary}
    \end{longtable}
\end{landscape}

\subsection{Categories of Drift Detectors\label{sec:detection-methods}}
\begin{figure}
 	\definecolor{dimgray}{rgb}{0.41, 0.41, 0.41}
    \centering
     \resizebox{\textwidth}{!}{
    \begin{tikzpicture}[level distance=4em,level/.style={sibling distance=15em/#1}]
    \node {Drift Detectors}
    child {node {Two-Sample Analysis}
        child {node {Loss-based}
            child {node {\emph{\begin{tabular}{c}{\color{dimgray}reconstruction},\\{\color{dimgray}anomaly detection},\\{\color{dimgray}density estimation}
            \end{tabular}}}
            }}
      child {node {Virtutal classfier}
        child {node {\emph{\begin{tabular}{c}D3, MB-DL,\\$kdq$-Tree, \\LDD-DIS\end{tabular}}}
        }}
      child {node {Statistical test}
        child {node {\emph{\begin{tabular}{c}Kolmogorov-\\Smirnov (KS),\\kernel two-sample\\ test (MMD)
            \end{tabular}}}
            }}
    }
    child {node {Meta-Statistic}
        child {node {\emph{\begin{tabular}{c}{\color{dimgray}ADWIN},\\ShapeDD
        \end{tabular}}}
        }
    }
    child {node {Block}
      child {node {Independence Test}
        child {node {\emph{DAWIDD}}}
        }
      child {node {Model-based}
          child {node {\emph{Drift Segmentation}}}
      }
      child {node {Multi-Change Point}
          child {node {\emph{KCpD}}}
      }
    };
\end{tikzpicture}}
    \caption{Taxonomy of Drift Detection/Localization approaches discussed in this paper. Methods in marked in gray are supervised.}
    \label{fig:driftdetction-tree}
\end{figure}
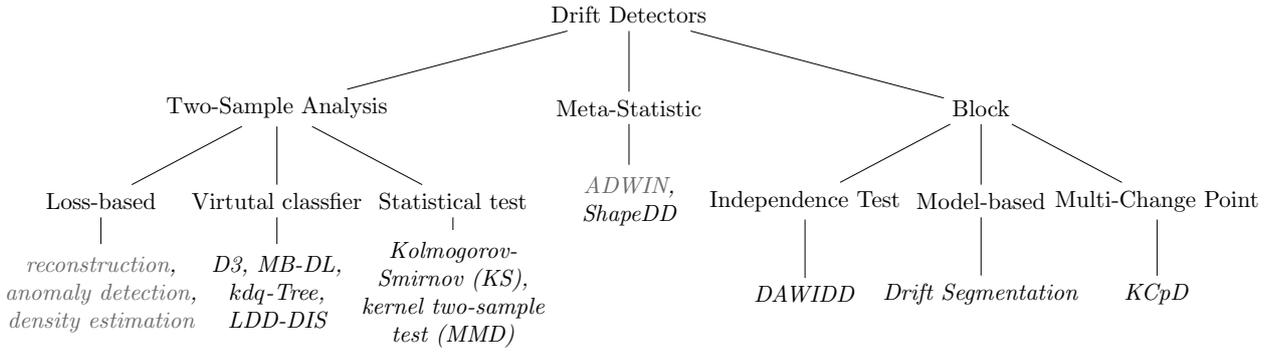

So far, we formally defined the properties a drift detection algorithm should fulfill and described on an algorithmic level how different approaches can be implemented. 
In this section, we focus on concrete approaches. We propose a categorization according to the main strategies of the approaches, relying either on an analysis of two samples, meta-statistics, or a block-based strategy. We present methods organized according to the taxonomy presented in \Figname~\ref{fig:driftdetction-tree}. An overview of the approaches considered in this survey is presented in Table~\ref{tab:summary}. We provide the overall strategy, whether the method follows a model-based, statistical test-based, loss-based, or clustering heuristic-based scheme, and how the stages discussed in the last section are realized.

\subsubsection{Two-Sample Analysis Based}

Formally drift is defined as the difference between two time points. This can be detected by applying a two-sample test, i.e., a statistical test that checks whether or not two data distributions are the same. Drift detectors based on this idea are the most common ones in the literature. 
In order to perform such a test we split our sample $S(t)$ into two samples $S_-(t)$ and $S_+(t)$ and then apply the test to those. The construction of the descriptor, distance measure, and normalization (stages 2-4) are then left to the used testing scheme. Besides classical statistical tests, there also exist more modern approaches that make use of advanced machine learning techniques.

As stated above, in order to apply this scheme we need to split the obtained sample into two sub-samples which are then used for the test. This step is crucial as an unsuited split can have a profound impact on the result. Choosing an inappropriate step can diminish the test's performance. In severe cases, it can make the drift vanish and thus undetectable as we consider time averages of the windows. However, there exists theoretical work that suggests that the averaging out does not pose a problem practice~\citep{shape}.

From a more algorithmic perspective, there are essentially three ways the testing procedure is approached. Loss-based and virtual classifier-based approaches rely on machine learning techniques while statistical test-based approaches rely on statistical tools. We will discuss those in the following.

\paragraph*{Loss-based approaches}
A large family of loss-based approaches uses machine learning models to evaluate the similarity of newly arriving samples to already received ones. In this case, the reference window (stage 1) is implicitly stored in a machine learning model which is also used as a data descriptor (stage 2). The dissimilarity is usually given by the model loss. It is further analyzed using drift detectors which are commonly used in the supervised setup~\cite{gama2004learning,baena2006early,adwin,frias2014online,basseville1993detection} and serve as a normalization (stages 3 \& 4).

There are several candidates implementing this strategy. One of the most common model choices are auto-encoders which compress and \emph{reconstruct} the data~\citep{fail}. Here, the idea is that the models are successfully compressing and reconstructing data generated by the distribution at training time but have difficulties generalizing to data generated by another distribution, i.e. if concept drift occurred in between changing the data distribution. Thus, an increase in the mean reconstruction error indicates the presence of drift.

Other popular model choices are models like 1-class SVMs or Isolation Forests. Originating from \emph{anomaly detection}, they provide an anomaly score that estimates how anomalous a data point is. The rationale for using this approach for drift detection is that data points from a new concept look like anomalous data points from the point of view of the old concept. An increase in the mean anomaly score indicates drift.  

Finally, \emph{density estimators}, which are designed to estimate the likelihood of observing a sample, can be applied to detect drift. Here the idea is that a sample from a new concept is assumed to be unlikely to be observed in the old concept, resulting in a low occurrence probability. Thus, a decrease in the mean observation probability indicates drift~\cite{kawahara2009change,yamanishi2002unifying}.

Although these methods are quite popular as they are closely connected to supervised drift detection, they also face the same issues. In particular, it has been shown on a theoretical~\citep{ida2023} and practical~\citep{icpram} level that such approaches do not work well in discovery tasks monitoring for drift as the drift might be irrelevant to the decision boundary that can be learned by the model. In particular, the resulting methods are neither surely drift detecting nor universally valid as the connection of model loss and drift is rather loose. Drift might not influence the classification as attempted by the model and thus not be detectable by means of model-loss-based detection schemes. The suitability strongly depends on the characteristics of the drift and the chosen model class. We will thus not focus on them for the rest of this paper.

\paragraph*{Virtutal classifier based approaches}
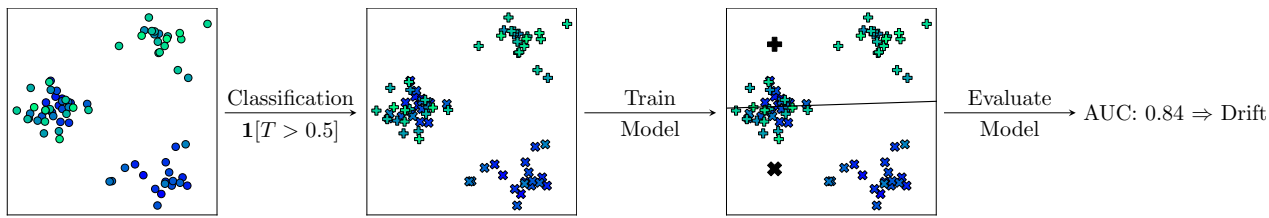
\begin{figure}
    \centering
    \resizebox{\textwidth}{!}{
    \begin{tikzpicture}
    \node (stage1) at (0,0) {\includegraphics[width=0.2\textwidth]{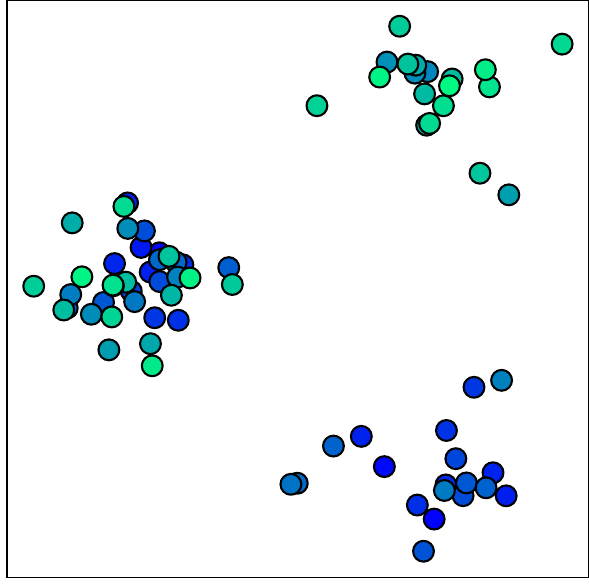}};
    \node (stage2) [right =6 em of stage1] {\includegraphics[width=0.2\textwidth]{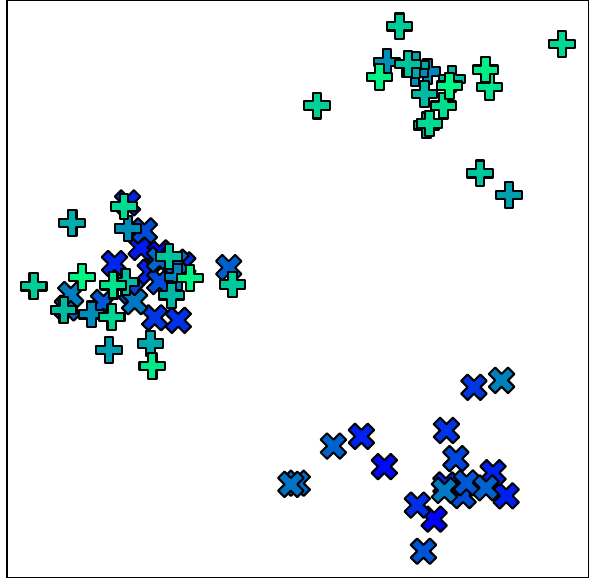}};
    \node (stage3) [right =6 em of stage2] {\includegraphics[width=0.2\textwidth]{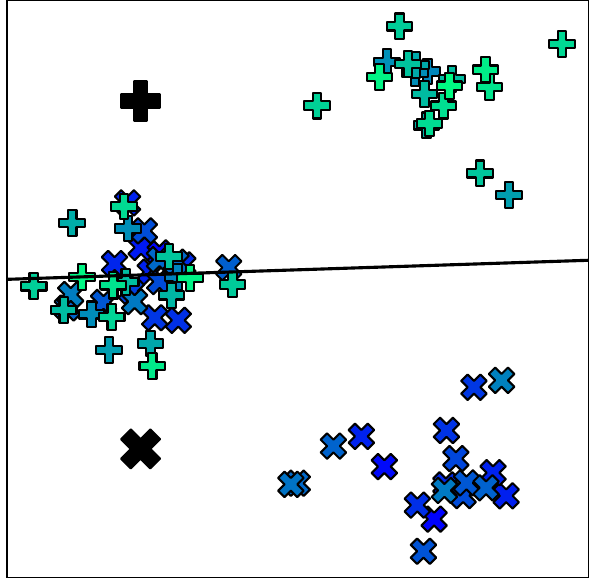}};
    \node (stage4) [right =6 em of stage3] {AUC: 0.84 $\Rightarrow$ Drift};
    \draw [-stealth] (stage1.east) -- node[above] {Classification} node[below] {$\1[T > 0.5]$} (stage2.west);
    \draw [-stealth] (stage2.east) -- node[above] {Train} node[below] {Model} (stage3.west);
    \draw [-stealth] (stage3.east) -- node[above] {Evaluate} node[below] {Model} (stage4.west);
\end{tikzpicture}
    }
    \caption{Visualisation of Virtual Classifier based Drift Detection. 1. Collect data (moment of arrival is color coded: dark blue to green), 2. Mark all samples arrived before a certain time as class -1 (cross) and after as class +1 (plus), 3. Train model to distinguish class -1 and +1, 4. Evaluate model, if performance better than random chance then there is drift.
    \label{fig:virtual-classifier}}
\end{figure}
A different approach using machine learning models is based on the following idea of virtual classifiers\citep{kifer2004detecting,hido2008unsupervised}: If a classifier performs better than random guessing, then it must exploit some properties in the data. In particular, the distribution of the single classes cannot be the same. Otherwise, it would not be able to tell the samples apart. 

This idea can be employed for drift detection as follows (see \Figname~\ref{fig:virtual-classifier} for an illustration): All data points of a reference sample $S_-(t)$ and a current state sample $S_+(t)$ are stored (stage 1). All data points in the reference sample $S_-(t)$ are labeled as class $-1$ and all data points in the current state sample $S_+(t)$ as $+1$. They are used to train a model describing the data stream (stage 2). It can either be trained anew or updated every time new samples are arriving. The model's performance, e.g., its accuracy, serves as a similarity measure for our drift detection scheme (stage 3). Depending on the precise setup, i.e., the relative size of samples, used performance measure, etc., common classification metrics can serve as normalized scores indicating drift if the model performs significantly better than random chance (stage 4).

For a practical realization of this approach, it makes sense to use $k$-folds to make optimal usage of the provided data without testing on the train set. Furthermore, using ideas from statistical learning theory, the model loss can be further refined in order to obtain actual $p$-values\citep{dries2009adaptive,kifer2004detecting}. However, this does not work well in practice as the bounds provided by learning theory are too loose. Notice, the choice of the used model class determines which types of drift can be detected in which intensity as it needs to be capable of capturing the change~\citep{ida2022}. Furthermore, the authors showed that for many learning models virtual classifiers provide surely drift detecting algorithms assuming an appropriate split point is chosen. It is also suggested that in these cases the resulting algorithms are also universally valid. Also, as pointed out by \cite{shape} the chance of choosing an invalid split point is essentially zero. 

We will consider D3~\citep{d3} as an example candidate of this class. In the default implementation by the authors, it uses two sliding windows of equal size (stage 1), logistic regression (stage 2), and ROC-AUC which is already a normalized score (stages 3 \& 4). 
There are several other approaches that use similar ideas but different models and/or evaluation metrics~\cite{LDD}.

\paragraph*{Statistical test based approaches}
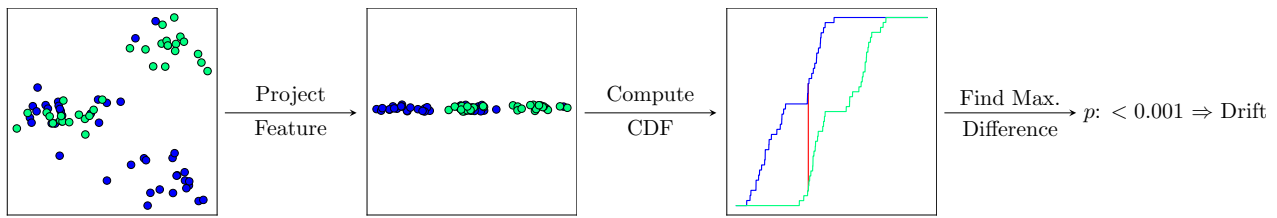
\begin{figure}
    \centering
    \resizebox{\textwidth}{!}{
    \begin{tikzpicture}
    \node (stage1) at (0,0) {\includegraphics[width=0.2\textwidth]{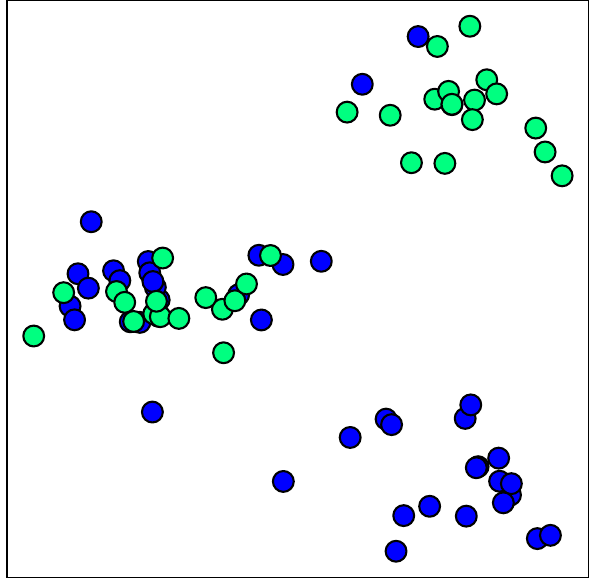}};
    \node (stage2) [right =6 em of stage1] {\includegraphics[width=0.2\textwidth]{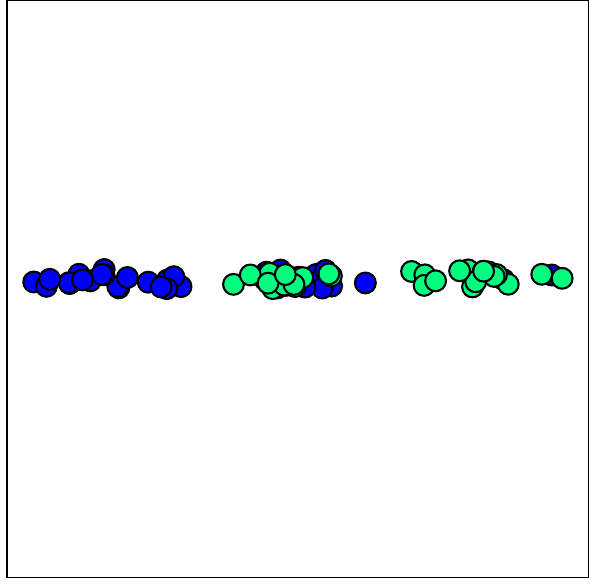}};
    \node (stage3) [right =6 em of stage2] {\includegraphics[width=0.2\textwidth]{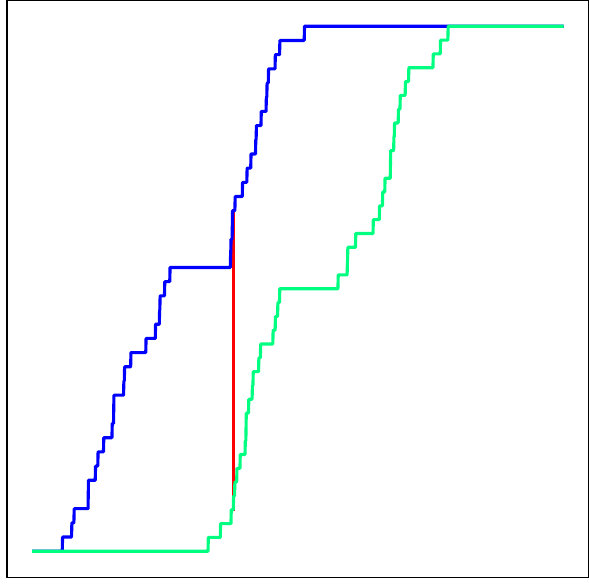}};
    \node (stage4) [right =6 em of stage3] {$p$: $<0.001$ $\Rightarrow$ Drift};
    \draw [-stealth] (stage1.east) -- node[above] {Project} node[below] {Feature} (stage2.west);
    \draw [-stealth] (stage2.east) -- node[above] {Compute} node[below] {CDF} (stage3.west);
    \draw [-stealth] (stage3.east) -- node[above] {Find Max.} node[below] {Difference} (stage4.west);
\end{tikzpicture}
    }
    \caption{Visualisation of Kolmogorov-Smirnov test for Drift Detection. 1. Collect data (two-windows; $S_-(t)$ blue and $S_+(t)$ green), 2. Feature-wise projection, 3. Compute largest difference (red line) between CDFs ($\hat{F}_{t-}$ of $S_-(t)$ and $\hat{F}_{t+}$ of $S_-(t)$) of feature-wise before and after distribution, 4. Use analytic $H_0$ distribution to obtain $p$-value.
    \label{fig:kstest}}
\end{figure}
The approaches discussed so far are usually based on intuitive considerations and can be considered as ad-hoc approaches. Another way to deal with the problem is to turn to mathematical and statistical tools that were designed for this purpose, namely two-sample tests. In contrast to the previous approaches those usually come with theoretical justifications but sometimes lack knowledge of practical situations.

Most classical two-sample tests are designed for one-dimensional data only. The most prominent candidate from this category that plays an important role in drift detection is the \emph{Kolmogorov-Smirnov (KS) test}. We have visualized the testing procedure in \Figname~\ref{fig:kstest}. The test requires two collections of one-dimensional samples as input (stage 1). Apart from this requirement, the data acquisition strategy can be chosen freely. It then computes the empirical cumulative distribution function (cdf) (stage 2):
\begin{align*}
    \hat{F}_{S_\pm(t)}(x) := \sum_{s \in S_\pm(t)} \1[s < x]
\end{align*}
which counts the number of observations smaller than the input. The similarity is then computed as the maximal distance of the two cdfs (stage 3)
\begin{align*}
    \hat{d}(S_-(t),S_+(t)) &:= \sup_{x} \left| \hat{F}_{S_+(t)}(x) - \hat{F}_{S_-(t)}(x) \right|.
\end{align*}
It has been shown in the original paper that in case $S_-(t)$ and $S_+(t)$ follow the same distribution (i.e. $\D_{W_-(t)} = \D_{W_+(t)}$, which is the case if there is no drift), then for large sample sizes $\hat{d}$ follows a distribution that does not depend on the distributions $\D_{W_\pm(t)}$~\citep{an1933sulla} allowing us to directly compute a $p$-value which can be used as a normalized scale (stage 4). 

\begin{algorithm}[!t]
	\caption{Kolmogorov-Smirnov Windowing}
	\label{algo:ks}
	\begin{algorithmic}[1]
		\Procedure{KSWIN}{$(x)$ data stream $d$-dimensional, $n_{\max2}$ current window size, $n_{\max1}$ reference window size, $p_{\text{detect}}$ detection threashold} \;
		\State Initialize Windows $W_1 \gets [], W_2 \gets []$\;
		\While{Not at end of stream $x$} \;
	    \State $W_2 \gets W_2 + [x]$ \Comment{Add new sample}
	    \If{$|W_2| > n_{\max2}$}
            \State $x \gets \textsc{pop}(W_2)$ \Comment{Move sample from current to reference window}
	        \State $W_1 \gets W_1+[x]$ 
	    \EndIf
        \If{$|W_1| > n_{\max1}$}
	        \State $\textsc{pop}(W_1)$ \Comment{Drop oldest sample}
	    \EndIf
	    \If{$|W_1| > n_{\min}$}
            \State $p \gets 1$
            \State $W_1' \gets \textsc{UniformSubsample}(W_1, n_{\max2})$
            \For{$i \in \{1, \dots, d\}$}
                \State $w_1 \gets \{x_i \mid x \in W_1'\}$\;, $w_2 \gets \{x_i \mid x \in W_2\}$ \Comment{Extract $i$-th feature from each sample}
                \State $p_d \gets \textsc{TestKS}(w_1,w_2)$
                \State $p \gets \min(p_d,p)$
            \EndFor
	        \If{$p < p_{\text{detect}}/d$}
	            \State Alert drift
	        \EndIf
	    \EndIf
		\EndWhile
		\EndProcedure
	\end{algorithmic}
\end{algorithm}

If we are dealing with higher than one-dimensional data, we can apply the procedure to every dimension separately (see Algorithm~\ref{algo:ks}). The $p$-values can then be combined using various techniques. One of the simplest is to take the minimum as it suffices that there is drift in one of the features (Bonferroni correction). A drawback of this approach is that we do not consider drift that only affects correlations as those are not considered. One way to deal with this issue is to use randomly chosen projections~\citep{ida2022,fail}. However, this does not perform well in practice~\citep{esann2023}. There are incremental versions of the KS-test that allow for a fast computation in a streaming setup~\citep{dos2016fast}.

Another important statistical test in drift detection is the \emph{kernel two-sample test}~\citep{mmd,fail} which is based on Maximum Mean Discrepancy (MMD). MMD can be thought of as the optimal performance of a collection of models:
\begin{align*}
    \text{MMD}(P,Q) := \max_{\Vert f\Vert_\mathcal{H} \leq 1} \left| \E_{X \sim P}[f(X)] - \E_{Y \sim Q}[f(Y)] \right|.
\end{align*}
In order to compute this efficiently one makes use of kernel methods, which allow us to estimate the MMD using
\begin{align*}
    &\widehat{\text{MMD}}_b(X_1,\dots,X_n \; , \; Y_1,\dots,Y_m) \\&:= \frac{1}{n^2}\sum_{i,j = 1}^n k(X_i,X_j) - \frac{2}{nm} \sum_{i = 1}^n \sum_{j = 1}^m k(X_i,Y_j) + \frac{1}{m^2} \sum_{i,j = 1}^m k(Y_i,Y_j)
    \\&= \left(\begin{array}{c}\frac{1}{n}\\\vdots\\\frac{1}{n}\\-\frac{1}{m}\\\vdots\\-\frac{1}{m}\end{array}\right)^\top
    \underbrace{\left(\begin{array}{cccccc}
        k(X_1,X_1) & \cdots & k(X_1,X_n) & k(X_1,Y_1) & \cdots & k(X_1,Y_m) \\
        \vdots & \ddots & \vdots & \vdots & \ddots & \vdots \\
        k(X_n,X_1) & \cdots & k(X_n,X_n) & k(X_n,Y_1) & \cdots & k(X_n,Y_m) \\
        k(Y_1,X_1) & \cdots & k(Y_1,X_n) & k(Y_1,Y_1) & \cdots & k(Y_1,Y_m) \\
        \vdots & \ddots & \vdots & \vdots & \ddots & \vdots \\
        k(Y_1,X_1) & \cdots & k(Y_1,X_n) & k(Y_1,Y_1) & \cdots & k(Y_1,Y_m) \\
    \end{array}\right)}_{= K} \left(\begin{array}{c}\frac{1}{n}\\\vdots\\\frac{1}{n}\\-\frac{1}{m}\\\vdots\\-\frac{1}{m}\end{array}\right)
\end{align*}
where $k$ is a kernel function and the matrix $K$ is the kernel matrix

Similar to the KS test, the MMD test requires two collections of sample points (stage 1). However, in contrast to KS those do not need to be one-dimensional, indeed, the test can even deal with non-vectorial data as long as a kernel can be defined on that data. The descriptor is then given by the kernel matrix $K$ based on the reference and current state sample (stage 2). This allows the computation (an estimate) of the MMD by multiplying it with an appropriate weight vector from both sides (stage 3). As the resulting score is heavily influenced by the choice of the kernel and the dataset a normalization is necessary. To do so the authors~\citep{mmd} suggest either using a normalization based on the higher moments of the $\widehat{\text{MMD}}_b$ estimator or using a permutation boot-strap approach that compares the MMD of the original split with randomly permutated ones which both provide us with $p$-values (stage 4). Unfortunately, the moment-based approach tends to be less robust. Thus, we will make use of the permutation approach. 
There are several more approaches that follow similar lines or arguments based on various descriptors or metrics~\citep{harchaoui2008kernel,harchaoui2009regularized,chen2015graph,bu2016pdf,bu2017incremental,rosenbaum2005exact}.

A drawback of all the aforementioned approaches is that the selection of the split point is somewhat random but still has a huge influence on the resulting performance. Furthermore, as we perform the testing on sometimes heavily overlapping time windows we face the multiple testing problem, i.e., the fact that statistical analysis and interference tools become less reliable if applied to the same set of observations over and over again. One way to deal with this issue is using meta-statistics which do not consider every time point/time window in isolation but rather compare the resulting values. 

As the notion of validity of a drift detector is derived from considering the problem as a statistical test, most test base approaches are valid if test specific criteria are meat, e.g., as we will see in Section~\ref{sec:drift-experiments} the feature-wise Kolmogorov-Smirnov tests assumes that the drift is not contained in the correlation only. However, it is not clear how to derive a surely drift detecting algorithm for a statistical test: Although, the $p$-value tends to 0 as the number of samples increases, assuming there is drift, it does not tend to 1 in the non-drifting cases so that an appropriate threshold has to be chosen which can be complected in practice. Furthermore, as in all two-sample cases, a split point has to be chosen which can have a strong effect on the performance of the test~\citep{ida2022,shape} (see Section~\ref{sec:drift-experiments}). However, the chance of choosing a split point that invalidates the detection is very unlikely~\citep{shape}.

\subsubsection{Meta-Statistic Based}

So far we have been dealing with two-sample approaches. In a sense, those are the simplest approaches as they consider every time point in the stream separately which leads to issues like the multiple testing problem, sub-optimal sensitivity, and high computational complexity. 
Meta-statistic approaches try to deal with some of these issues by not considering each estimate separately but rather combining the values of several estimates to get better results. 
Though there are several algorithms that combine the result of multiple drift detectors (ensemble approaches), those are usually applied at the same time point. To the best of our knowledge, there are only very few algorithms that fall into this category. We will describe two algorithms in detail.

\paragraph*{AdWin} 
\begin{figure}
    \centering
    \resizebox{\textwidth}{!}{
    \begin{tikzpicture}
    \node (stage1) at (0,0) {\includegraphics[width=0.2\textwidth]{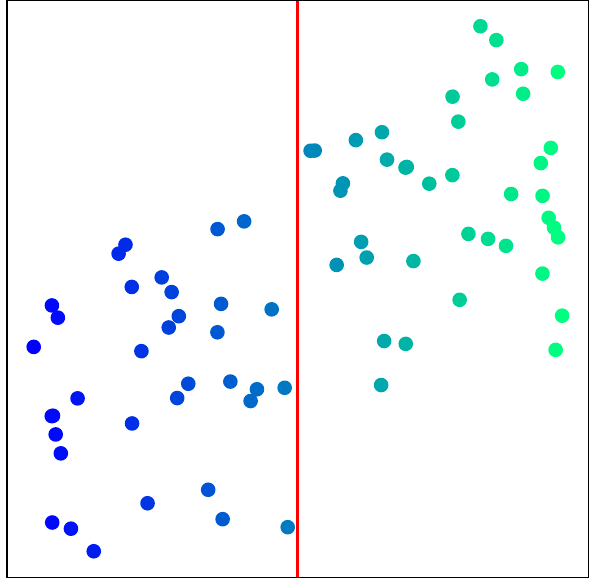}};
    \node (stage2) [right =6 em of stage1] {\includegraphics[width=0.2\textwidth]{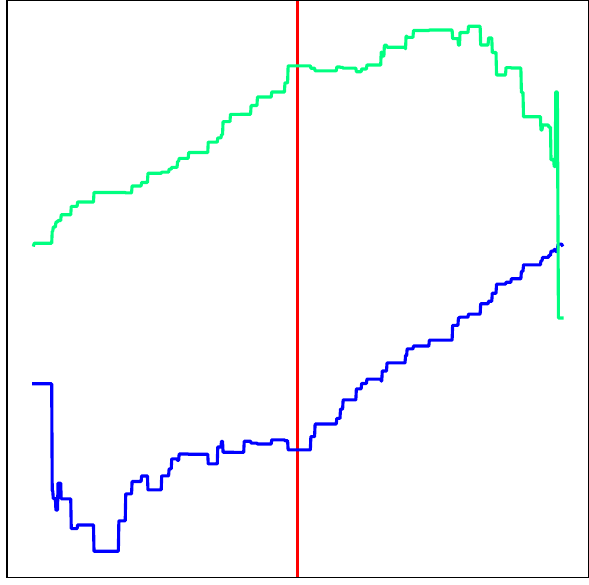}};
    \node (stage3) [right =6 em of stage2] {\includegraphics[width=0.2\textwidth]{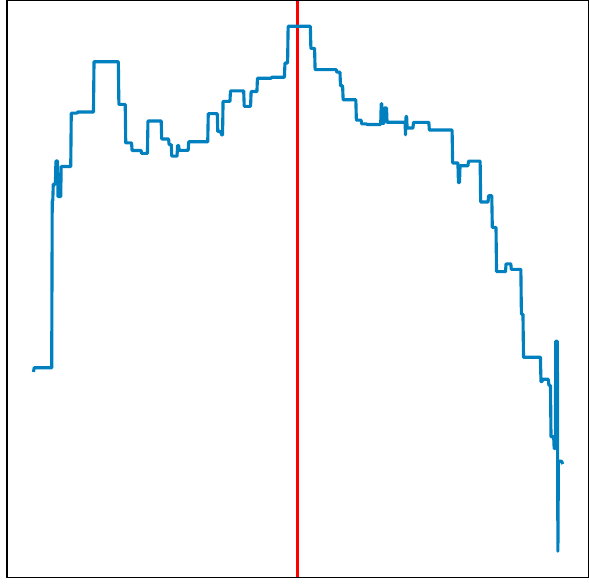}};
    \node (stage4) [right =6 em of stage3] {$p$: $<0.001$ $\Rightarrow$ Drift};
    \draw [-stealth] (stage1.east) -- node[above] {Compute} node[below] {Moving Means} (stage2.west);
    \draw [-stealth] (stage2.east) -- node[above] {Compute} node[below] {Difference} (stage3.west);
    \draw [-stealth] (stage3.east) -- node[above] {Find Max.} node[below] {Difference} (stage4.west);
\end{tikzpicture}
    }
    \caption{Visualisation of AdWin Drift Detection. 1. Collect data ($S(t)$ values between 0 and 1; red line marks drift), 2. Compute moving means ($\mu_{s-}(t)$ blue $\mu_{s+}(t)$ green), 3. Find largest difference, 4. Use analytic $H_0$ distribution to obtain $p$-value.}
    \label{fig:adwin}
\end{figure}
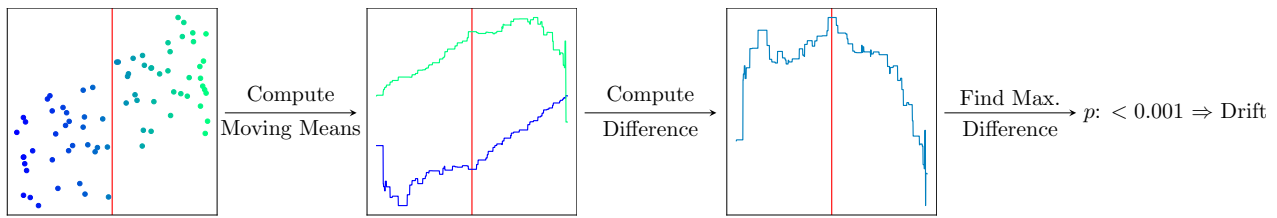
AdWin~\citep{adwin} stands for ADaptive WINdowing and is one of the most popular algorithms in supervised drift detection. It takes values between 0 and 1 as input, which are interpreted as model losses. However, other inputs such as $p$-values are also possible. We have illustrated the workings of AdWin in \Figname~\ref{fig:adwin}: First, it stores the values in a single growing or sliding window $S(t)$ (stage 1). For every time point $s \in W(t)$, this window is split $S(t) = S_{s+}(t) \cup S_{s-}(t)$ and the mean value $\mu_{s\pm}(t)$ (and variance) are computed for both halves (stage 2). The (variance normalized) difference of mean values then provides the statistic of interest (stage 3):
\begin{align*}
    \hat{d}(t) &= \sup_{s \in W(t)} \left| \mu_{s+}(t) - \mu_{s-}(t) \right|.
\end{align*}
Using the assumption that the samples are Bernoulli random variables, i.e., take on the values 0 and 1 only, as is the case for classification errors one can compute $p$-values for the $H_0$-hypothesis that the performance of the model only becomes better over time (stage 4). If the $H_0$-hypothesis is rejected then the model's accuracy has decreased indicating a drift. The time point of the largest discrepancy is considered as the moment the drift took place. 

There are algorithmic solutions that allow for an efficient, incremental computation of the AdWin statistic. However, as stated before the connection of model-loss as drift is rather vague~\citep{icpram,ida2023} thus we will exclude AdWin from further considerations. 

\paragraph*{ShapeDD} 
\begin{figure}
    \centering
    \resizebox{\textwidth}{!}{
    \begin{tikzpicture}
    \node (stage1) at (0,0) {\includegraphics[width=0.2\textwidth]{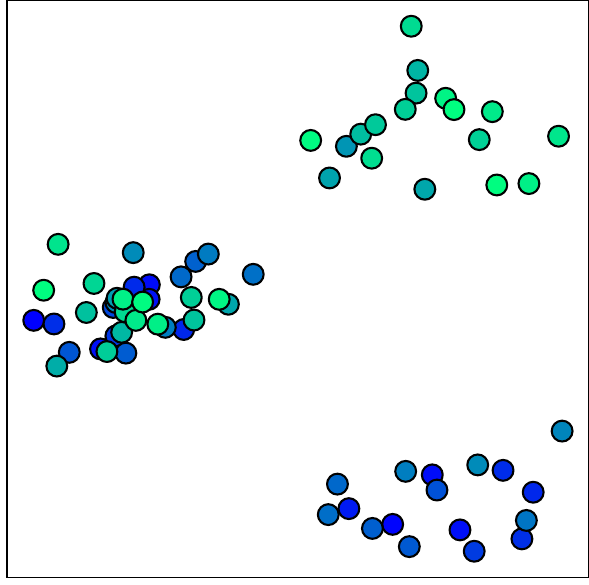}};
    \node (stage2) [right =6 em of stage1] {\includegraphics[width=0.2\textwidth]{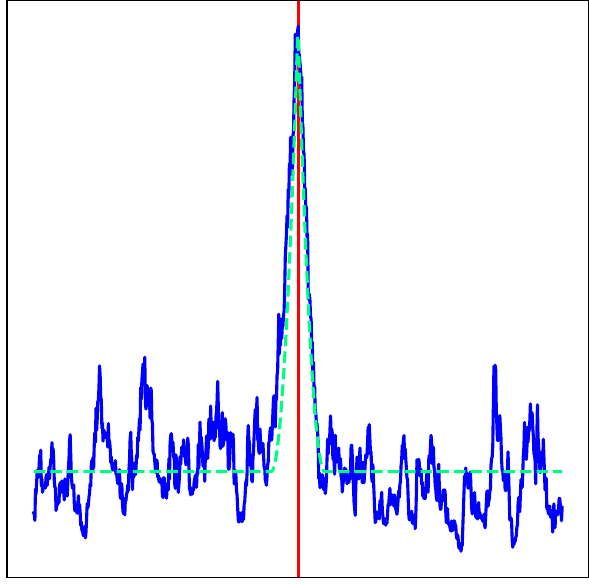}};
    \node (stage3) [right =6 em of stage2] {\includegraphics[width=0.2\textwidth]{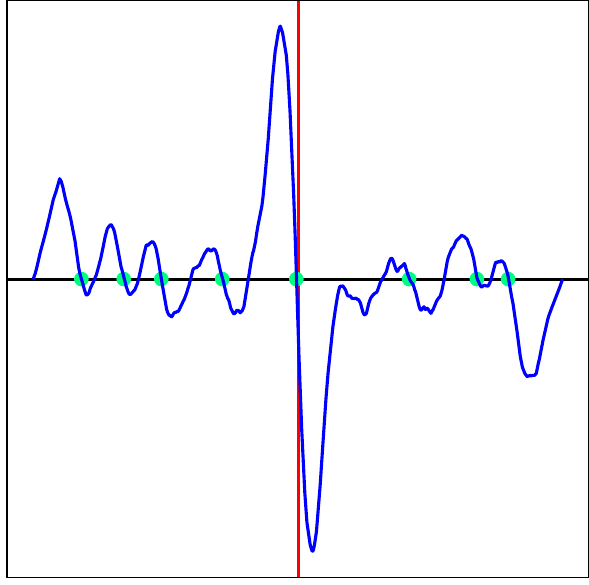}};
    \node (stage4) [right =6 em of stage3] {$p$: $<0.001$ $\Rightarrow$ Drift};
    \draw [-stealth] (stage1.east) -- node[above] {Moving} node[below] {MMD} (stage2.west);
    \draw [-stealth] (stage2.east) -- node[above] {Shape} node[below] {Match} (stage3.west);
    \draw [-stealth] (stage3.east) -- node[above] {Compute} node[below] {Bootstrap} (stage4.west);
\end{tikzpicture}
    }
    \caption{Visualisation of Shape Drift Detector. 1. Compute MMD for all time points ($\hat{\sigma}$, dotted line shows theoretically expected shape $\sigma$, red line indicated time point of drift), 2. Match obtained shape with theoretically expected one ($\hat{\sigma}*w$), 3. Candidate points are where the match score changes from positive to negative (black line is 0, dots mark candidates), 4. Compute $H_0$ distribution for MMD using permutation bootstrap at candidate points to obtain $p$-value.\label{fig:shape}}
\end{figure}
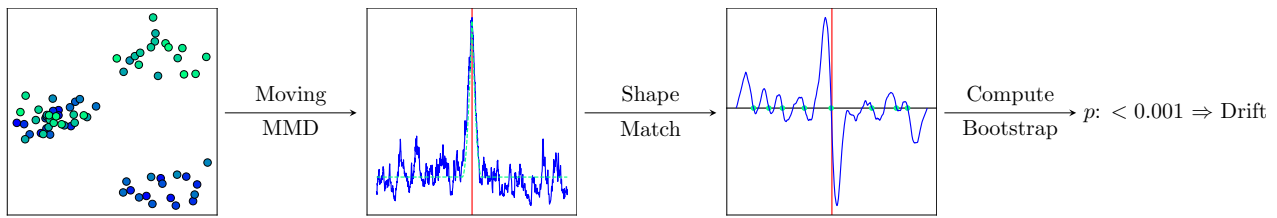
The Shape Drift Detector~\citep{shape} is another meta-statistic-based drift detector. 
In contrast to AdWin it focuses on the discrepancy of two consecutive time windows, a quantity referred to as drift magnitude~\citep{webb2017understanding}:
\begin{align*}
    \sigma_{d,l,\D_\cdot}(t) = d(\D_{[t-2l,t-l]}, \D_{[t-l,t]}).
\end{align*}
Here $d$ is allowed to be any (quasi-)metric. This includes several dissimilarity measures used in other drift detectors: total variation norm~\citep{webb2017understanding}, Hellinger distance~\citep{hdddm}, MMD~\citep{fail,mmd}, Jensen-Shannon-metric (A metric derived from the Kullback-Leibler divergence~\citep{kdqtree}), feature-wise maximum KS statistic~\citep{kswin}, and classification accuracy linear models with non-linear preprocessing~\citep{d3}.  
The core finding of \citep{shape} is that the drift detector is based on is the fact that (for quasi-norms $d$) the graph of $\sigma$ takes on a characteristic shape (see \Figname~\ref{fig:shape}) in case there is drift. This shape does not depend on the dataset or the used distance measure but only on the length of the sliding window. Thus, in contrast to many other drift detectors, ShapeDD directly takes the fact that it makes use of sliding windows into account. In case there are several drift events, the same shape is repeated once for each drift assuming the drifts are sufficiently far apart. Finding those parts of $\sigma$ that match the characteristic shape allows for a precise pinpointing of the drift event.

More algorithmically, in the default setup as described by the authors, ShapeDD uses the MMD. We thus use two consecutive sliding windows to compute the MMD using the kernel matrix $\hat{\sigma}(t)$ (stages 1-3). This can be done in an online fashion in linear time with respect to the window length. 
This has two advantages: First, the MMD is known to work quite well for a large number of different datasets. Second, the computationally expensive part of the MMD-based drift detector is the computation of the normalization rather than the simple statistic. 

To match the shape we then compute the convolution $(\hat{\sigma} * w)(t)$ with $w(t) = -1/l$ for $-2l \leq t < -l$, $=1/l$ for $-l \leq t < 0$ and 0 otherwise, which can be done iteratively in constant time and check for consecutive values where the sign changes from positive to negative, indicating a perfect match of the shape. The obtained points are then used to compute the MMD in a two-window fashion using the found position as the split point (stage 4).
As a consequence, most potential split points are not considered in the first place, reducing the average computational complexity of the method.
Furthermore, it is less likely to find several candidate points in close proximity, reducing the change of a multiple testing problem and thus increasing the reliability of the method.
Furthermore, in some cases, the precise time point of the drift event can be of interest, which is not provided by the two-window approaches. Additionally, it is unlikely that the same drift is found twice which is also not true for the other approaches.
However, the characteristic shape is, in fact, an artifact that results from the way the sampling procedure interacts with a single drift event, and thus no longer holds true if we consider a different windowing scheme (stage 1), several drift events in close succession, or gradual drift.

One way to solve the latter issue is to make use of even more advanced meta-statistics that analyze the entire data block at once.

Besides from that the authors discuss in the original paper that ShapeDD resolves the problem of choosing the right split point, which together with the validity of the MMD two-sample test shows that the method is valid for all drift processes with abrupt drifts that are sufficiently far apart. However, as ShapeDD essentially reduces the drift detection to the application of MMD, it suffers the same issues regarding being surly drift detecting.

\subsubsection{Block Based}

In contrast to all other drift detectors considered so far, block-based methods do not assume a split of the data into two windows at any point. Instead, they take an entire data segment into account and analyze it at once. To the best of our knowledge, so far there is only one block-based drift-detection algorithm in existence. 

\paragraph*{Independence Test based}  
DAWIDD~\citep{dawidd} is derived from the formulation of concept drift as a statistical dependence of data $X$ and time $T$ and thus resolves drift detection as a test for statistical independence. Here we will make use of the HSIC-test~\citep{hsic} which similarly to MMD makes use of kernel methods. However, instead of searching for a map that discriminates the two datasets it searches for a pair of maps that increases the correlation, i.e., $\sup_{f : \T \to \R, g: \X \to \R} \text{cov}(f(T),g(X))$ where similar to MMD $f$ and $g$ are found using kernel-methods. The test requires a single collection of data points and thus a sliding window (stage 1). If available, the real observation time points can be used, otherwise \citep{dawidd} suggested to simply making use of the sample id, i.e., sample $X_i$ was observed at time $T_i = i$. Using HSIC we compute the kernel matrix of data $K_X$ and time $K_T$ as a descriptor (stage 2). The HSIC statistic is then a measure for the dependence of data $X$ and time $T$ and is estimated by $\text{trace}(K_X H K_T H)$, where $H = I - n^{-1}\mathbf{1}\mathbf{1}^\top$ is the kernel-centering matrix (stage 3). Similar to MMD the HSIC can be normalized using higher moments or a permutation bootstrap approach (stage 4). Due to better performance, we make use of the latter. Notice that if the actual observation time is not available we can use the same time kernel matrix $K_T$ and thus precompute $H K_T H$ as well as the permutated versions resulting in a drastic reduction in computation time. 

Compared to the other approaches, DAWIDD makes the fewest assumptions on the data or the drift. Not only can it deal with gradual, and reoccurring drift but can even be applied in cases where there is no linear, temporal order as in the case of federated learning where every computational node can experience drift at different points in time. However, this generality comes at the cost that we usually need more data to obtain the same level of certainty. We can consider this as the usual complexity-convergence trade-off found in all areas of machine learning. 

As DAWIDD is again a statistical test it is also universally valid but suffers the same issues regarding surely drift detectability. However, in contrast to the two-sample test base methods, we do not have to choose a valid split point which can be problematic in practice. 

\paragraph*{Clustering based} 
Another way to perform block-based drift detection is provided by clustering-based methods. In contrast to independence test-based approaches, these assume that there are only finitely many, abrupt drifts in the considered time interval. The methods then proceed by clustering the data points with the additional restriction that all data points that belong to one cluster have been observed at consecutive time points. This is a reasonable approach as the variance of several data points coming from the same concept is small, while the variance of data points coming from different concepts is large. Furthermore, since the time points of all observations belonging to one concept have to be a time interval and every time point has to belong to one interval it suffices to find the boundary points of the intervals which then correspond to change points of the drift process. Using a distributional variance measure $V$, i.e., a generalization of variance, such algorithms solve an optimization problem of the following form for a predefined number $n$:
\begin{align*}
    \underset{t_0 < \dots < t_n}{\text{arg}\;\min} \sum_{i = 0}^{n-1} w(t_{i+1}-t_i) V(\D_{(t_i,t_{i+1}]}),
\end{align*}
where $\T = (t_0,t_n]$ and $w$ is a weighting function. 

An instantiation of this approach was proposed by \cite{comte2004new,harchaoui2007retrospective} making use of a kernalized version of variance, i.e., $V(P) = \sup_{\Vert f \Vert \leq 1} \text{var}_{X \sim P}(f(X))$, that is closely connected to the MMD and constant weighting $w(\Delta) = \Delta/n$~\citep{arlot2019kernel}. Thus, for $X_1,\dots,X_m \sim P$ we have the estimate $\hat{V}(X_1,\dots,X_m) = \frac{1}{m}\sum_{i = 1}^m k(X_i,X_i) - \frac{1}{m^2}\sum_{i,j = 1}^m k(X_i,X_j)$. Due to the linear nature of the kernel-variance, i.e., it can efficiently be computed from the integral kernel matrix, the problem can efficiently be solved using a dynamic programming approach that assures to always find the optimal solution in polynomial time. The resulting algorithm is commonly called Kernel Change-point Detection (KCpD). 
Later on, \cite{arlot2019kernel} introduced a heuristic to estimate the number of change points by making use of ideas that originate in model selection. More precisely, one considers the same objective as before for different numbers of split points $n$. If there are more than $n$ change points then we can find a good split and end up with a much smaller value for $n+1$. On the other hand, if there are no more than $n$ change points, then the obtained decrease of the objective is only due to statistical insufficiencies and therefore comparably small. 

From a more algorithmic point of view, KCpD essentially searches for blocks along the main diagonal of the kernel matrix so that the mean value of the entries inside the blocks is maximized. The number of blocks is then chosen such that more blocks no longer increase that value significantly. 

Notice that there is a close connection between clustering-based algorithms and independence test-based algorithms: In the case of HSIC, by restricting the class of functions $f: \T \to \R$ to indicator functions on intervals, i.e., $f(t) = 1$ for $a < t \leq b$ and $0$ otherwise, we force the test to search for functions $g$ such that $g(X)$ is large whenever $T \in (a,b]$ and small everywhere else. Since the norm of the functions is bounded, this implies that the variance of $g(X) \mid T \in (a,b]$ is small. 
Thus, KCpD can be considered as an instantiation of an independence-based detection scheme. 

Since KCpD is a mainly heuristic methods it is hard to make any statement about its limiting behaviour. However, in the statistic of the 1-split point case is very similar to the once considered by \cite{ida2022} furthermore it is well known that in many cases kernel-estimates have uniform convergence rates. It is thus reasonable that one can derive universally valid surely drift detesting methods that make use of the same ideas.

There are more algorithms that use similar ideas~\citep{keogh2001online}.

\paragraph*{Model based} Besides the classical kernels which are  predefined and not dataset specific, we can also construct new kernels using machine learning models. In \citep{ida2022}, random forests with a modified loss function that is designed for conditional density estimation, so-called moment trees \citep{momenttrees}, are used to construct such kernels. To do so the model is trained to predict the time of observation $T$ from the observation $X$. The resulting kernels show drastic improvements in drift detection tasks~\citep{ida2022}. We can also apply this procedure directly to obtain model-based block-based approaches that can be thought of as an extension of the classifier-based two-window approaches to continuous time by removing the time discretization. The relation between the resulting approaches to DAWIDD is then very similar to the relation of MMD to the model-based two-window approaches.

\subsection{Analysis of Strategies}
\label{sec:drift-experiments}
So far, we categorized different drift detection schemes and described them according to the four stages discussed in Section~\ref{sec:detection-stages}. In this section, we will consider the different strategies on a more practical level and investigate experimentally in which scenarios which drift detection method is most suitable. For this purpose, we identified four main criteria that describe the data stream and the drift we aim to detect: we investigate the role of the \emph{drift strength}, the influence of drift in \emph{correlating features}, the \emph{data dimensionality}, and the \emph{number of drift events}. To cover the strategies described in Section~\ref{sec:detection-methods}, we select one representative technique per category. As these approaches are structurally similar, from a theoretical viewpoint they carry the same advantages and shortcomings. We will present and discuss our findings in the remainder of this section.

\paragraph*{Experimental Setup}
\begin{figure}
    \begin{center}
            \begin{tabular}{ccc}
                \includegraphics[width=0.3\textwidth]{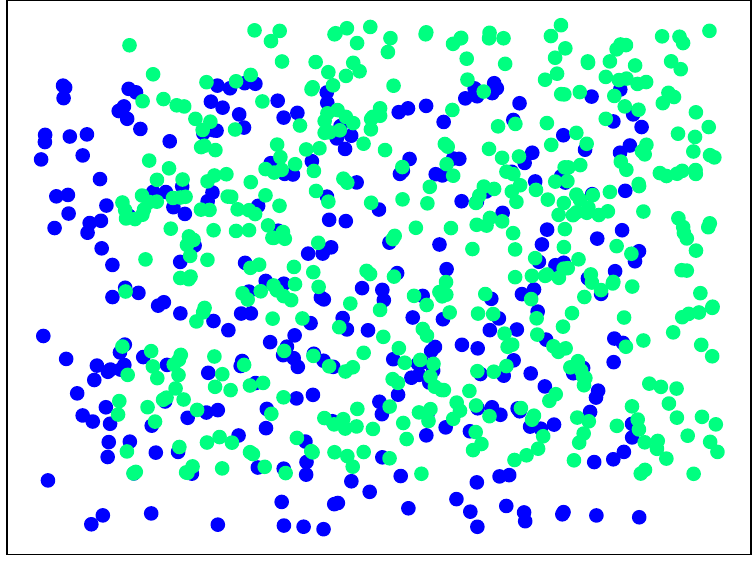} &
                \includegraphics[width=0.3\textwidth]{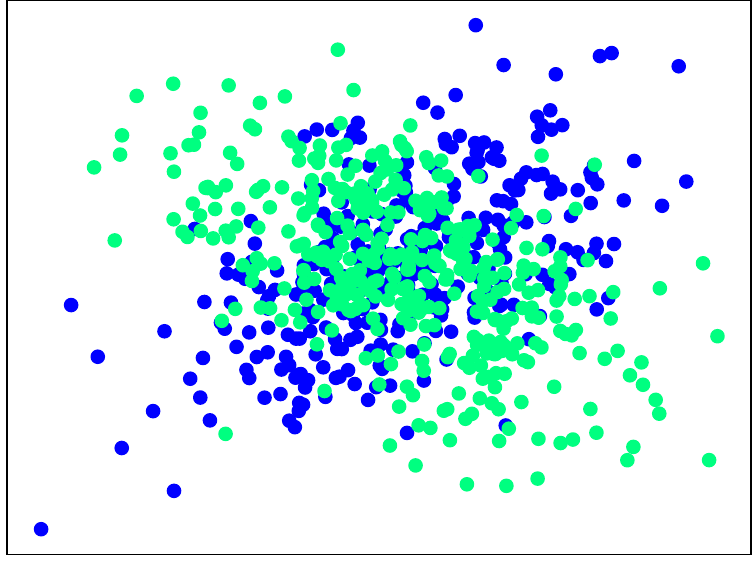} &
                \includegraphics[width=0.3\textwidth]{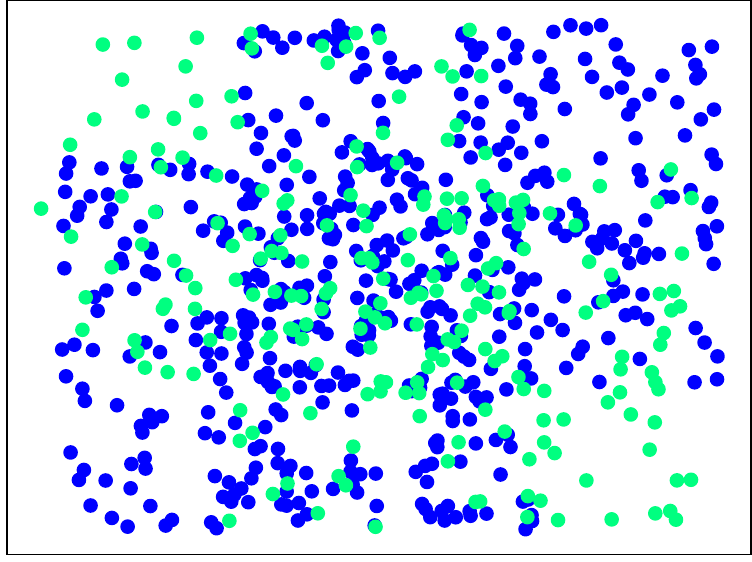} \\
                uniform & Gauss & two overlap \\
            \end{tabular}
    \end{center}
    \caption{Illustration of used datasets (default parameters, original size). Concepts are color coded (Before drift: blue, after drift: green).}
    \label{fig:datasets}
\end{figure}
\emph{Datasets:} 
For our experiments, we consider three 2-dimensional, synthetic datasets with differently structured abrupt drift (see \Figname~\ref{fig:datasets}). We use modifications of these datasets to evaluate the properties of the discussed drift detection methods. 

For the first dataset, we draw samples from a \emph{uniform} distribution on a square. The drift is induced by performing a uniform shift along the diagonal, i.e., in $x$- and $y$-direction. Here, the drift intensity describes the length of that shift. Potential extra dimensions contain uniform noise. 
In this dataset, the drift is thus described by means of a shift, which is not uncommon in practice, but a rather simple setup.

The second dataset is generated by drawing samples from a \emph{Gauss} (normal) distribution. The drift intensity is characterized by the absolute value of the covariance. The drift changes the sign of the covariance, i.e., the drift corresponds to a $90^o$ rotation of the distribution. Potential additional dimensions are normally distributed with covariance 0.
In this dataset, the drift is thus encoded in the correlation only, however, normal distributions are a relatively easy problem to deal with for many algorithms.

For the third dataset, samples are drawn from \emph{two overlapping} uniform distributions on squares. The intensity describes the offset of the two squares. The drift corresponds to a $90^o$ rotation of the distribution. The noise dimensions are also uniformly distributed. In this dataset, drift is again encoded in correlation only. However, the distribution is not as simple as a normal distribution.

Based on these basic datasets we generate data streams consisting of 750 samples with drift times randomly picked between $t=100$ and $t=650$ by varying the following parameters:
\begin{itemize}
    \item We vary the drift \emph{intensity} of the previously described datasets, the default is set to $0.125$.
    \item We modify the \emph{number} of drift events in the given data streams, the default value is 1.
    \item Finally, we consider adding noisy dimensions. As described above, each stream contains two drifting dimensions; additional dimensions are noise according to the dataset description. We  control the total number of \emph{dimensions}, the default is set to 5, i.e., 3 non-drifting noise dimensions.
\end{itemize}

\emph{Methods:}
We make use of the following algorithms that cover every major type and sub-type of drift detector discussed in section~\ref{sec:drift-detection}: D3\footnote{We use the implementation provided by the authors \url{https://github.com/ogozuacik/d3-discriminative-drift-detector-concept-drift}}, KS, MMD, ShapeDD, KCpD\footnote{We use the implementation provided by \cite{JH2020,JCRAH2021}.}, and DAWIDD. 
For D3 we used the default logistic regression as well as random forests as model. For MMD, ShapeDD, and DAWIDD we used the RBF-kernel and $2{,}500$ bootstrap permutations. 

Except for KCpD all methods provide a drift score which we make use of for the evaluation. For KCpD use the extension presented in~\cite{arlot2019kernel} which estimates the number of drifts (aka change points) given a predefined threshold parameter $\alpha$ that plays a similar role to a $p$-value threshold. We thus use the smallest $\alpha$ value as drift score. 

In order to apply the two-sample (D3, KS, MMD) and block-based (DAWIDD, KCpD) algorithms, we split the stream into chunks of 150 and 250 samples with an overlap of 100 samples of consecutive chucks. We apply the drift detectors separately to each chunk. For the two-window-based approaches, we always choose the midpoint as the split point. This way, we obtain a drift score, $p$-value, ROC-AUC in the case of D3, for each chunk. 
For the meta-statistic approaches (ShapeDD) we apply the algorithms directly to the stream obtaining a drift score for each sample. To obtain the chunk-wise drift score, we take the minimum over all scores associated with samples that belong to that chunk. 

Furthermore, as KCpD also allows for a precise pinpointing of the drift we also apply it to the entire stream and then make use of the same strategy as in the case of ShapeDD. Notice that this setup is not comparable to the others as it makes use of far more data and is not online. However, the pinpointing capabilities can be important in further downstream tasks. 

\emph{Evaluation:}
For each setup we perform 500 runs on independently sampled streams. 
To evaluate the methods we make use of the ROC-AUC that is defined as the area under the receiver operating characteristic (ROC) curve. Given a number of class scores for binary labeled samples, the ROC-curve plots the true positive rate against the false positive rate for various decision thresholds on the class scores. This way, we get an impression on how many false positives we have to tolerate if we want a certain number of true positives. By taking the area under the curve (AUC) this information is condensed into a single number taking on values between 0 and 1 with 1 a perfect separation, 0.5 random chance.
The ROC-AUC has the benefit that we do not have to specify a decision threshold, rather it provides an upper bound on the performance for every threshold. Notice that it is problematic to choose a threshold: on the one hand, it is not realistic in practice to optimize the threshold, because we usually do not have ground truth, but on the other hand, it is necessary as an inappropriate choice can render the method useless. The ROC-AUC measures how well the scores for the drifting and non-drifting cases can be separated and thus provides an upper bound on the expected performance. Furthermore, the ROC-AUC is not affected by class imbalance and thus a particularly good choice as the number of chunks with and without drift is not the same for most setups.

\paragraph*{Drift strengths}
Depending on the application it might be desirable to employ a drift detector that is capable of robustly detecting very small drifts. To evaluate how well the discussed strategies capture small drifts we run experiments with increasing drifts on all described datasets. 
From a theoretical perspective, it is reasonable from most setups that weaker drifts are harder to detect. 
The sensitivity of the method often severely  depends on the choice of meta-parameters. 
For example: If we use D3 with different models, the performance of the model has an impact on how well the derived drift detection works. Thus, if we make use of a comparably simple model like logistic regression we expect to be less sensitive to drift compared to models like $k$-nearest neighbor or random forests. Similar effects can be found in all drift detectors regarding the respective descriptors, i.e., the choice of the kernel has an impact on the performance of MMD, ShapeDD, and DAWIDD, or the way we choose the projection for KS, i.e., random, sparse, axis-aligned~\citep{ida2022,fail,esann2023}.

\begin{figure}
    \centering
    \begin{tabular}{ccc}
        \includegraphics[width=0.32\textwidth]{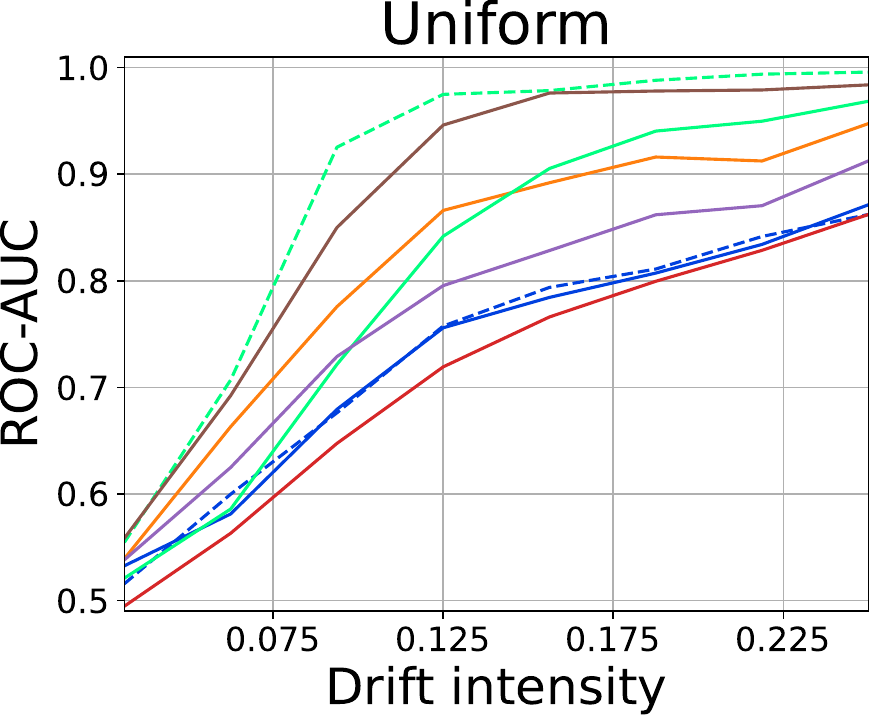} &
        \includegraphics[width=0.32\textwidth]{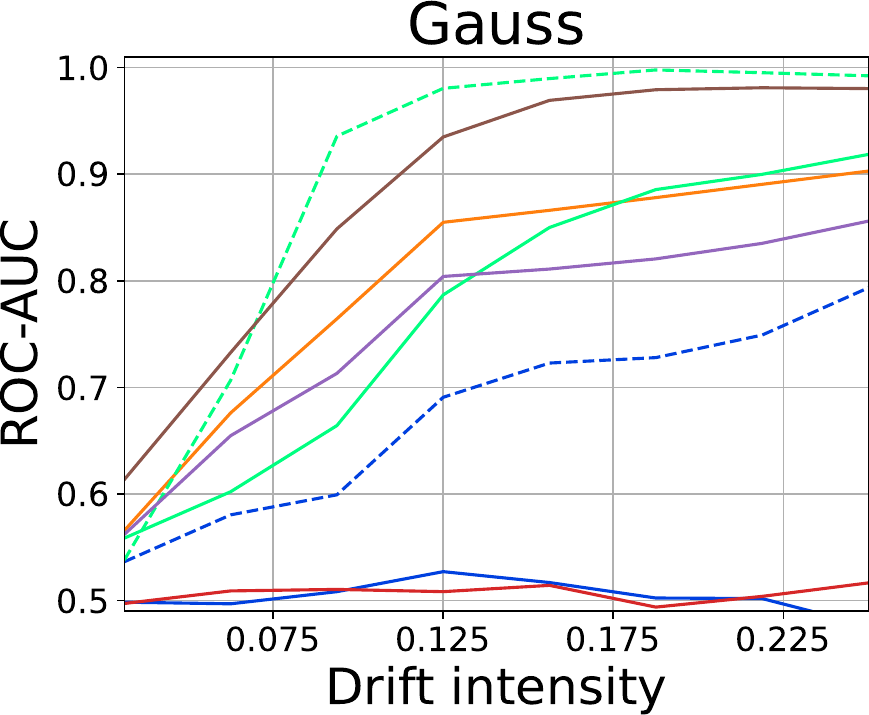} &
        \includegraphics[width=0.32\textwidth]{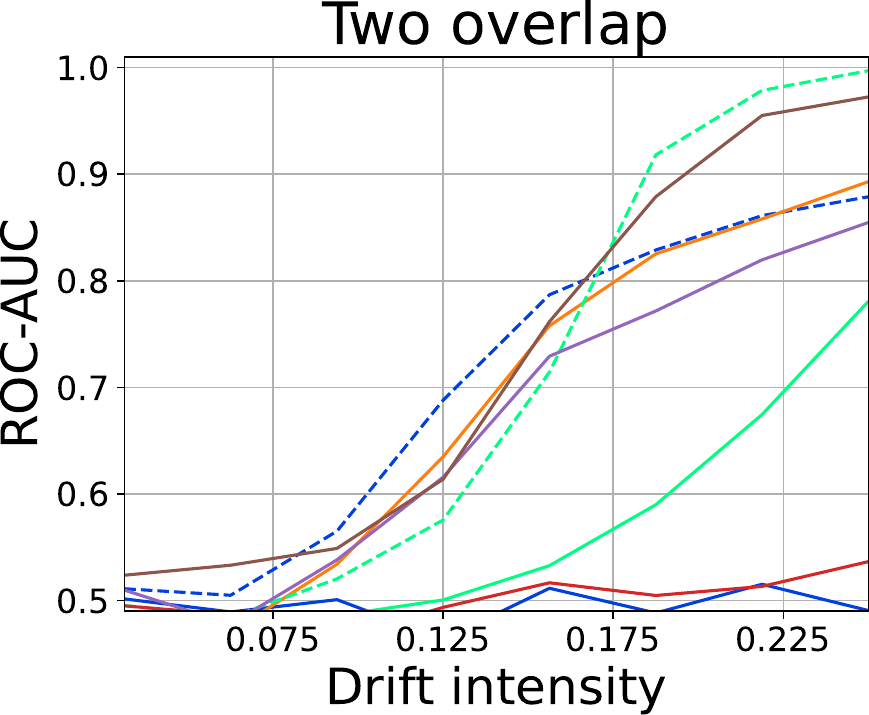} \\
        \multicolumn{3}{c}{\includegraphics[height=3.5em]{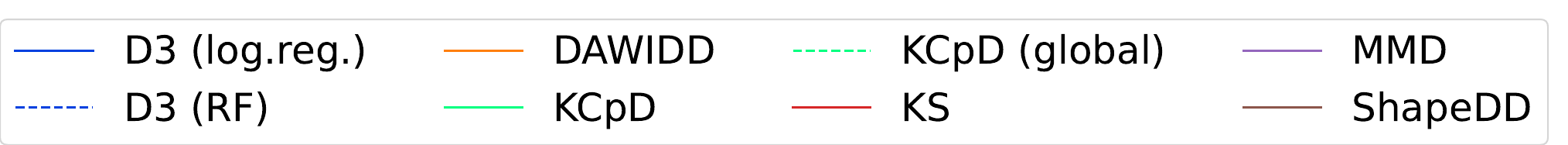}}
    \end{tabular}
    \caption{Drift Detection performance for various intensities.}
    \label{fig:intensity}
\end{figure}

Our results are visualized in Fig \ref{fig:intensity}. As expected, all methods improve their detection capabilities with increasing drift strengths. However, for ShapeDD we observe a stronger increase, reaching a high detection accuracy already for small drift strengths. As ShapeDD makes use of the same test as MMD this implies that the optimized split point choice often results in large improvements. We will study this effect more closely later on. Furthermore, as predicted the performance of D3 heavily depends on the combination of model and dataset. While logistic regression works better than random forest on the uniform dataset, its performance on the two other datasets is essentially random guessing. This is to be expected as linear models cannot learn the necessary non-linear decision boundaries. Furthermore, for 150 samples D3 with random forest even outperforms ShapeDD for all small intensities on the two overlapping squares dataset. This is to be expected as the drift there is contained in the missing corners which is hard to find using the radial basis functions of the RBF kernels, but easy for the axis-aligned decision trees. This is also an example that shows that meta-statistics can decrease the performance when the signal is too weak to be picked up by the preprocessing. Furthermore, we want to point out that DAWIDD is in second place for all datasets and the smaller parameters, and at least third place. KCpD in the batch variant requires larger intensity to outperform DAWIDD. The global variant of KCpD outperforms all online algorithms which is not surprising considering the amount of used data, however, it is usually closely matched by ShapeDD.

Thus, we suggest incorporating as much domain knowledge into the choice or construction of the descriptor as possible. Furthermore, we recommend the usage of meta-statistic or block-based methods. 

\paragraph*{Drift in correlating features}
In real-world applications, drift might only be detected if the correlation of the features is analyzed. To evaluate how well different detectors perform in such settings we consider feature correlations in this experiment. As the correlation is realized on the datasets and not their parameterizations, i.e. in the uniform dataset, the drift manifests itself in the individual features while for the other two, it can only be detected in their correlation, we compare the model performance across different experiments.

\begin{figure}
    \centering
    \begin{tabular}{ccc}
        \includegraphics[width=0.32\textwidth]{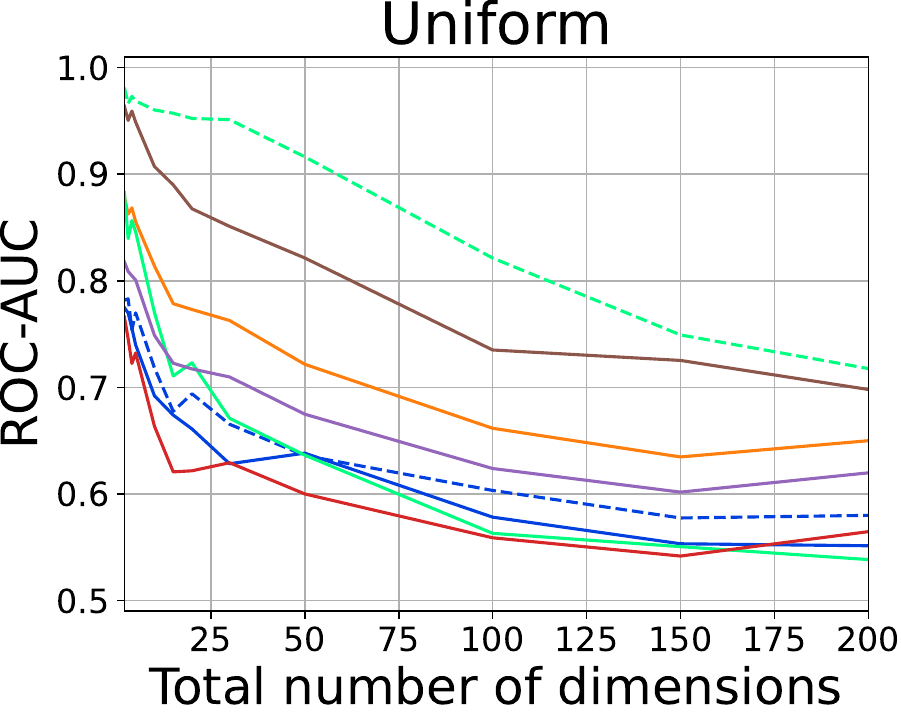} &
        \includegraphics[width=0.32\textwidth]{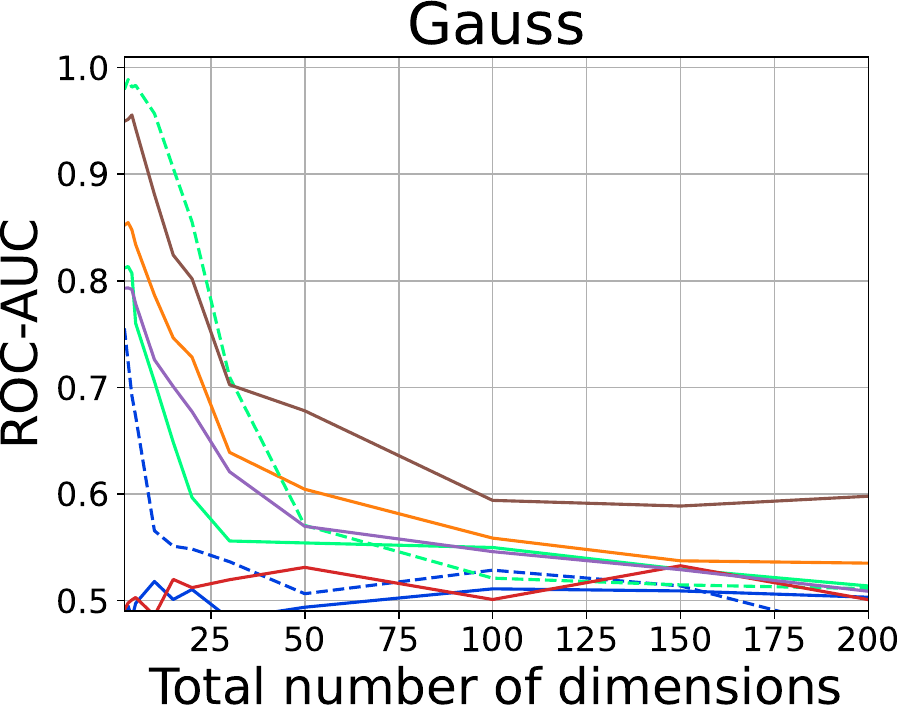} &
        \includegraphics[width=0.32\textwidth]{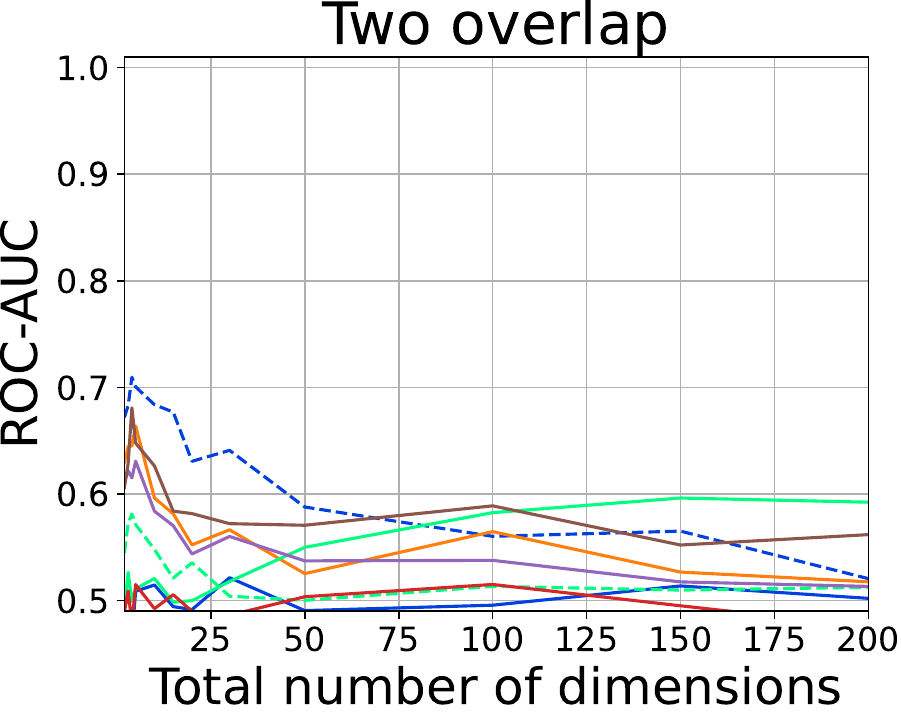} \\
        \multicolumn{3}{c}{\includegraphics[height=3.5em]{src/out/dec/overview_legend.pdf}}
    \end{tabular}
    \caption{Drift Detection performance for various number of total dimensions.}
    \label{fig:dims}
\end{figure}

As can be seen in \Figname~\ref{fig:intensity} (Gauss, Two overlap) KS shows a performance close to random chance. This is to be expected as in both cases the drift is entirely encoded in the feature correlation and thus lost to a feature-wise analysis as performed by KS. The axis-aligned decision trees used by D3 with random forests show a similar issue in the case of \Figname~\ref{fig:dims} (Gauss) whereas the kernel-based methods show far more comparable results across all datasets. 

We thus advise only using methods that make heavy use of feature-wise analysis if drift in the correlations only is either less relevant or very unlikely. If this is not an option, ensemble-based drift detectors may provide an appropriate solution.

\paragraph*{High dimensional data streams}
In many practical applications, one has to cope with high dimensional data streams containing measurements from many different sensors. In these settings, the drift might be reflected in a few features only. We will investigate, whether a high dimensionality disturbs some of the drift detection strategies. For this purpose, we again run experiments increasing the number of non-drifting features.

Our results are visualized in \Figname~\ref{fig:dims}. We observe a decline in the performance of all methods with an increasing number of non-drifting features. However, the KS-based approach quickly declines to a non-acceptable level. Interestingly, D3 also suffers a comparably fast decline in performance although tree-based methods usually perform intrinsic feature selection. However, the decline is usually less steep compared to the other methods. Additionally, in case of the Gauss dataset (in \Figname~\ref{fig:dims}), where the features are strongly correlated and hard to find the random forests we see a particularly steep decline. We also observe that the additional information provided to the global KCpD does not seem to help the method, which can be explained by the choice of the RBF kernel. 

Thus, we advise to choose appropriate preprocessing techniques or descriptors in order to select or construct suited features.

\begin{figure}
    \centering
    \begin{tabular}{ccc}
        \includegraphics[width=0.32\textwidth]{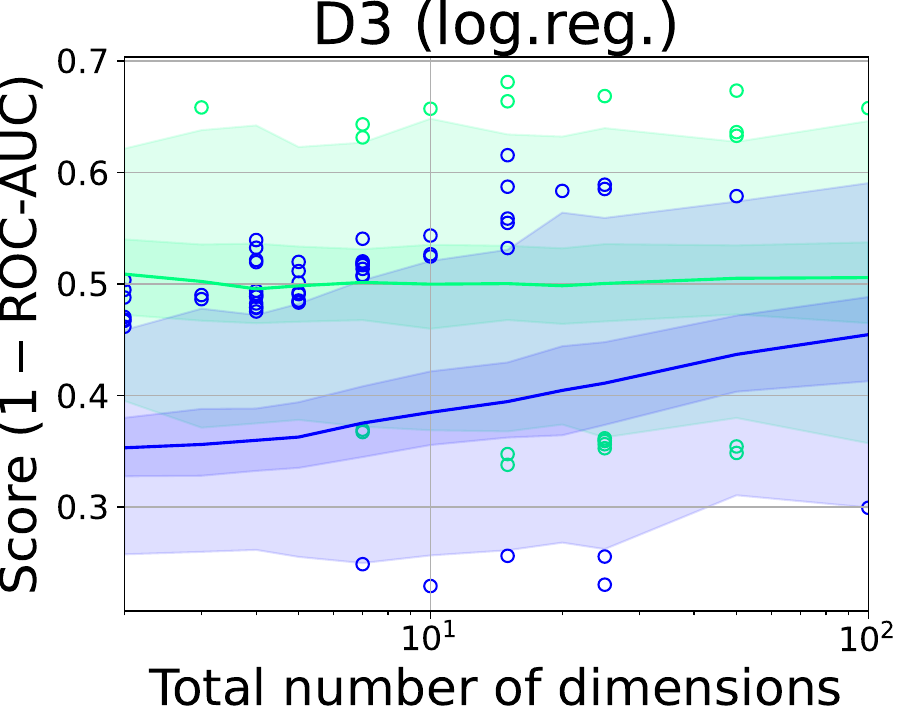} &
        \includegraphics[width=0.32\textwidth]{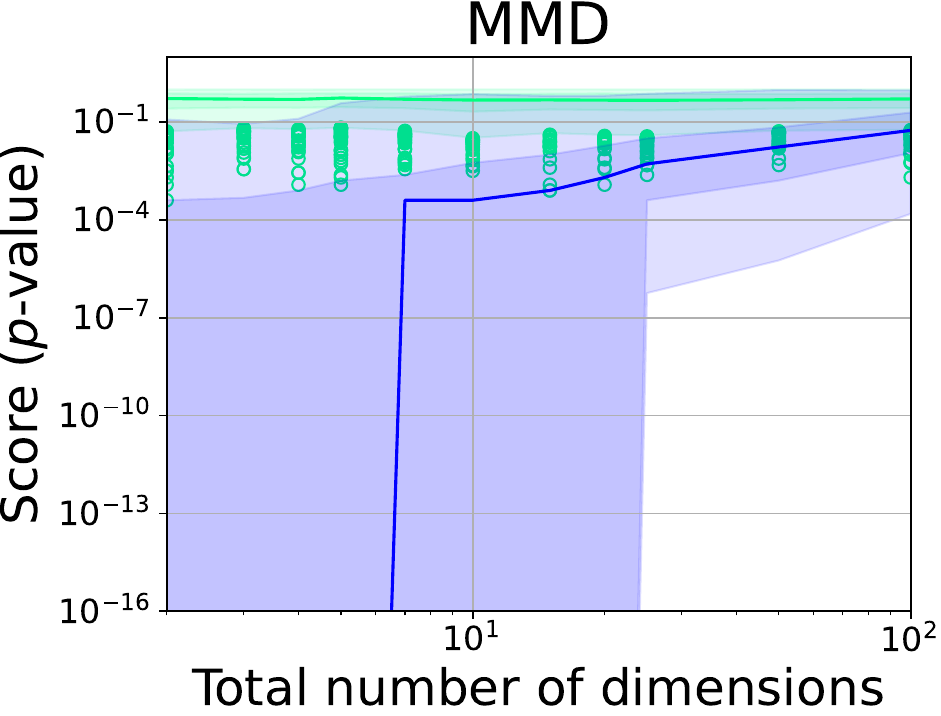} &
        \includegraphics[width=0.32\textwidth]{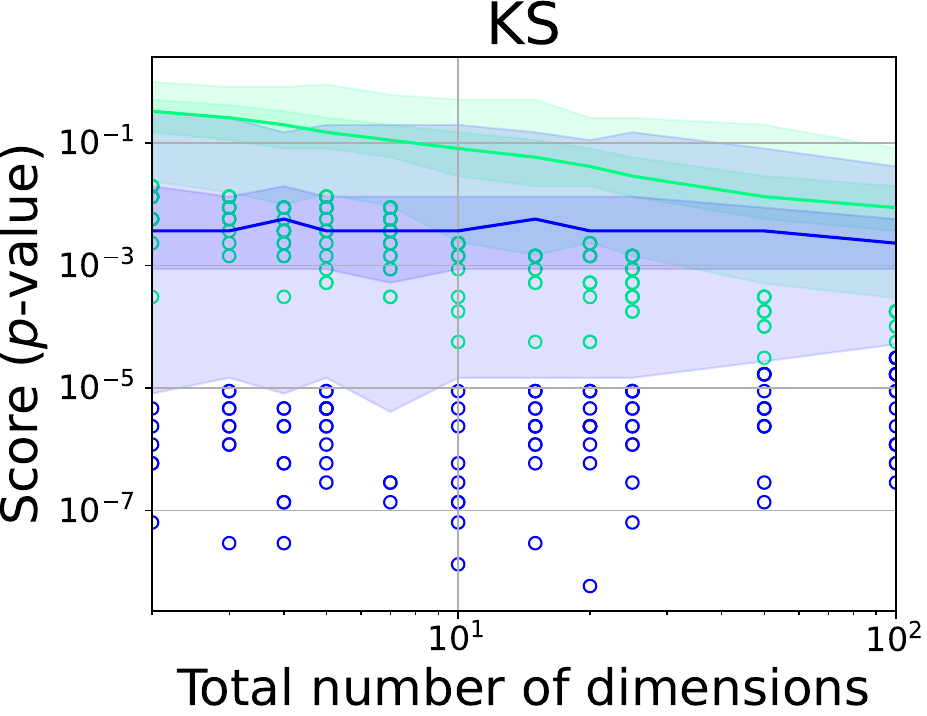} \\
        \multicolumn{3}{c}{\includegraphics[height=1.75em]{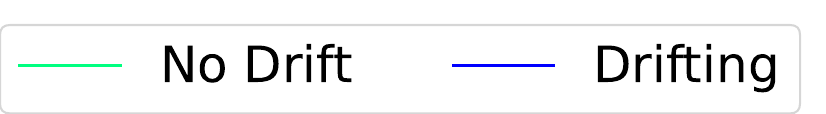}}
    \end{tabular}
    \caption{Effect of total number of dimensions on drift score for various drift detectors. Graphic shows median (line), $25\%-75\%$-quantile (inner area), $\min-\max$-quantile (outer area), and outliers (circles).}
    \label{fig:comp_dims}
\end{figure}

We further analyzed the behavior of the different methods in a 	test bed setting. We used the uniform dataset only and drift right in the middle, which we also used as split point. We then compared the obtained drift scores ($p$-values) for drifting and non-drifting samples, each containing 250 data points. We repeated the procedure 500 times for various numbers of dimensions. The results are shown in \Figname~\ref{fig:comp_dims}. 

As can be seen, for D3 and MMD detecting the drift in the drifting sample becomes harder for more dimensions, the score assigned to the non-drifting stream however stays the same. This is not too surprising as the signal-to-noise ratio becomes worse for more dimensions. 

For KS it is the other way around: while the score for the drifting sample stayed the same, the score for the non-drifting sample declined, indicating that it becomes more like a drifting sample. This effect is due to the fact that just by random chance, some dimensions look like there could be drift although there is none. This effect is essentially the multi-testing problem posing a huge problem for dimension-wise drift detectors.

We thus advise that in case of high dimensions with high cost in case of false alarms, one should refuse  drift detectors that operate dimension-wise.

\paragraph*{Number of drift events}

\begin{figure}
    \centering
    \begin{tabular}{ccc}
        \includegraphics[width=0.32\textwidth]{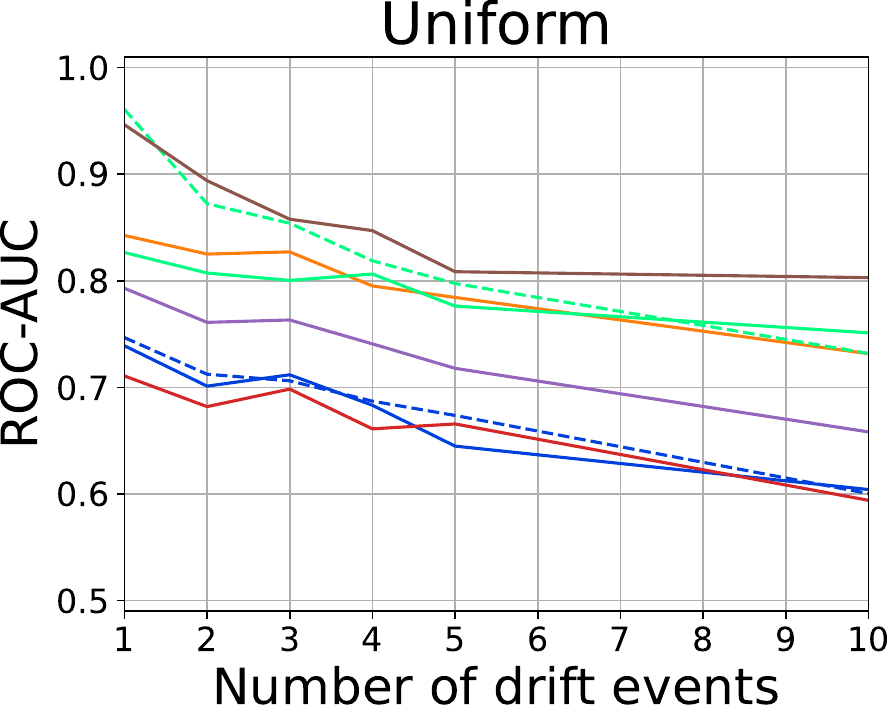} &
        \includegraphics[width=0.32\textwidth]{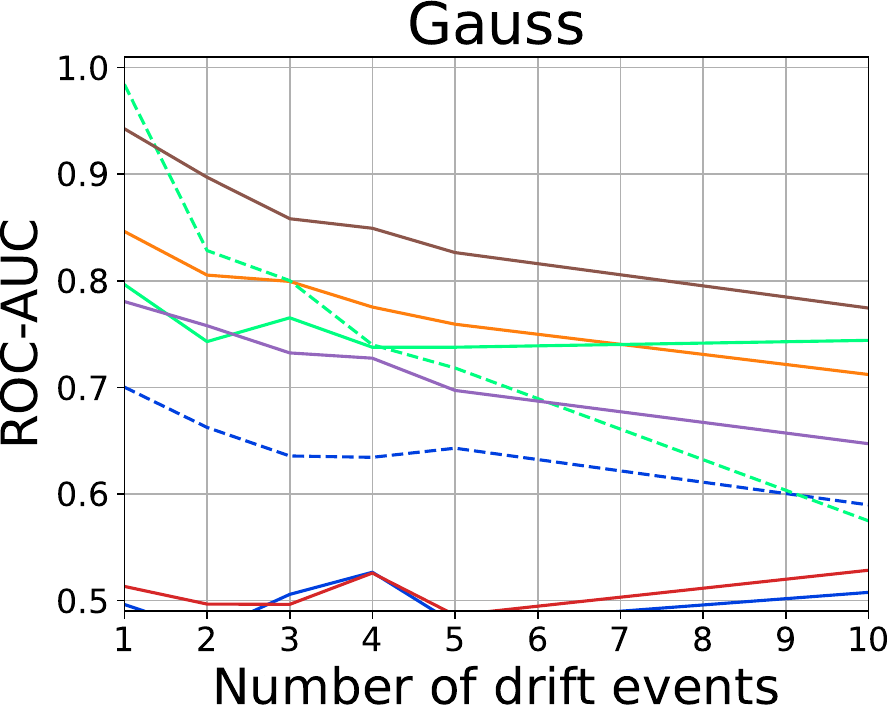} &
        \includegraphics[width=0.32\textwidth]{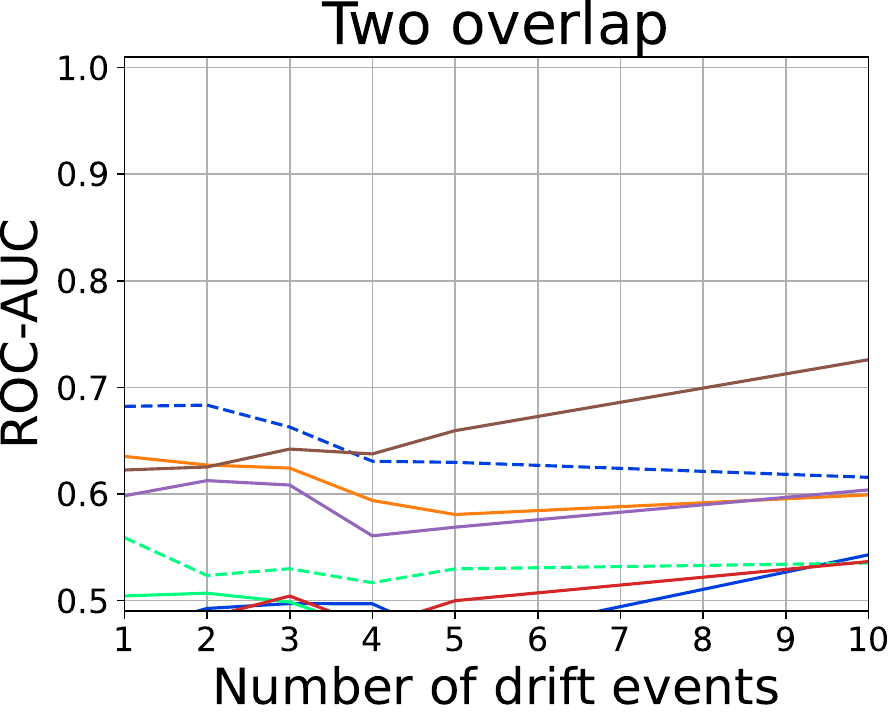} \\
        \multicolumn{3}{c}{\includegraphics[height=3.5em]{src/out/dec/overview_legend.pdf}}
    \end{tabular}
    \caption{Drift Detection performance for various numbers of drift events.}
    \label{fig:number}
\end{figure}

Another crucial parameter is the number of drift events in a single window. Most drift detectors assume that there is only one drift event per window. While that might not be such a big issue for two-window approaches~\citep{shape}, meta-statistic-based algorithms need to take the resulting interference patterns into account and accommodate for them adequately. The ShapeDD for example makes the explicit assumption that there is at most one drift event in the scanner window. However, in particular, streams that alternate between two states may pose a problem to two-window-based drift detectors. For example, if we observe concepts ABA and split right in the middle of B then the mean distribution in both windows is the same and the drift detector will fail to detect the drift. Although overlapping windows reduce the chance of such scenarios, they can still cause some problems.  

As before, we use the same experimental setup, this time increasing the total number of drift events. The results are presented in \Figname~\ref{fig:number}. As can be seen, all drift detectors struggle when confronted with multiple drifts. 
Interestingly, in particular, the global KCpD is strongly affected by the number of drifts. 

We thus advise making use of block-based drift detectors if several drift events are to be expected. In particular, we suggest not to make use of meta-statistic-based methods unless they can explicitly deal with the setup. 

\paragraph*{D3 Model choices}

\begin{figure}
    \centering
    \begin{tabular}{ccc}
        \includegraphics[width=0.32\textwidth]{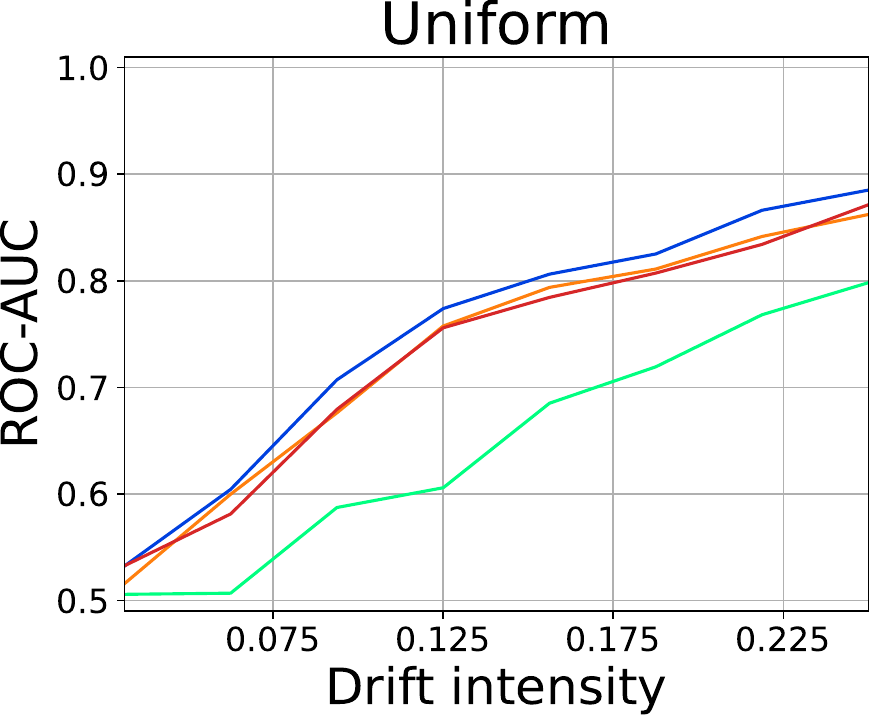} &
        \includegraphics[width=0.32\textwidth]{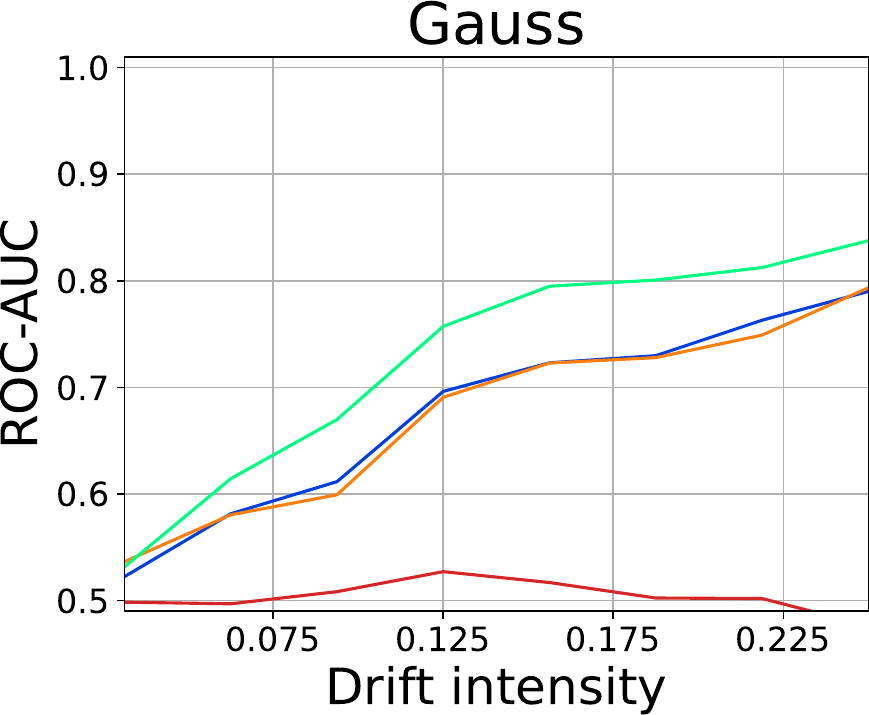} &
        \includegraphics[width=0.32\textwidth]{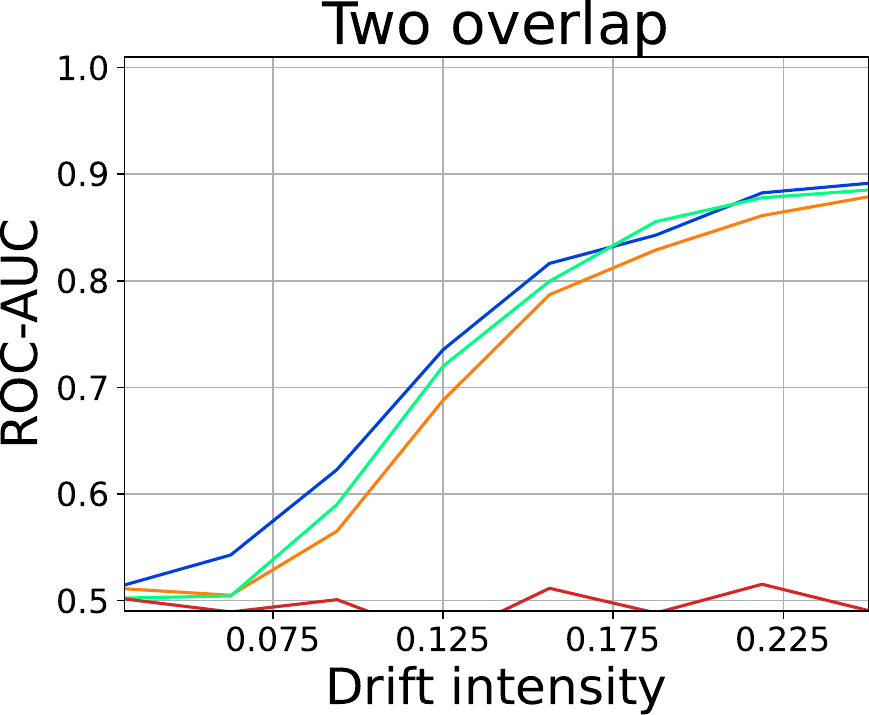} \\
        \includegraphics[width=0.32\textwidth]{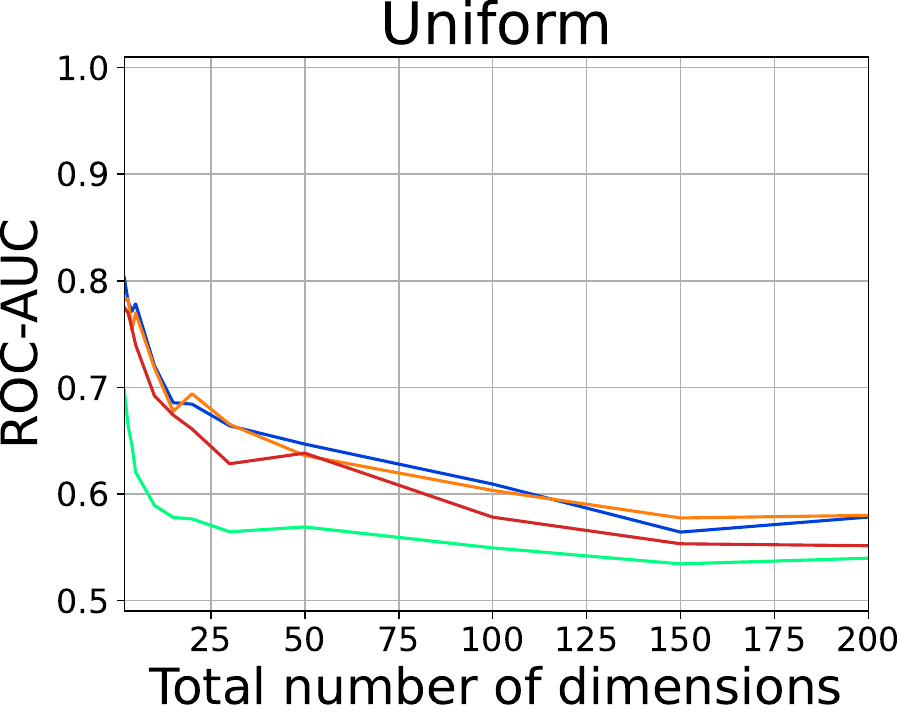} &
        \includegraphics[width=0.32\textwidth]{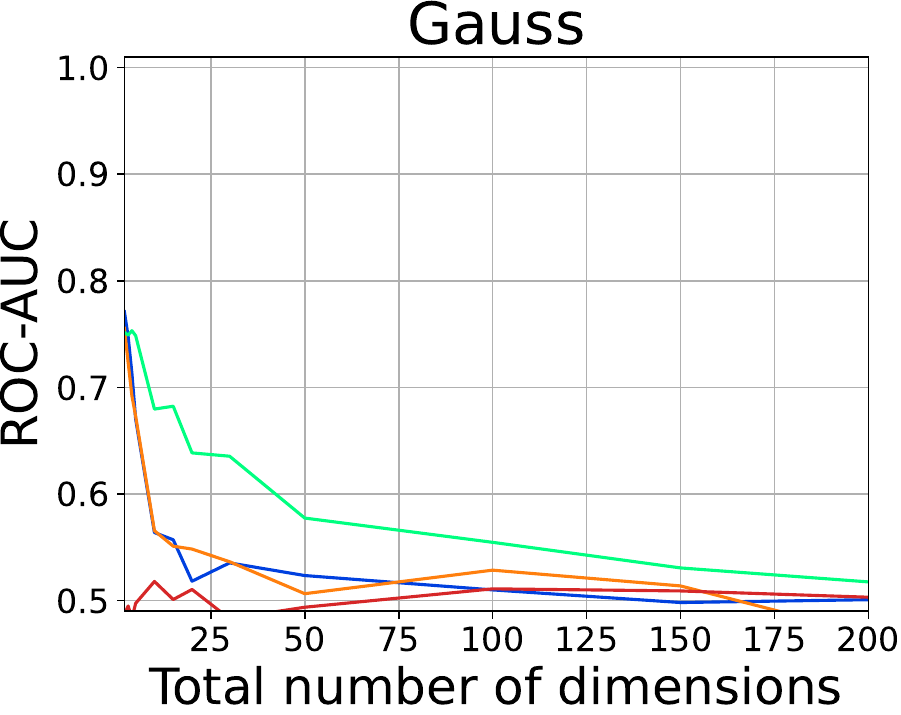} &
        \includegraphics[width=0.32\textwidth]{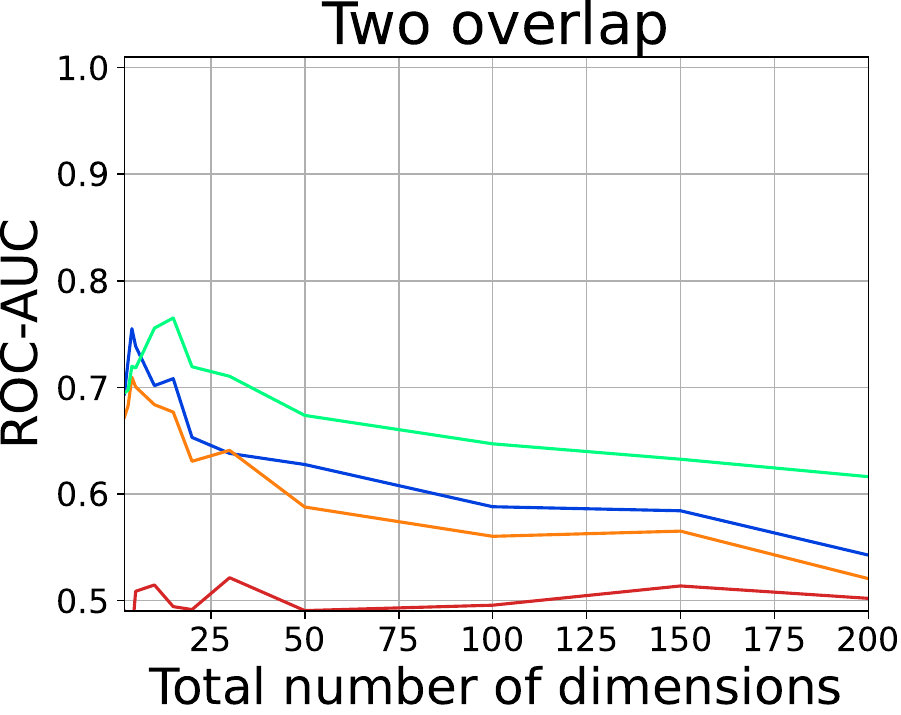} \\
        \includegraphics[width=0.32\textwidth]{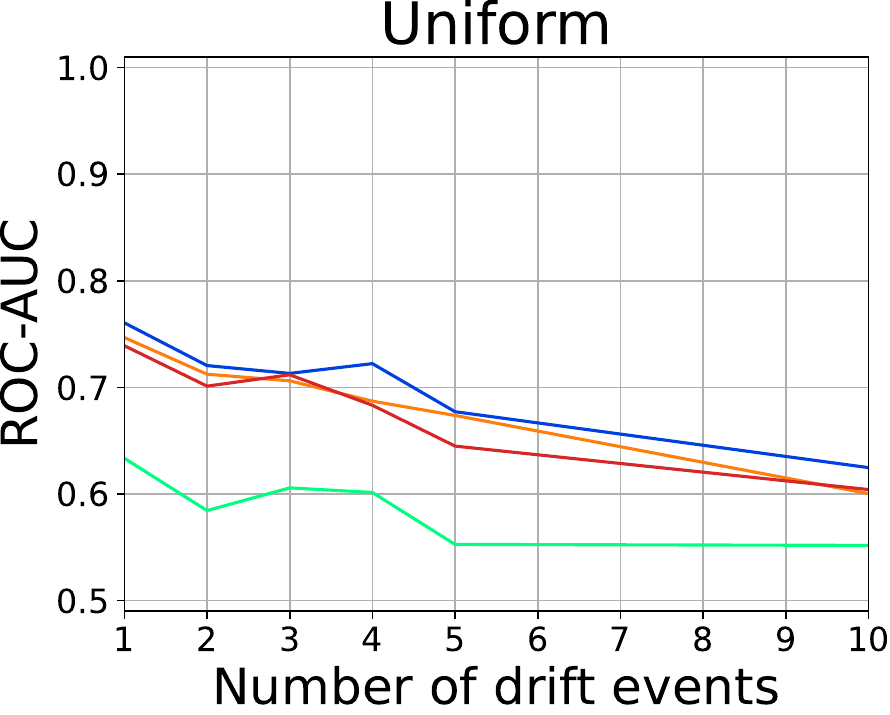} &
        \includegraphics[width=0.32\textwidth]{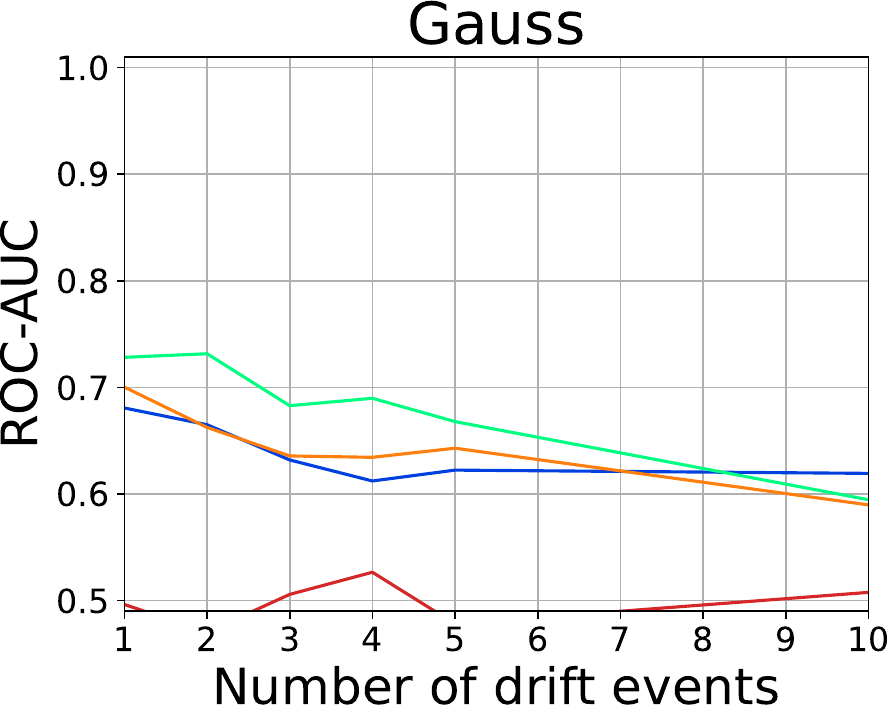} &
        \includegraphics[width=0.32\textwidth]{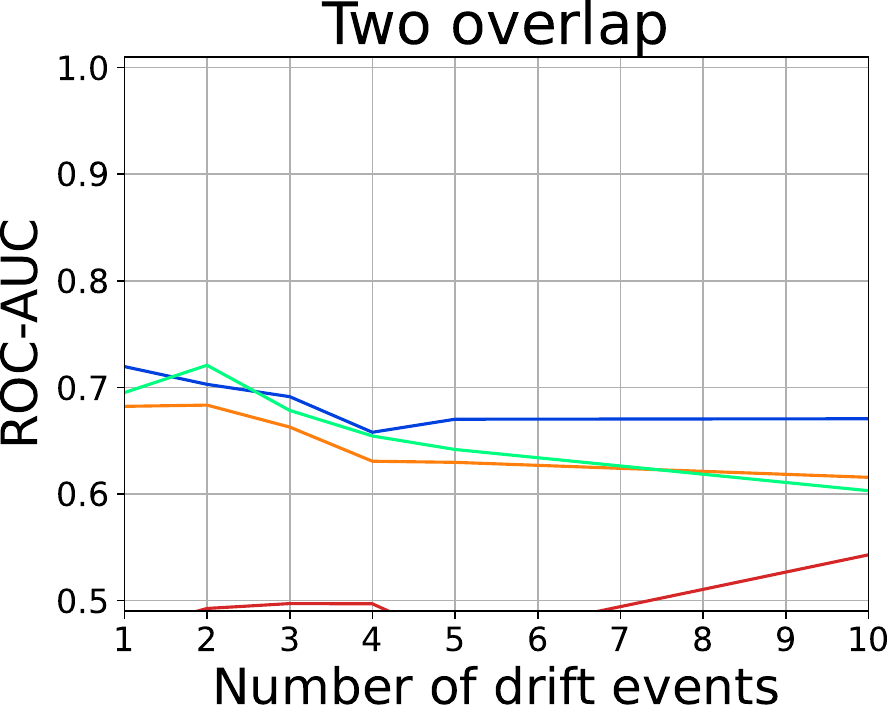} \\
        \multicolumn{3}{c}{\includegraphics[height=1.75em]{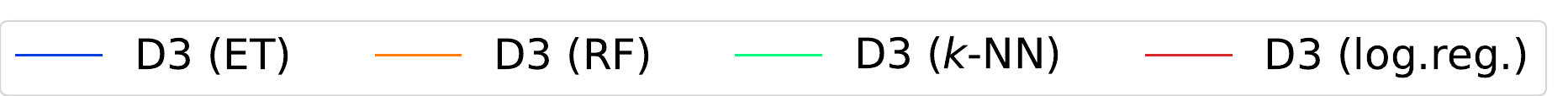}}
    \end{tabular}
    \caption{Drift Detection performance for various models used by D3.}
    \label{fig:model-impact}
\end{figure}

As already pointed out in the experiment on intensity, different specific instantiations of the algorithm may have a significant impact on the detection performance. We showcase this fact in the case of D3 by considering different learning models. We made use of the same experimental setup as above and used the following models: logistic regression, random forests (RF), extra tree forests (ET) (a variant of random forests), and $k$-nearest neighbor classifier ($k$-NN). The results are presented in \Figname~\ref{fig:model-impact}.

As can be seen the model has a major influence on the method's performance. While the $ k$-NN-based model is outperformed by all other models on the uniform dataset, it clearly outperforms all other models on the Gauss dataset. On the other hand, similar models result in similar performance as can be seen in cases of RF and ET.
What surprised us was that we were not able to observe their capability to select the most important features that is theoretically available to all models except $k$-NN, which still outperformed the other models in 2 out of 3 cases with high dimensionality. 

To conclude, we suggest using as much prior knowledge as possible to build a suitable descriptor for the data at hand. This result matches the observations of \citep{ida2022} where the authors argued that the descriptor is actually more important than the metric that is derived from it.  

\paragraph*{Effect of split point}

\begin{figure}
    \centering
    \begin{tabular}{ccc}
        \includegraphics[width=0.32\textwidth]{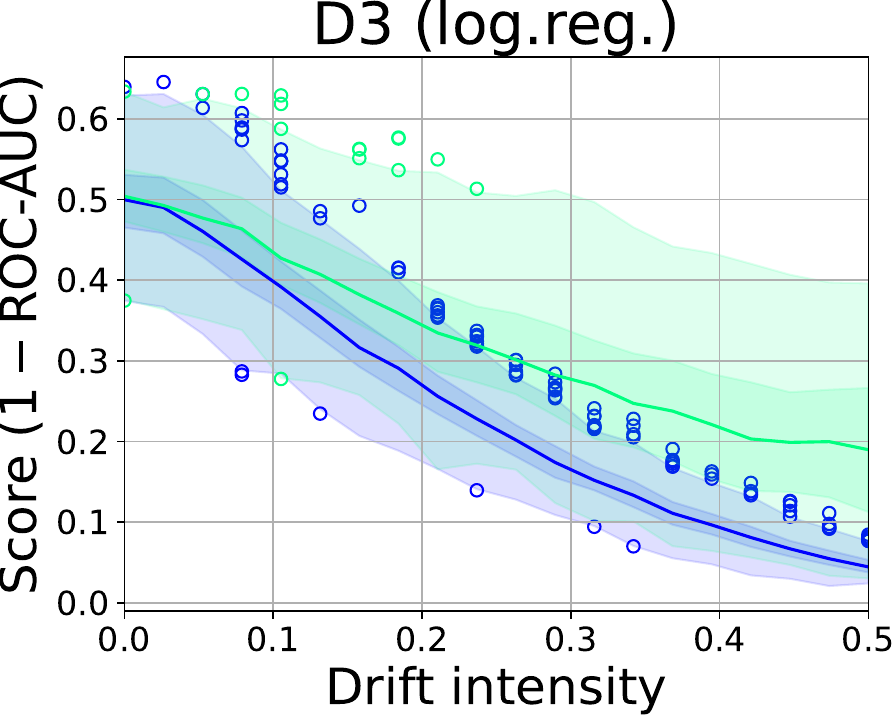} &
        \includegraphics[width=0.32\textwidth]{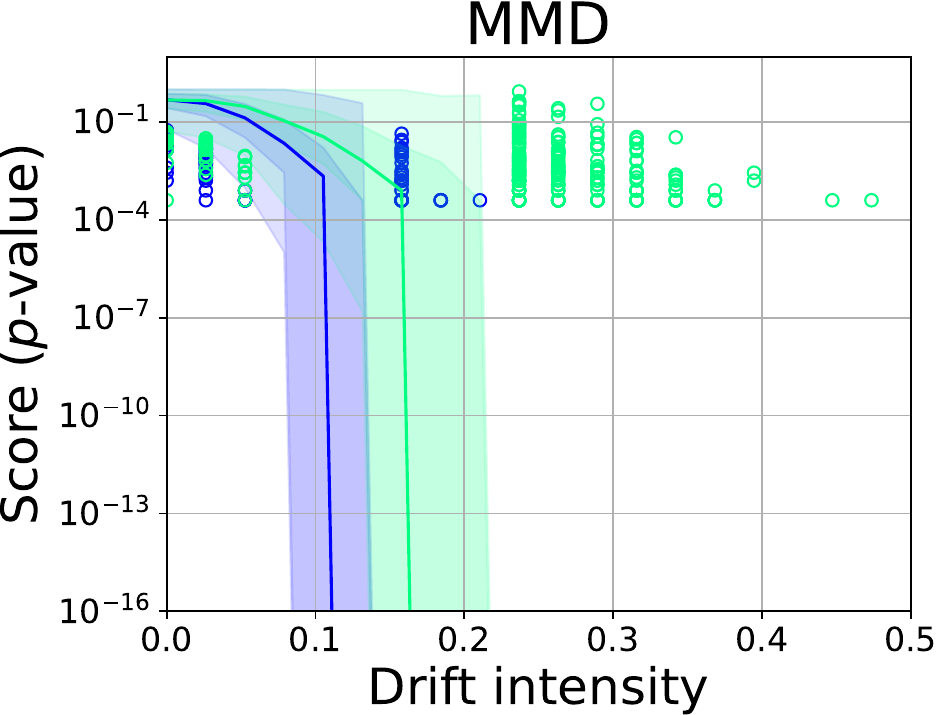} &
        \includegraphics[width=0.32\textwidth]{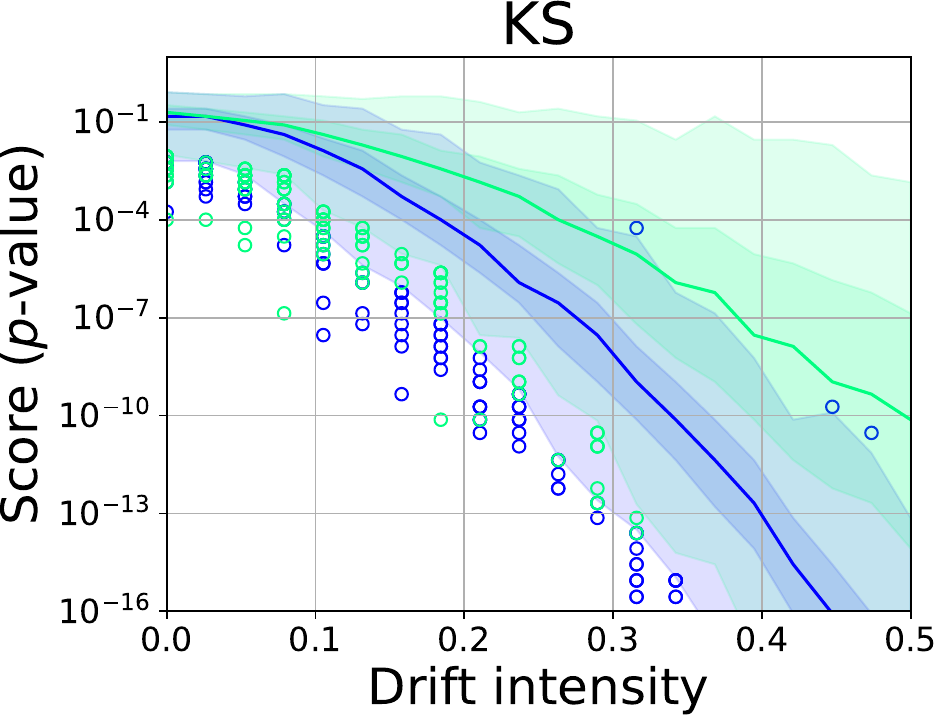} \\
        \multicolumn{3}{c}{\includegraphics[height=1.75em]{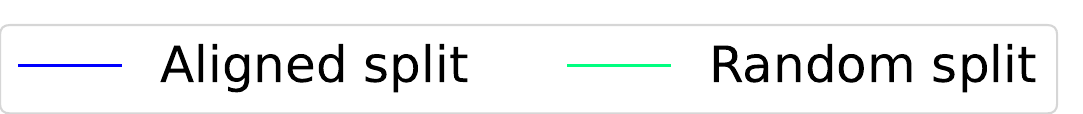}}
    \end{tabular}
    \caption{Effect of split point on detection performance of various drift intensities. Graphic shows median (line), $25\%-75\%$-quantile (inner area), $\min-\max$-quantile (outer area), and outliers (circles).}
    \label{fig:comp_split}
\end{figure}

Last but not least we further analyzed the effect of the used split point for the two-window approaches. We used the uniform dataset only to create two types of samples each containing 250 samples. In the first type, the drift occurs right in the middle, which is also used as the split point for the methods. For the second, the drift happens at a randomly chosen point within the sample. We compare the obtained scores ($p$-values) for various drift intensities, the results are shown in \Figname~\ref{fig:comp_split}.

As can be seen, all three methods perform significantly better when the drift point is known. Also, they show drastically smaller variance. The latter is to be expected as depending on where the drift point is relative to the used split point and the window size the problem can become simpler or harder. 

Considering our results, it is reasonable that approaches that preselect a candidate split point rather than simply choosing one ignorant of the underlying data provide a valid strategy to optimize performance. This reflects the discrepancy of performance when comparing MMD with ShapeDD, with the latter being essentially a strategy to find good candidate split points. We thus advise the user to investigate options to preselect a good candidate split point, either through prior knowledge or by choosing an appropriate algorithm. 

\paragraph*{Loss-Based approaches}

\begin{figure}
    \centering
    \begin{tabular}{ccc}
        \includegraphics[width=0.32\textwidth]{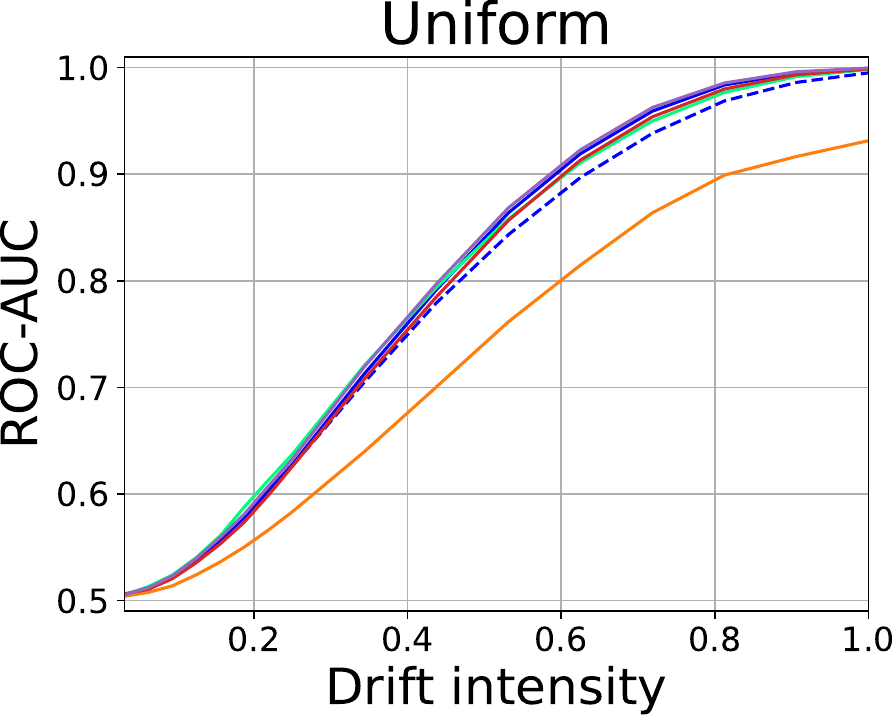} &
        \includegraphics[width=0.32\textwidth]{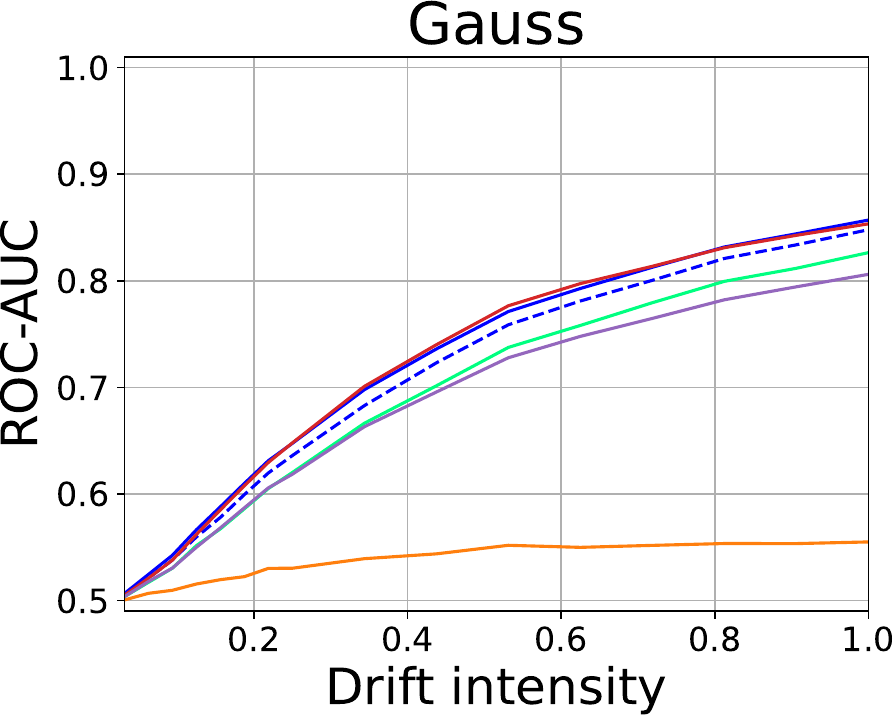} &
        \includegraphics[width=0.32\textwidth]{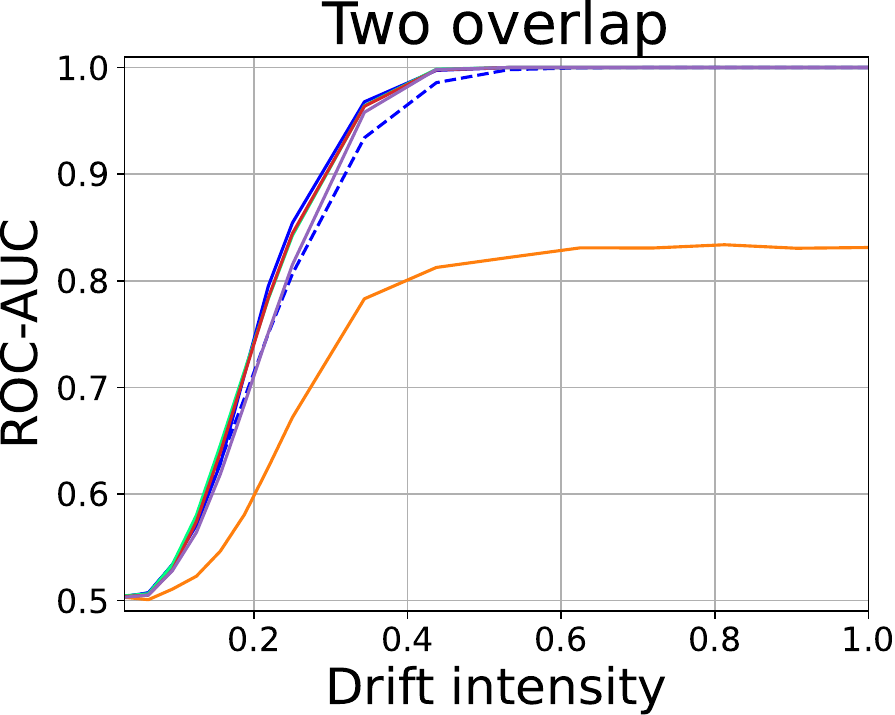} \\
        \includegraphics[width=0.32\textwidth]{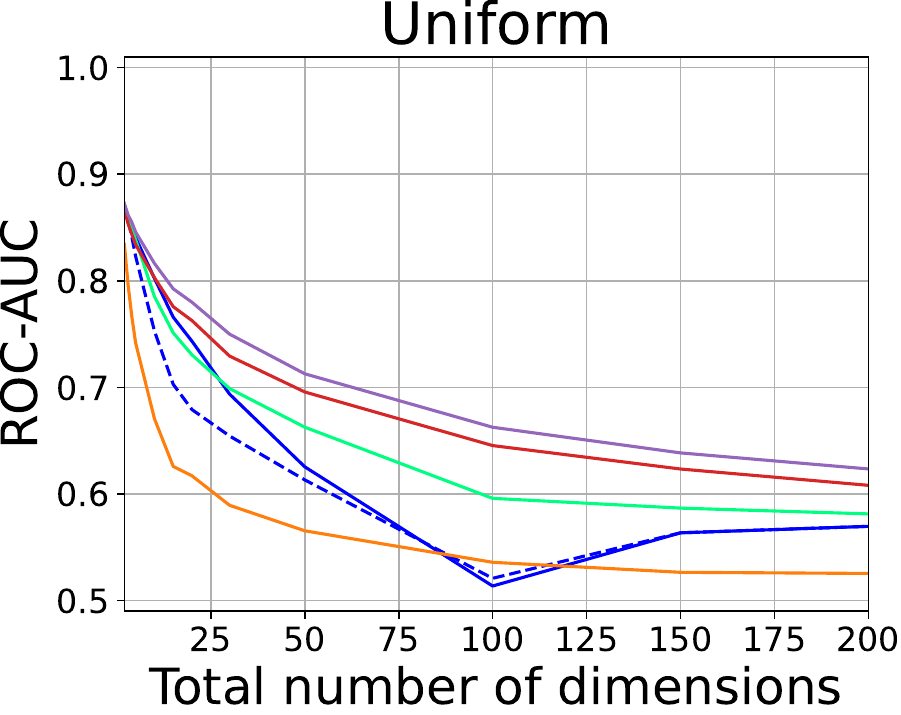} &
        \includegraphics[width=0.32\textwidth]{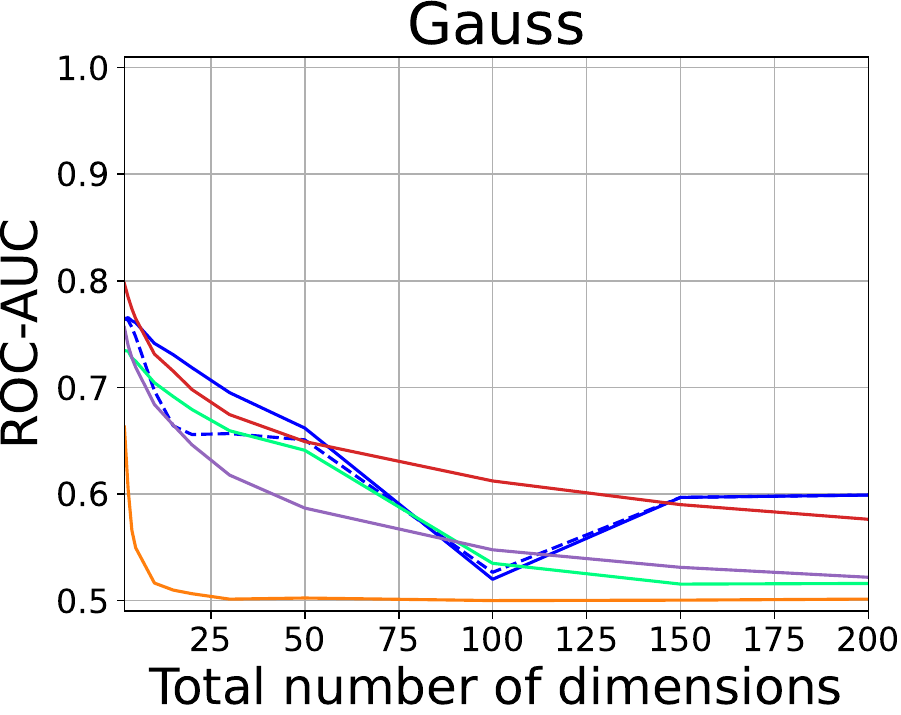} &
        \includegraphics[width=0.32\textwidth]{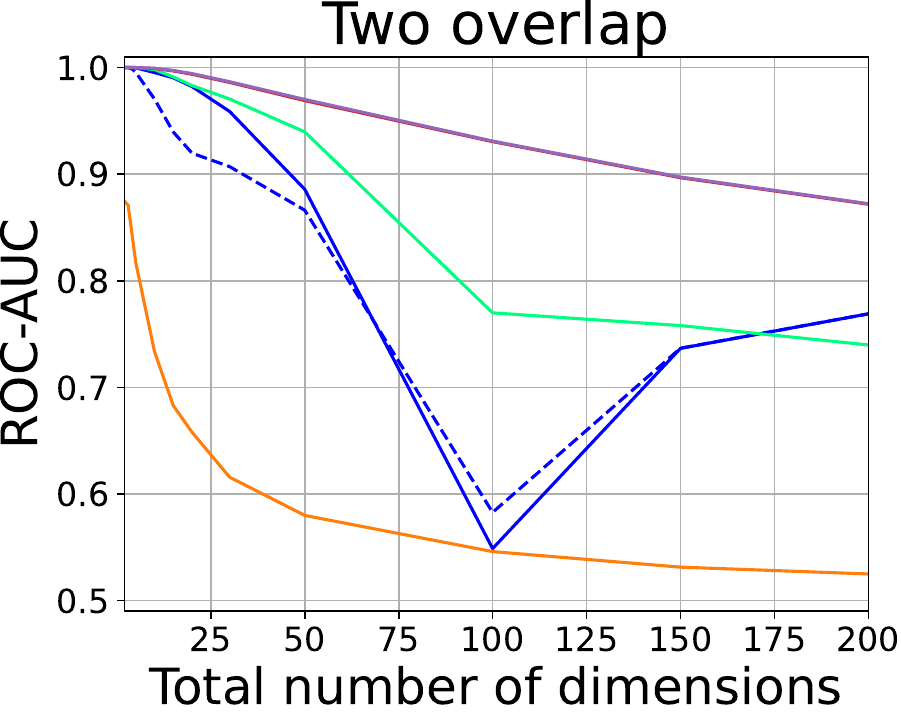} \\
        \multicolumn{3}{c}{\includegraphics[height=1.75em]{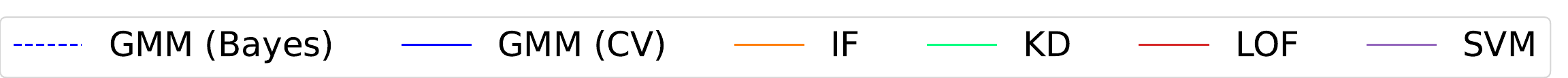}}
    \end{tabular}
    \caption{Drift Detection performance for various model/loss-based approaches. Experiments with varying intensity use a different scale than previous experiments.}
    \label{fig:model-loss-based}
\end{figure}

We also considered loss- and outlier-based approaches, namely one-class-SVM (SVM; RBF kernel), local outlier factor (LOF; $k = 10$), isolation forests (IF), and density-based approaches kernel-density (KD; RBF kernel) estimate, and Gaussian mixture model (GMM; $\leq 10$ components selected via cross-validation (CV) or Dirichlet prior (Bayes)). We made use of the scikit-learn~\citep{scikit-learn} standard implementation and default parameters if not specified above. In case of the outlier-based approaches we used the outlier score, in case of the density-based approaches we used sample probability. We considering the same datasets and parameters as before but used a simplified setup: We first trained the models on the first 100 samples (no drift) and then applied them to the remaining stream. As a consequence the number of drift becomes irrelevant to the detection task. The results are shown in \Figname~\ref{fig:model-loss-based}.

As the stream contained data from the initial training distribution and the drifted distribution, we expect the samples from the other distribution to be marked as outliers or assigned a small probability. However, even for an intensity of 0.25 the ROC-AUC falls below 0.7 for all methods but and two out of three datasets. Notice that the experiments using D3 clearly indicate that a significantly better performance is possible. We thus increased the default value for intensity to 0.5.

As expected we found that an increase of intensity leads to better performance. Still most methods require a comparably strong drift to achieve good performance. In particular, isolation forests seem to perform comparably poor. The number of dimensions also has a negative impact on the performance, resulting in lower scores as compared to the approaches studied in the previous sections.

We thus found additional empirical evidence for the results of \cite{icpram} which challenge the suitability of loss-based approaches for drift detection from a formal mathematical perspective and therefore suggest the reader not to make use of loss-based approaches.

\subsection{Conclusion and Guidelines for Drift Detection}
After providing a formal definition of a drift detector, we presented and categorized many different approaches for drift detection. Table~\ref{tab:summary} and \Figname~\ref{fig:driftdetction-tree} provide a condensed summary of the proposed taxonomy and summarize how different methods are implemented according the the common staged scheme as visualized in \Figname~\ref{fig:dd-stages}. To wrap up our analysis on drift detection, we will summarize the main findings of our experiments with representative implementations of the discussed strategies.

A main finding is that as much domain knowledge as possible should be incorporated when designing drift detection schemes. This concerns selecting appropriate preprocessing techniques, constructing and engineering suitable features, and choosing fitting descriptors in stage 1\&2 of the process. Over all experiments,  we found that it is advisable to use meta or block-based methods. Next to the finding that choosing good split points is crucial for obtaining good detection capabilities, we found some very situation-specific insights: A feature-wise analysis should only be performed if it is expected that the drift does not inflict itself in correlations. Otherwise, relying on ensemble-based techniques seems to be the better solution. When working with high dimensional data, one should avoid using dimension-wise methodologies, especially if false alarms are costly in the considered application. Finally, if multiple drifts are expected, applying block-based detectors is particularly suitable. 
Finally, but maybe most importantly loss-based strategies should be avoided when the target of the drift detection is monitoring for anomalous behavior.
  
\section{Drift Localization and Segmentation\label{sec:drift-localization}}
Solely detecting and determining the time point of the drift is not sufficient in many monitoring settings. In order to take appropriate action, more questions concerning the drift have to be answered. In this section, we focus on the \emph{where} -- our goal is to identify the drifting components in space.

\subsection{Problem Setup and Challenges\label{sec:drift-localization-def}}

\begin{figure}
    \centering
    \begin{tabular}{cc}
        \includegraphics[width=0.32\textwidth]{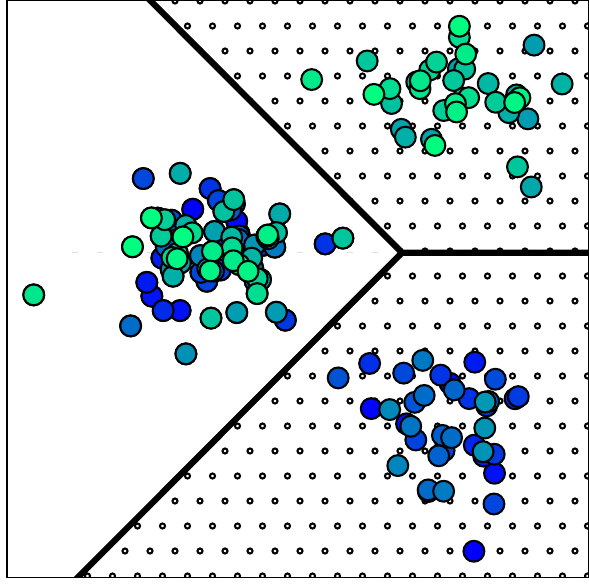} &
        \includegraphics[width=0.32\textwidth]{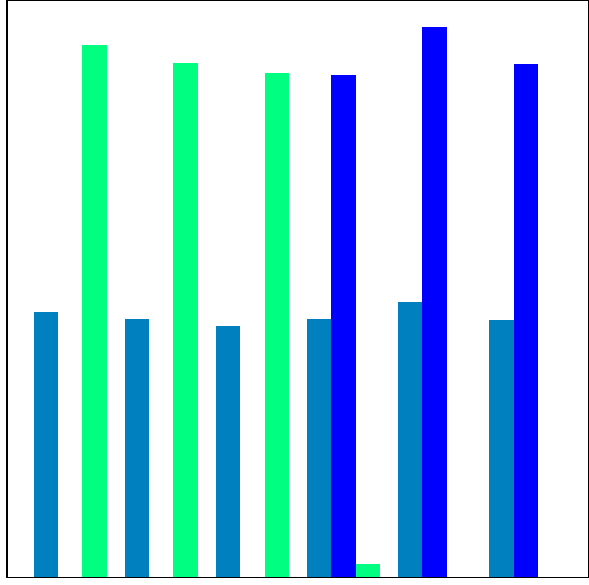} \\
    \end{tabular}
    \caption{Visualization of distribution consisting of three segments (optimal drift segmentation; borders given by black lines). The two segments on the right form the minimal drift locus (dotted area). Left: Sample drawn from distribution color indicates time point of observation, evolving from dark blue to light green, right: Time point distribution per segment. }
    \label{fig:localizaiont-visulazation}
\end{figure}

The task of determining where in data space the detected drift takes place or manifests is referred to as \emph{drift localization}. 
Informally, the problem of drift localization can be expressed as ``finding those regions in the data space that are affected by the drift''~\citep{lu_learning_2018,kdqtree}. We have illustrated this in \Figname~\ref{fig:localizaiont-visulazation}, the dotted area is the area of interest.
A slightly different angle on this question would be investigating
``whether or not a given sample is affected by the drift''~\citep{LDD,localization}. Both questions are relevant in practical applications. If we know which parts of the data space are affected by the drift, we know which analysis has to be redone. On the other hand, if we know which data points are affected by the drift we can update our dataset more efficiently, i.e., we do not need to discard all old data points but only the affected ones. 

Both questions can be raised interchangeably: if we know which parts of the data space are affected by drift, we can mark all data points therein as drifting. If, on the other hand, we know for every data point whether or not it is drifting, we can mark the corresponding parts of the data space as drifting. However, in practical applications, identifying the drifting samples is usually more feasible. In particular, we can consider it as yet another statistical test with the $H_0$ hypothesis ``The data point $x$ is not affected by drift''.

To summarize, we want to separate those parts of the dataset/space where the drift manifests from those irrelevant to the drift. However, this is challenging as the definition of drift is not local in the sense that it makes no statement about the inner workings of the drift process. Rather, it simply states that there is some kind of difference in the distributions over time. Yet, from a mathematical point of view, there is no obvious way to talk about the behavior locally, i.e., at a single point. Thus, before we can work on a solution for the task, we first need to specify what we actually mean. For this purpose, we will make use of the formalization of drift localization presented in \citep{localization}.

When discussing drift in the unsupervised setup one usually imagines something like a Gaussian moving through space, i.e., $\X = \R^d$ and $\D_t = \mathcal{N}(\mu_t, \sigma)$ where $\mu_\cdot : \T \to \R^d$ is the moving mean of the Gaussian. However, as is well known Gaussians span the entire space and are thus not suited for analyzing the local properties of the drift. 

Instead, we suggest to approximate the drift process using a mixture model of uniform distributions on a grid, i.e., denote by $L^{(n)}_{i_1,\dots,i_d} = [i_1/2^n,(i_1+1)/2^n) \times \dots \times [i_d/2^n, (i_d+1)/2^n)$ the grid cell starting at $(i_1,\dots,i_d)$ with length $2^{-n}$ and grid based approximation by $\widehat{\D}_t^{(n)} = \sum_{i_1,\dots,i_d \in \mathbb{Z}} \lambda_{i_1,\dots,i_d}(t) \mathcal{U}(\:L^{(n)}_{i_1,\dots,i_d}\:)$ where $\lambda^{(n)}_{i_1,\dots,i_d}(t) = \D_t(\:L^{(n)}_{i_1,\dots,i_d}\:)$ are the time dependent weights assigning with which probability a grid cell will be present in the data at the considered time. Notice that $\widehat{\D}_t^{(n)}$ is a valid approximation in the sense that it converges weakly to $\D_t$ as $n \to \infty$. On the other hand, $\widehat{\D}_t^{(n)}$ has drift if and only if at least one weight function $\lambda_{i_1,\dots,i_d}(t)$ is not constant. 

Clearly, if $\D_t$ has no drift, then neither does $\widehat{\D}_t^{(n)}$. On the other hand, if $\D_t$ has drift and we choose $n$ sufficiently large then $\widehat{\D}_t^{(n)}$ has drift, too. However, if we choose $n$ comparably small, then small drifts are lost to us, e.g., consider $\X = \R, \T = [0,1]$ and $\D_t = \delta_{t/2^m}$ then $\widehat{\D}_t^{(n)}$ has drift if $n > m$. 

\begin{figure}
    \centering
    \begin{tabular}{ccc}
        \includegraphics[width=0.3\textwidth]{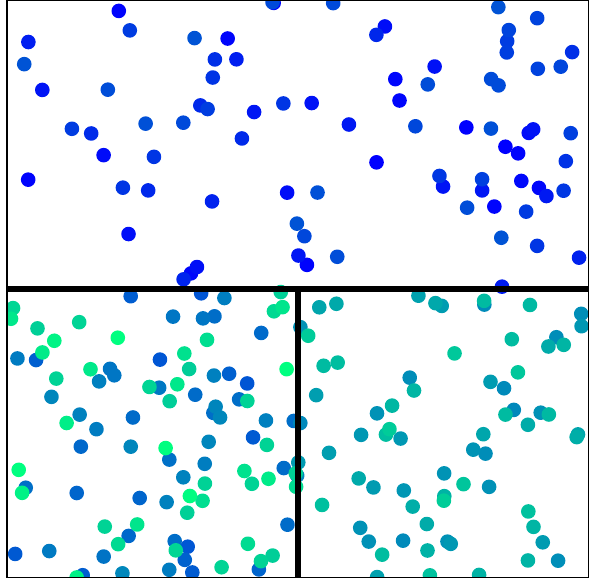} &
        \includegraphics[width=0.3\textwidth]{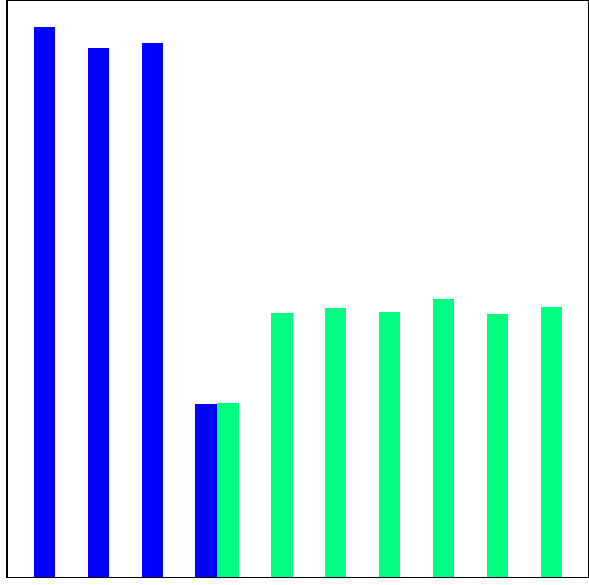} &
        \includegraphics[width=0.3\textwidth]{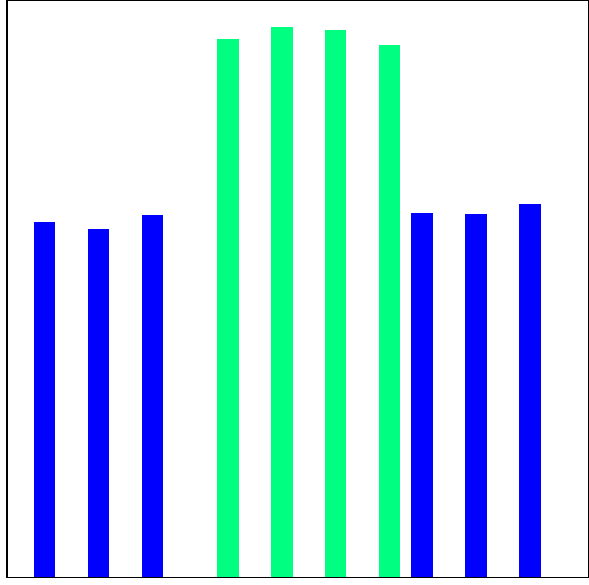}\\
        Segments & Above vs. below & Below only: left vs. right \\
    \end{tabular}
    \caption{Visualization of distribution with three segments, all having drift. Left: optimal drift segmentation, time is color coded, black lines mark segment border, middle/right: Time point distribution of segments. First drift moves distribution from upper halve to left below, second from left below to right below, third from right below to left below. The upper halve is a drift segment (no drift inside), the lower halve is not as the distribution moves from left to right and back.}
    \label{fig:segmentation-visulazation}
\end{figure}

In a sense we decompose the drift into two kinds: 1) the drift that ``moves'' probability between the cells which becomes apparent in a change of the weights $\lambda^{(n)}_{i_1,\dots,i_d}(t)$, and 2) the drift that is happening inside the cells $L^{(n)}_{i_1,\dots,i_d}$ that is not approximated by $\widehat{\D}_t^{(n)}$ and thus lost in our observation (see \Figname~\ref{fig:segmentation-visulazation} for an example illustration).
Notice that this effect is also used drift detection: virtual classifiers~\citep{kifer2004detecting} try to find a set $L \subset \X$ (corresponds to the class +1) such that the samples in the reference window $W_-(t)$ are more likely to fall into $L$ compared to the samples in the current window $W_+(t)$ the drift detection is then based on measuring the drift that moves the probability between $L$ and $L^C$. In \Figname~\ref{fig:segmentation-visulazation} the classifier may choose $L$ as the upper halve and then observe the discrepancy presented in the middle diagram. 
An interesting case occurs when we can choose $n$ so large that $\D_t$ has no more drift of the second kind, i.e., no drift inside any of the cells $L^{(n)}_{i_1,\dots,i_d}$. In this case, the entire information on the drift is encoded in the weights $\lambda^{(n)}_{i_1,\dots,i_d}(t)$ and even if approximating $\D_t$ locally by uniform distributions is no good fit the approximation errors are irrelevant for the analysis of the drift. This can be interpreted as the fact that the drifting behaviour inside $L^{(n)}_{i_1,\dots,i_d}$ is homogeneous, i.e., which particular point we consider is irrelevant as the temporal distribution is always the same. 

Notice that this idea offers an algorithmic solution for example by employing an quadtree-like scheme using drift detection on each leaf as stopping criterion. However, there is no need to make use of grid cells for the approximation. The important idea is that if we can partition the data space into segments $L_1,\dots,L_n$ such that there is no drift inside each of the segments, the drift is entirely encoded in the way the weight functions $t \mapsto \D_t(L_i)$ deviate from constants. 
We will refer to sets $L \subset \X$ for which there is no drift inside as \emph{drift segments}. 
In order to give a formal definition thereof we first need to understand what we mean by ``there is no drift inside $L$''.

Intuitively, describing the drift inside a set $L$ we basically mean to perform some kind of spatial restriction of the drift process from $\X$ to $L$ and then apply the usual definition. As pointed out in the beginning, the problem that there is no way to describe $\D_t$ point-wise arises. Thus, we cannot simply define the restriction to be equal to $\D_t$ on $L$ and 0 everywhere else as one would do with a function.
We thus need to come up with a formal definition of what restriction means. We would expect it to fulfill at least the following properties:
\begin{enumerate}
    \item It should again be a drift process on $\X$.
    \item It should be concentrated on the set we are restricting on.
    \item It should resemble the original drift process on the set we are restriction on.
\end{enumerate}

Let us construct such a process. In order to fulfill the first and third property, let us start with the drift process $(\D_t,P_T)$. This clearly is a drift process on $\X$ that resembles $(\D_t,P_T)$. The second property can be interpreted as: every set $A \subset \X \setminus L$ has probability 0. To realize this we can simply ``remove'' $L^C$ from every set before taking measure, i.e., we consider $\D_t(A \cap L)$. This however is no longer a measure on $\X$ as $\D_t(\X \cap L) = \D_t(L) < 1$. To cope with that we normalize by $\D_t(L)$, i.e., we consider $\D_t(A \cap L) / \D_t(L) = \D_t(A \mid L)$ which is just the conditional probability given $L$. This is only defined if $\D_t(L) > 0$ which does not need to be the case for all $t$. We can interpret this as ``at time $t$ we do not make any observations in $L$'' so it does not make any sense to talk about the restriction to $L$ here. We thus also modify $P_T$ to assign the value 0 to all such time points. This can be done in two ways: we can consider the \emph{active time} of $L$, i.e., the set $\T_L = \{t \mid \D_t(L) > 0\}$, and then consider $P_T(W \mid \T_L)$ using similar arguments as above or we consider the probability that we observe a sample in $L$ during $W$ which leads to $\D(W \times L \mid \T \times L)$. Here the latter choice is more natural form a modeling perspective as the holistic distribution of the drift process $(\D_t(\cdot \mid L), \D(\cdot \times L \mid \T \times L))$ is given by $\D( \cdot \mid \T \times L)$ which is well defined if $\D_\T(L) > 0$ and we can use the other arguments as before. Furthermore, notice that if $\D_t$ has a density $f$ with respect to some measure $\mu$, i.e., $\D_t(A) = \int_A f(x) \d \mu(x)$, then the (scaled and 0 extended) restriction of $f$ onto $L$, i.e., $f_{|L}(x) = f(x) / \int_L f(x) \d \mu(x)$ if $x \in L$ and 0 otherwise, is exactly the density of $\D_t( \cdot \mid L)$ with respect to $\mu$. In particular, this does not depend on the choice of $\mu$.

We can now formally define the notion of drift segments:
\begin{definition}
    Let $(\D_t,P_T)$ be a drift process from $\T$ to $\X$. Let $L \subset \X$ be a $\D_\T$ non-null set, then the \emph{restriction} of $\D_t$ onto $L$ is as the drift process with kernel $A \mapsto \D_t(A \mid L) = \D_t(A \cap L)/\D_t(L)$ and time distribution $W \mapsto \D(W \times L \mid \T \times L)$, with $\D$ the holistic distribution of the original drift process. We refer to $\T_L = \{t\in \T \mid \D_t(L) > 0\}$ is the \emph{active time} of $L$.

    A $\D_\T$ non-null set $L \subset \X$ is called a \emph{drift segment} if the restriction $(\D_t(\cdot \mid L), \D(\cdot \times L \mid \T \times L))$ has no drift. A drift segment is called maximal iff it is maximal with respect to set inclusion, i.e., if for every $L \subset L'$ we either have $\D_\T(L' \setminus L) = 0$ or the restriction with respect to $L'$ has drift. 

    A collection of drift segments $L_i,\; i \in \N$ that cover $\X$, i.e., $\cup_i L_i = \X$, is called a \emph{drift segmentation}. If all segments are maximal then the segmentation is \emph{optimal}.
\end{definition}

The notion of a maximal drift segment comes from the observation that if $L$ is a drift segment, then every subset $L'' \subset L$ is a drift segment, too. Thus, maximal enforces that the segments are of reasonable size. 

We will now define drift localization. As drift can be arbitrarily complicated, we describe the drifting region as the complement of the non-drifting region, i.e., which part of the distribution has to be ``removed'' in order to make the drift disappear.
As pointed out above, the term of a drift segment only encodes homogeneous drifting behaviour, i.e., there is no drift going on inside of each segment. However, there can still be drift between the segments, e.g., in \Figname~\ref{fig:segmentation-visulazation} there are three segments, but every single point in the data space if affected by drift. If we truly want no drift, then we also have to take the drift between the segments into account. However, using all the consideration done so far this is nothing more than to state that $\D_t(L)$ is constant. As a consequence we have that $\D(W \times L \mid \T \times L) = P_T(W)$. This then leads to the following definition:

\begin{definition}
\label{def:localization}
A \emph{drift locus} is a measurable set $L \subset \X$ such that $(\D_t( \cdot\mid L^C),P_T)$ has no drift and $t \mapsto \D_t(L)$ is $P_T$-a.s. A drift locus $L$ is \emph{minimal} if it is contained in every other drift locus $L'$ up to a $\D_\T$-null set, i.e. $\D_\T(L \setminus L')= 0$.
We refer to the process of finding the minimal drift locus as \emph{drift localization}.
\end{definition}

The idea of the drift locus is that it is the complement of a segment that does not show any drift, and also no shuffling. The notion of minimality is then analogous to the maximality of the non-drifting segment. Notice that since $\D_t(L)$ is constant, we do not need to consider the active time $\T_L$ as it is either $\T$ or $\emptyset$ ant the latter would imply a $\D_\T$-null set. 

As discussed in \citep{localization} the notion of a minimal drift locus has several nice properties. Among those is the fact that in all practically relevant cases, there is a unique minimal drift locus so the notion of drift localization makes sense from a theoretical point of view. Furthermore, the minimal drift locus is not empty if and only if there is drift. In the following, we will discuss how to obtain a drift localization given data. 

\subsection{A general scheme for drift localization}

\begin{wrapfigure}{r}{0.4\textwidth}
    \centering
    \resizebox{0.4\textwidth}{!}{
    \begin{tikzpicture}
    \node (x0) at (0,0) {$x_0$};
    \node (x1) [right of = x0] {$x_1$};
    \node (x2) [right of = x1] {$x_2$};
    \node (x3) [right of = x2] {$\dots$};
    \node (x4) [right of = x3] {$x_t$};
    \node (x5) [right of = x4] {$x_{t+1}$};
    \node (x6) [right of = x5] {$\dots$};
    \node (x7) [right of = x6] {$x_{t+n}$};
    \node (x8) [right of = x7] {$\dots$};
    \node (x9) [right of = x8] {$x_T$};

    \node (start) [above =3.5em of x0] {};
    \node (end) [above =3.5em of x9] {};
    \node (windowStart) [above =3em of x4] {[};
    \node (windowEnd) [above =3em of x7] {]};
    \node (t) [above of=windowEnd, yshift=-0.5cm] {$t$};
    \node (window) [left of=t, xshift=-0.5cm, yshift=-0.5em] {$W(t)$};
    \draw [-stealth] (start.west) -- (end.east);

    \node (sample) [ellipse, draw,minimum height=1cm, minimum width=4cm, right of = x4, xshift=0.6cm]  {};
    \node (sampleT) [above right=-0.2em and -0.2em of sample] {$S(t)$};
    \node (dist) [above = 1.5em of sample] {$\D_{W(t)}$};
    
    \draw [-stealth] (dist.south) -- (sample.north);

    \node (dd) [rectangle, draw, minimum width=6cm, minimum height=14cm, below =3.5em of sample.south] {};
    \draw [-stealth] (sample.south) --   (dd.north);
    \node (ddt) [below = 1em of dd.north] {\textbf{Drift Localization}};
    \node (s1) [rectangle, draw, minimum width=2cm, minimum height=2cm, below= of dd.north, xshift=-1.5cm] {\includegraphics[width=2cm]{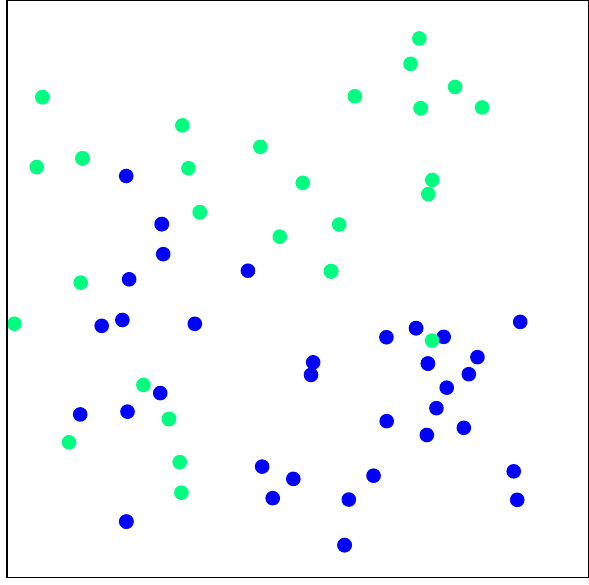}};
    \node (s2) [rectangle, draw, minimum width=2cm, minimum height=2cm, below = of s1] {\includegraphics[width=2cm]{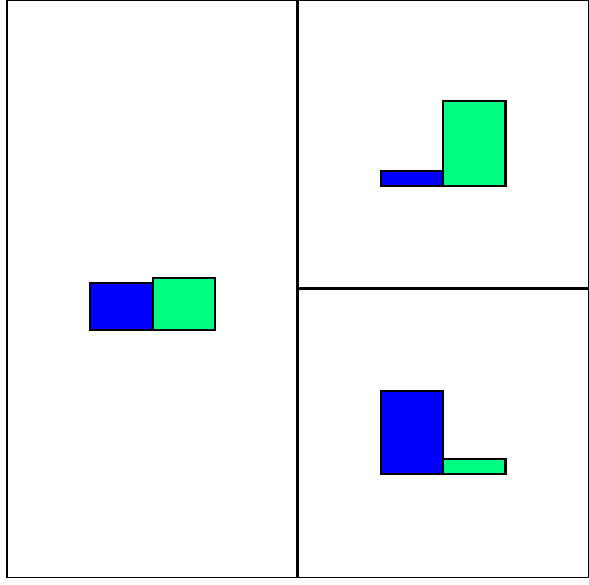}};
    \node (s3) [rectangle, draw, minimum width=2cm, minimum height=2cm, below = of s2] {\includegraphics[width=2cm]{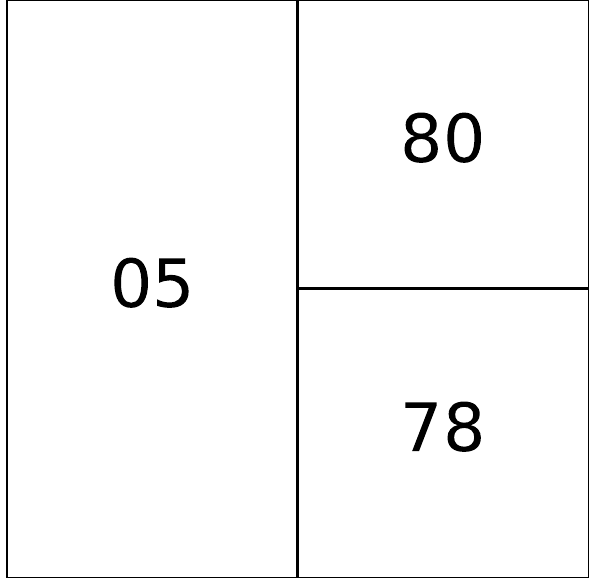}};
    \node (s4) [rectangle, draw, minimum width=2cm, minimum height=2cm, below = of s3] {\includegraphics[width=2cm]{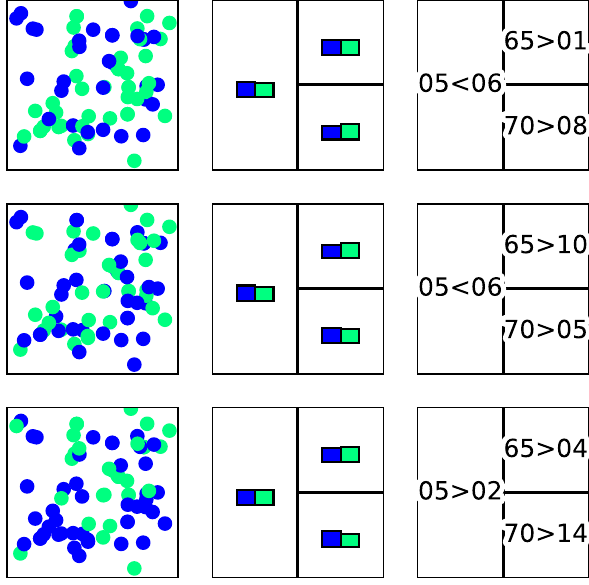}};
    \node (s1t) [right =0.2em of s1.east] {\small \begin{tabular}{c} Stage 1:\\ Acquisition\\of data   \end{tabular}};
    \node (s2t) [right =0.2em of s2.east] {\small \begin{tabular}{c} Stage 2:\\ Building\\a descriptor    \end{tabular}};
    \node (s3t) [right =0.2em of s3.east] {\small \begin{tabular}{c} Stage 3:\\Computing\\dissimilarity   \end{tabular}};
    \node (s4t) [right =0.2em of s4.east] {\small \begin{tabular}{c} Stage 4:\\Normalization   \end{tabular}};
    \node (out) [below = of dd.south] {drift segments};
    \draw [-stealth] (dd.south) -- (out.north);
    \draw [-stealth] (s1.south) -- (s2.north);
    \draw [-stealth] (s2.south) -- (s3.north);
    \draw [-stealth] (s3.south) -- (s4.north);
    
\end{tikzpicture}}
    \caption{Visualization of drift detection from a data stream. Given a data stream, for each time window $W(t)$ a distribution $\D_{W(t)}$ generates a sample $S(t)$. A drift detection algorithm estimates whether the $S(t)$ contains drift or not by performing a four-stage detection scheme. Illustrated algorithm uses: tow-windows (stage 1), tree-histograms (stage 2), leaf-wise total variation norm (stage 3), and bootstrap normalization (stage 4).}
    \label{fig:localization-stages}
\end{wrapfigure}
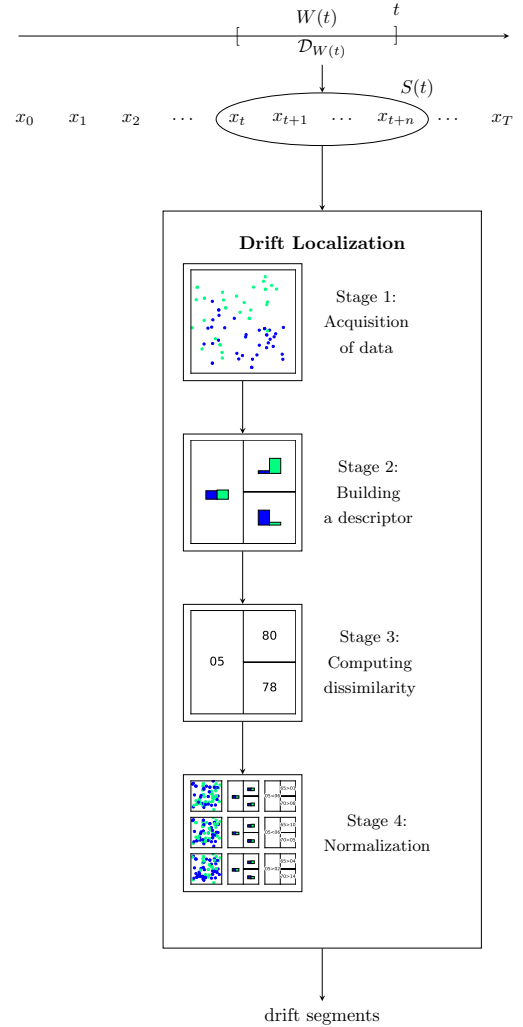

To understand the workings of drift localization algorithms, we can essentially apply the same 4-stage scheme we already used for drift detection, see \Figname~\ref{fig:localization-stages}.
\paragraph*{Stage 1: Acquisition of data} 

As a first step, we again need a strategy for selecting which data points are to be used for further analysis. Most approaches rely on some instantiation of sliding window strategies. Again we are free to consider sliding, fixed, or growing, as well as implicit reference windows. However, as we usually require a large amount of data for the localization task, to the best of our knowledge there are no methods that make use of an implicit reference window.
Similar preprocessing steps as for drift detection, such as a deep latent space embedding, are reasonable tools that have been applied successfully in the literature~\citep{neucomp}. 

\paragraph*{Stage 2: Building a descriptor}
Just as drift detection, drift localization algorithms split the data processing into two steps building a descriptor from data first and then analyzing it. In contrast to drift detection, those usually offer a quite direct connection between locations in data space and the structure of the descriptor. Commonly used are decision trees~\citep{kdqtree,localization,segmentation} or $k$-neighbour based descriptors~\citep{LDD,localization}. However, depending on the analysis algorithm nearly arbitrary machine learning models can be used as descriptors~\citep{localization}.

\paragraph*{Stage 3: Computing dissimilarity} 
Based on the descriptor a drift score is computed. In contrast to drift detection, where the score is used to describe the global amount of drift, in drift localization it computes the amount of drift in a region of the data space~\citep{kdqtree,localization,segmentation} or a single data point~\citep{LDD,localization}. In particular, for methods that make use of region-wise computations this can be considered as performing a common drift detection that only takes a small region of the data space into account~\citep{kdqtree}. Indeed, many dissimilarities commonly used in drift detection are essentially based on the idea, that there is drift if we can partition the data space in such a way that the number of samples observed before and after the drift in a single partition differ significantly. In particular, all aforementioned drift detection algorithms make use of this idea in one way or another.

\paragraph*{Stage 4: Normalization} Similar to drift detection, the obtained dissimilarities are typically very setup-specific and do not allow a direct conclusion regarding whether or not a certain sample or region is drifting. A common strategy is to make use of some kind of bootstrap statistical test in order to either find the parameters under the $H_0$-hypothesis~\citep{LDD} or directly compute a $p$-value for the point or region~\citep{kdqtree,localization}. 

\subsection{Approaches}

Although, there are several methods for drift detection and some that allow further analysis of the drift, drift localization in the sense described above is a less popular research question. Most methods that admit such an option either use it as a subroutine of drift detection or allow for it as a mere byproduct~\citep{kdqtree} rather than an explicit aim for drift analysis technology \citep{lu_learning_2018}.
The majority of algorithms that allow localization are based in one way or another on performing drift detection on a local scale~\citep{kdqtree,LDD,localization}, i.e., instead of analyzing the entire dataset at once, only local subspaces are analyzed. This has the drawback that we usually have less data to work with. On the other hand, as we are already local in data space we can make use of far simpler detection schemes. 

For example, if we make use of a grid-based binning then the total variation norm is approximated by counting the samples of each distribution per bin, taking the difference of those numbers, and summing up:
\begin{align*}
    \widehat{\Vert P - Q \Vert} = \sum_{i = 1}^k \left| \frac{1}{m}\sum_{j = 1}^m \1[X_j \in G_i] - \frac{1}{n}\sum_{j = 1}^n \1[Y_j \in G_i] \right|,
\end{align*}
where $G_1,\dots,G_k \subset \X, \; \cup_{i = 1}^k G_i = \X,\; G_i \cap G_j = \emptyset$ for $i \neq j$ are the grid-cells, $X_1,\dots,X_m \sim P$, $Y_1,\dots,Y_n \sim Q$ i.i.d. Similar estimation strategies can be applied to all sorts of distances~\citep{ida2022}.

Just as in (global) drift detection, the estimates are usually based on two-window approaches where we first select a split point and then compare the distributions before and after. As in drift detection, this split point can be chosen more or less arbitrarily. However, as the algorithms are usually less robust due to the even smaller amounts of data they are working with, it can be beneficial to first determine the actual split point using a suitable drift detector and then perform the localization. 
The common next step is to use the descriptor, which is constructed to work locally in data space to perform local drift detection. As pointed out above, such descriptors only need to count the number of points in the vicinity of the query point. As pointed out by \citep{localization} this can be considered as a probabilistic classification task, where the question of whether we make a statement about the local region or sample depends on the model used. 
This idea was further analyzed in \citep{localization} giving a theoretical justification for the most approaches available. In particular, it was shown that essentially every probabilistic classifier can be used for drift localization by analyzing its output.

In the following we will consider four exemplary approaches in more detail:

\paragraph*{$kdq$-Tree}
The algorithm is one of the oldest implementations for drift localization~\citep{kdqtree}. It is a two-window approach (stage 1), designed to work with vectorial data only. 
The main idea is to grow a $ kd$-tree-like data structure to obtain a binning (stage 2). More precisely, the trees are obtained by iterating over each dimension in every recursion step and splitting the area right in the middle of said dimension as long as enough data is available. This assures that the volume of every leaf shrinks exponentially with each recursion. It is important to note that, up to the stopping criterion, the tree does not take the data distribution into account.

Once the tree is grown, it computes the symmetrized Kullback-Leibler divergence to compare the number of samples coming from each window on every leaf which serves as the drift score (stage 3). 
This way we obtain a score for every leaf and thus region in the data space. Then a scanner statistic is used to compute a threshold (stage 4) which also depends on user-defined parameters. If the score of a leaf exceeds the threshold then the leaf area is considered to be drifting. 

\paragraph*{LDD-DIS}
The algorithm~\citep{LDD} is a two-window (stage 1), neighbor-based (stage 2) approach that computes a drift score for every data point. 
It is based on the \emph{Local Drift Degree} which a the ratio of the number of points in the $k$-neighbourhood query point. It divides the number of points in the same time window as the query point by the number of points in the other window, minus 1 (stage 3).
It is thus close to 0 if the ratio is even and deviates if there are far more samples from one window than the other. 
By an application of the central limit theorem, the authors show that for large $k$ the scores should follow a normal distribution if there is no drift. The parameters of this normal distribution are computed using a bootstrap permutation scheme. This distribution is then used for normalization (stage 4). 

Notice that the ratio that forms the heart of the LDD is closely related to the predicted probability of a $k$-neighbour classifier. 

\paragraph*{Model-Based Drift Localization}
In \citep{localization} a family of algorithms that make explicit use of machine learning models has been introduced. The algorithms can be classified as multiple-window-based approaches (stage 1), i.e., two windows or more. For simplicity and comparability, we will consider the two-window case here. 

Very similar to virtual classifiers~\cite{kifer2004detecting,hido2008unsupervised} and LDD a probabilistic classifier is trained to predict the window or time point each sample belongs to. This model serves as a descriptor (stage 2). The drift score, which the authors refer to as informativity, is then derived from the classifier prediction. In order to cope with class imbalance, i.e., different numbers of samples per window, the prediction is compared to the prediction of the constant model (stage 3), i.e., size ratio of windows, using the normalized Kullback-Leibler divergence.
As pointed out by the authors, from a Bayesian point of view, informativity can be interpreted as ``how much information on the time is lost, if we do not take the place into account''. It then is shown by the authors that informativity takes on values between 0 and 1. The mean informativity is 0 if and only if there is no drift, and the informativity at one point is larger than 0 if and only if that point belongs to the minimal drift locus. This serves as a theoretical justification for the presented method and other methods like LDD-DIS or $kdq$-trees.

Algorithmically, the authors suggest making use of a permutation bootstrap test that uses informativity as a point-wise statistic in order to cope with effects like overfitting (stage 4). However, it was also discussed that for many commonly used models like $k$-nearest neighbor, decision trees, or random forests, the corresponding $H_0$ distribution can be computed analytically. Thus, we do not have to refer to explicit computations. Furthermore, due to the supervised training scheme, we can make use of techniques like cross-validation to automatically choose model parameters.

\paragraph*{Drift Segmentation}
As stated in section~\ref{sec:drift-localization-def} drift localization can be considered as a downstream-tasks of what is referred to as drift segmentation~\citep{segmentation}, i.e., the decomposition of the data space into regions with homogeneous drifting behavior. Assuming a drift segmentation is provided to us, we can easily check for each segment whether there is drift using a simple drift detector (stages 3 \& 4). 
Drift segmentation can be approached using similar ideas as applied in \citep{localization}. 
However, instead of performing probabilistic classification, conditional density estimation is employed on a single sliding window (stages 1 \& 2), i.e., the model is trained to predict $x \mapsto \P_{T\mid X = x}$. This way, the necessity of a split point or several windows is no longer given. In this sense, drift segmentation relates to block-based drift detectors. 

Algorithmically, the different methods mainly differ in how they approach the construction of the descriptor (stage 2): The authors of \citep{segmentation} approached drift segmentation using a specially trained kind of decision tree, called the Kolmogorov tree, that uses the Kolmogorov-Smirnov test as a split criterion that reduces the statistical dependency of data and time very similar to DAWIDD. The leaves of the resulting tree then approximate the drift segments. In \citep{neucomp} it was pointed out that similar results can be obtained using any segmentation-based model. Also, using ideas from \citep{momenttrees,izbicki2017converting} instead of special training allows a large variety of models. Furthermore, even non-segmentation-based models can be used to derive an informed metric that can then be used for clustering, the resulting clusters then approximate the segments~\citep{neucomp}.

\subsection{An analysis}

So far, we analyzed the discussed drift localization methodologies on a conceptual level. In the remainder of this part, we focus on conducting a more practical analysis and providing guidelines for the practical usage of these methods.
\paragraph*{Experimental Setup}
\emph{Dataset:}
We consider a single, 2-dimensional, synthetic dataset that is sampled from a uniform distribution on a square. Drift is induced by a shift along the diagonal, i.e., in $x$- and $y$-direction, with different lengths (\emph{intensity}). Notice that this corresponds to the uniform setup in the drift detection experiment. 
Besides drift intensity, we again also consider the total number of \emph{dimensions} (as before), random \emph{rotations}, and the number of \emph{samples}. We assume that we know the time point of drift and that we are provided with an equal amount of samples before and after the drift. For the random rotation we first generate the dataset, center at 0, and then multiply with a $\lambda O + (1-\lambda) I$ for various $\lambda \in [0,1]$, where $O$ is a randomly sampled orthogonal matrix (random rotation) and $I$ is the identity matrix (data is axis aligned). We consider all combinations of parameters.

\emph{Method:}
We consider the $kdq$-Tree, LDD-DIS, model-based drift localization based on random forests~\citep{localization}. As two out of three methods are tree-based and thus align the segments with the coordinate axis, we also consider the effect of random rotation.

\emph{Evaluation:}
We perform a sample-based evaluation, i.e., we want to predict for each data point whether or not it is affected by drift. Here, we consider every sample in the overlap of the square as non-drifting and every other data point as drifting. 
To evaluate the methods we again make use of the ROC-AUC for the same reasons. In particular, in contrast to many other scores like F1 or accuracy, the ROC-AUC is not affected by the expected class imbalance.  

\begin{figure}
    \centering
    \begin{tabular}{c}
    \includegraphics[width=0.48\textwidth]{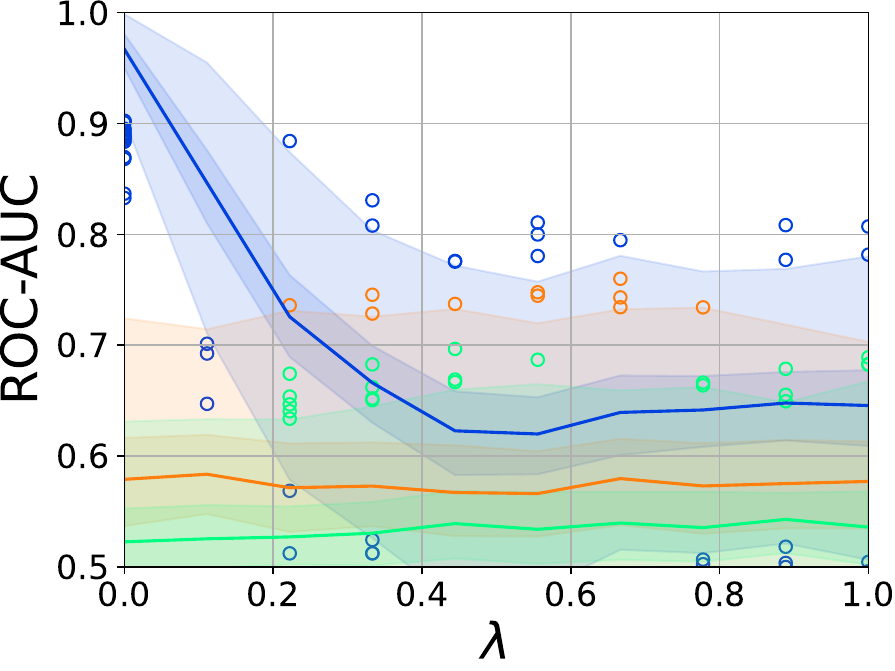}\\
    \includegraphics[height=1.75em]{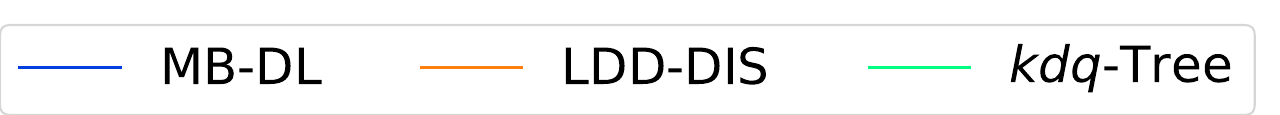}
    \end{tabular}
    \caption{Effect of axis alignment ($\lambda$) on localization performance (intensity: 0.05, samples: 750, dimensions: 5). Graphic shows median (line), $25\%-75\%$-quantile (inner area), $\min-\max$-quantile (outer area), and outliers (circles).}
    \label{fig:align-loc}
\end{figure}

\begin{figure}
    \centering
    \begin{tabular}{ccc}
        \includegraphics[width=0.32\textwidth]{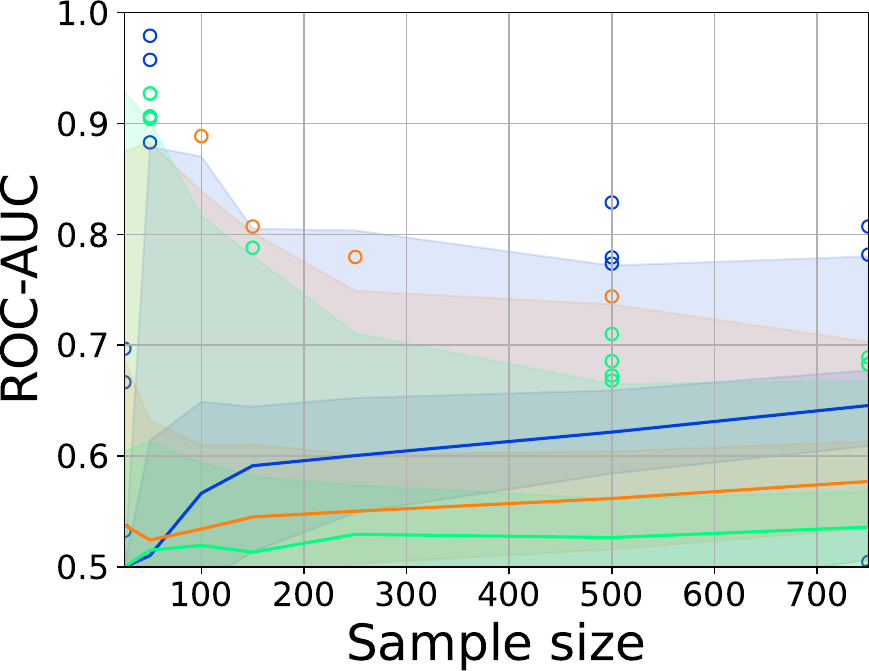} &
        \includegraphics[width=0.32\textwidth]{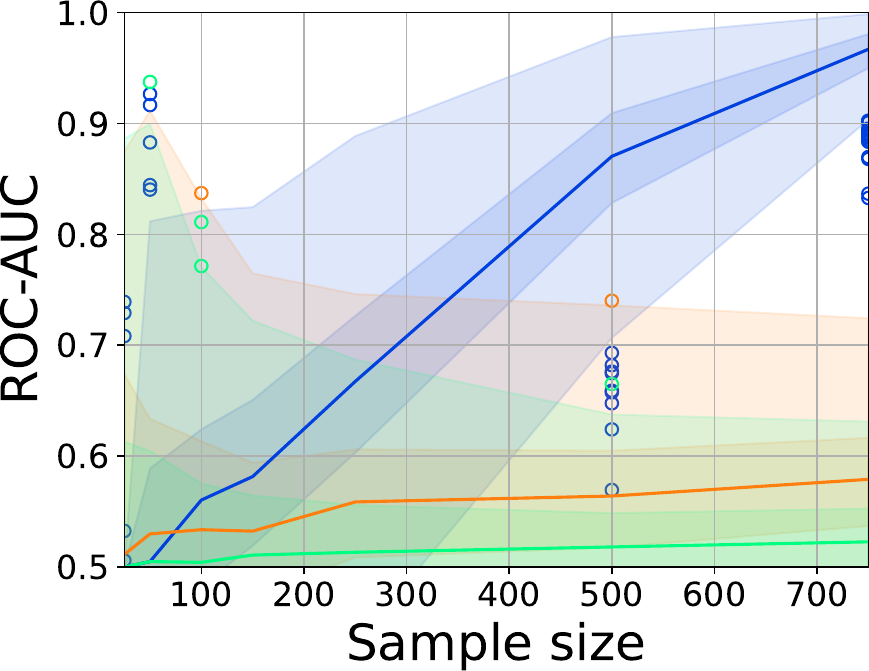} &
        \includegraphics[width=0.32\textwidth]{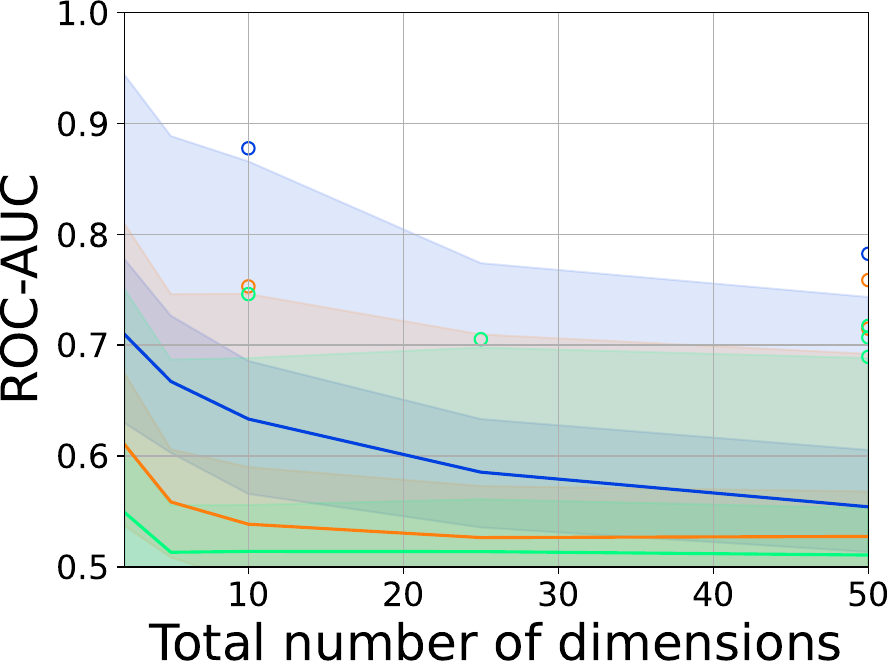} \\
        $\lambda = 1$ & $\lambda = 0$ & $\lambda = 0$ \\
        \multicolumn{3}{c}{\includegraphics[height=1.75em]{src/out/loc/legend.pdf}}
    \end{tabular}
    \caption{Effect of number of samples and dimensionality on localization performance for various choices of $\lambda$ (intensity: 0.05, samples: 250, dimensions: 5). Graphic shows median (line), $25\%-75\%$-quantile (inner area), $\min-\max$-quantile (outer area), and outliers (circles).}
    \label{fig:samp-dim-loc}
\end{figure}

\begin{figure}
    \centering
    \begin{tabular}{cc}
    \includegraphics[width=0.48\textwidth]{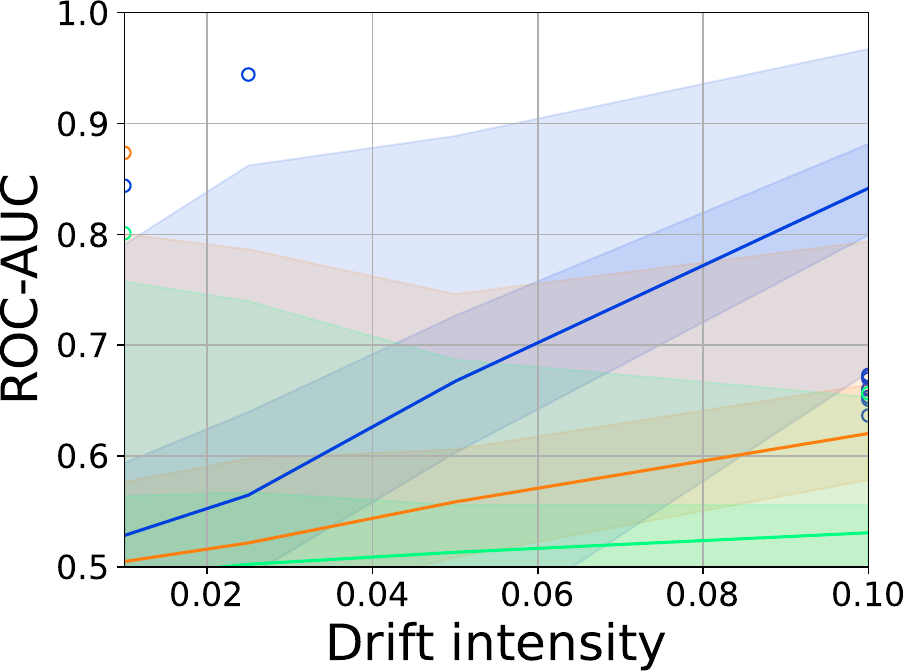}&
    \includegraphics[width=0.48\textwidth]{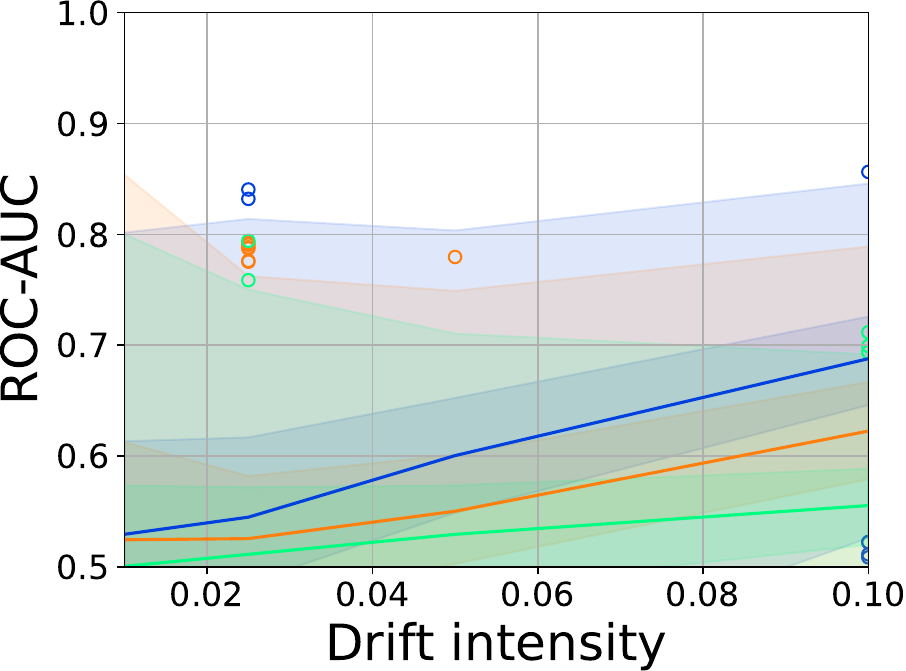}\\
    $\lambda = 0$ & $\lambda = 1$ \\
    \multicolumn{2}{c}{\includegraphics[height=1.75em]{src/out/loc/legend.pdf}}
    \end{tabular}
    \caption{Effect of drift intensity on localization performance for various choices of $\lambda$ (samples: 250, dimensions: 5). Graphic shows median (line), $25\%-75\%$-quantile (inner area), $\min-\max$-quantile (outer area), and outliers (circles).}
    \label{fig:intensity-loc}
\end{figure}

\paragraph*{Overall results}
The overall results of the experiment (see \Figname~\ref{fig:align-loc}-\ref{fig:intensity-loc}) show that the problem of drift localization is a comparably hard one and still requires additional research. The overall ranking of the methods in our simplified experiments places the MB-DL approach at the top, followed by LDD-DIS and $kdq$-Trees at the last place. This is consistent with the findings in the original paper where more complex datasets were analyzed\citep{localization}. 
In nearly all parameters $kdq$-Trees perform only slightly better than random, LDD-DIS barely ever reaches a score 0.6 or higher. This, together with the very high variance makes the analysis comparably hard.

\paragraph*{Axis-alignment}
As one can see in \Figname~\ref{fig:align-loc}, axis alignment is one of the most crucial parameters for MB-DL, for the other two approaches it is nearly irrelevant. For the MB-DL applied to a window of 150 samples or more, we observe an extreme decline in performance when we switch $\lambda = 0$ (perfectly axis-aligned) to $0.5$, after that, the performance stays at a constant, low level. This is to be expected as random forests use axis-aligned splits and thus can hardly cope with classifications that require making use of correlations as is the case if $\lambda > 0$. In the following, we will thus explicitly discuss the cases $\lambda = 0$ and $\lambda = 1$ separately. 

We thus suggest applying some kind of preprocessing or other model in case of drift that is mainly present in the correlation. 

\paragraph*{Sample size and dimensions}

As is expected all methods profit from more samples (see \Figname~\ref{fig:samp-dim-loc}). However, the increase in performance of $kdq$-Tree is not significant and might be due to random chance. In the case of MB-DL the increase for $\lambda = 1$ is only moderate and comparable to LDD-DIS, for $\lambda = 0$ the increase of performance is very strong and nearly linear until it reaches nearly perfect classification. 

Similar results can be found for the number of noise dimensions. While all methods suffer from high dimensionality, MB-DL does rather moderately in case $\lambda = 1$, this is to be expected as the detection scheme is trained in a supervised fashion and thus can perform feature selection in this case. Similar effects cannot be observed for $kdq$-Trees as they do not optimize the tree structure for the problem at hand, LDD-DIS which is not capable of feature selection in the first place, or MB-DL if $\lambda = 0$ as in this case feature selection is not possible.  

To conclude, drift localization requires a comparably large amount of data and high dimensionality poses a problem in particular if no feature selection is possible. Still, even in this case model-based approaches might still be the best choice. 

\paragraph*{Drift intensity}
Just as in drift detection, the larger the drift the easier the task becomes. This finding is systematic across consistent across all methods as summarized in \Figname~\ref{fig:intensity-loc}. Furthermore, we see a nearly linear relationship between the drift intensity and the increase of performance. This is true for both cases $\lambda = 0$ and $\lambda = 1$, but for the latter the increase is much steeper.

In particular, we suggest making use of a split point selection by using an appropriate drift detector like ShapeDD in order to increase localization performance. This seems to be the best way to increase performance in a general practical setup where other preprocessing or model choices are not available. 

\subsection{Conclusion and Guidelines}
Investigating the task of drift localization, we again provided a formal definition of the task and classified existing approaches into the presented four-staged framework. Overall, one can say that research on this task is still very limited with few approaches existing. In our experiments we only reported good results for few of them. Thus, further research is needed.

When applying drift localization methods in practical applications one should consider the following key findings of our experimental evaluation and analysis: First, of all one should use as much data as possible for the localization task. Besides, avoiding high dimensional data where feasible is desirable. To obtain good results it is crucial to select suitable drift points, e.g., by choosing reliable drift detection methods as discussed in the previous part. Finally, when reyling on tree-based methods, it is crucial to design an appropriate preprocessing if the drift inflicts itself in data correlations.

\section{Drift Explanations\label{sec:drift-explanation}}
So far, working on the \emph{whether}, \emph{when}, and \emph{where}, we focused on the problems of drift detection and localization. However, the knowledge which samples are affected by the drift at which time is still not sufficient in many real-world monitoring settings. Therefore, we will focus on a more detailed analysis of the drift in this section.

\subsection{Problem and Setup}

The question of \emph{explaining} drift, i.e., describing the potentially complex and high dimensional change of distribution in a human-understandable fashion, is a relevant problem as it enables an inspection of the most prominent characteristics of how and where drift manifests itself. Hence, it enables human understanding of the change and thus is a key ingredient for the informed decision-making of human operators. 

In contrast to the problems discussed so far, drift explanations are an inherently ill-defined problem. This is partly caused by the fact the term ``explaining'' is inherently ill-posed as can be seen by considering the wide range of different explanation schemes, methods, and frameworks present in the literature. Furthermore, the choice of a suitable explanation is highly domain and problem-specific.

Providing  a detailed, complete, and understandable description of  ongoing drift, requires a large amount of information covering all relevant aspects. This usually surpasses the level of information which is required to select change points or to estimate the rate of change: While drift can be detected based on a single drifting feature, its explanation might need to address the interplay of all drifting features.

This leads us to two general insights on the topic: First,  it is not possible to provide a formal restriction let alone a definition of what drift explanations are. Second, a large number of different explanation schemes -- one for every potential use case -- is actually a desirable state of affairs. 

\subsection{A General Scheme for Drift Explanations}
Even though we cannot provide a formal definition of drift explanations, we can still analyze approaches using a similar functional scheme as we did for detection and localization. Usually, the normalization stage is not required.

\paragraph*{Stage 1: Acquisition of data:} 
As a first step, we again need a strategy for selecting which data points are to be used for further analysis. Most approaches rely on some instantiation of sliding window strategies. Again we are free to consider sliding, fixed, or growing, as well as implicit reference windows. 
Similar preprocessing steps to drift detection and localization, such as a deep latent space embedding, are reasonable tools that have been applied successfully in the literature~\citep{neucomp}. 

\paragraph*{Stage 2: Building a descriptor:}
Just as drift detection and localization, drift explanation algorithms split the data processing into two steps building a descriptor from data first and then analyzing it. Similar to drift localization, those are usually chosen with respect to the explanation task at hand. Depending on the desired explanation, a large variety of descriptors is used, but binning approaches are very common~\citep{webb2016characterizing,webb2018analyzing,pratt2003visualizing}. However, as pointed out by \cite{neucomp} nearly every machine learning model can be used as descriptors.

\paragraph*{Stage 3: Computing explanations:} 
In the last of the explanation scheme, the descriptor is analyzed. This is indeed comparable to the computation of dissimilarity as a simple quantity derived from the descriptor. Indeed, many methods simply derive such numbers such as feature-wise change intensity or change in correlation~\citep{webb2016characterizing,webb2018analyzing,wang2020conceptexplorer,pratt2003visualizing}. Here, a further analysis by means of normalization is not necessary as the data is usually directly presented to and judged by a human operator. However, also some more advanced explanation methods are available \citep{neucomp}.

\subsection{Exemplary Cases}

While explainability has been a major research interest in recent years \citep{molnar2019interpretable,rohlfing2021explanation}, more complex explanation methods for drift are still limited.
Quite a number of approaches aim for the detection and quantification of drift  \citep{lu_learning_2018,webb2018analyzing}, or its visualization \citep{wang2020conceptexplorer,webb2017understanding,pratt2003visualizing}.
Furthermore, several methods focus on feature-wise representations of drift \citep{wang2020conceptexplorer,webb2018analyzing,webb2017understanding,pratt2003visualizing}.
However, these methods face  challenges  if high-dimensional data or non-semantic features are dealt with. 
To our knowledge, there is only one approach that directly targets concept drift using more complex XAI methods for explaining drift~\citep{neucomp}.

In the following, we will group the methods based on the question of whether they focus on feature-wise analysis only, or allow for the application of more complex XAI technologies.  

\paragraph*{Feature-Based Drift Explanations}

In \citep{webb2018analyzing,webb2017understanding} the authors make use of the (conditional) \emph{drift magnitude} to visualize the intensity and change of correlation of certain features. For sets of features $F,F'$ the drift magnitude is defined as
\begin{align*}
    \sigma_{\D_\cdot, l}^{F}(s,t) &= \Vert \D_{W_l(s)}(X_F) - \D_{W_l(t)}(X_F) \Vert_\text{TV} \\
    \sigma_{\D_\cdot, l}^{F \mid F'}(s,t) &= \int \Vert \D_{W_l(s)}(X_F \mid X_{F'}) - \D_{W_l(t)}(X_F \mid X_{F'}) \Vert_\text{TV} \d \D_{W_l(s) \cup W_l(t)}(X_{F'})
\end{align*}
where $W_l(t) = (t-l/2,t+l/2)$ is the window around $t$ with length $l$, $\D_W(X_F)$ is the projection of the drift process onto the features $F$, and $\D_W(X_F \mid X_{F'})$ is the conditioning of the conditioning of $\D_W(X_F,X_{F'})$ on $\D_W(X_{F'})$. The theoretical properties of the drift magnitude was analyzed in \citep{shape}. Notice that the drift magnitude on consecutive windows also forms the basis for the Shape Drift Detector.

To estimate the drift magnitude, \citep{webb2018analyzing,webb2017understanding} use sliding windows (stage 1), and bin histograms (stage 2) which are used to compute the total variation norm for different time-points (stage 3). The results of this computation are directly presented to the user. 

\emph{ConceptExplorer} is a tool presented in \citep{wang2020conceptexplorer}. It is designed for visual inspection of drift in particular in time-series data. The tool contains several analysis and visualization tools: drift detection algorithm and event-log-plot, automatic extraction of concepts, visualization, and interaction, feature selection and relevance tools, and cross-data source analysis. 
For drift detection and feature analysis, standard tools are used. The concept analysis is mainly performed by making use of a time-binned correlation matrix.

In \citep{pratt2003visualizing} the authors suggest using \emph{brushed, parallel histograms} in order to visualize concept drift. The data distribution for each dimension is displayed using a histogram, correlations are marked by lines connecting the dimension-wise projections. The implementation presented by the authors enables user interaction by allowing the user to select subsets of points, e.g., parts of the histograms, for which more information is desired.

To visualize drift, the authors use sliding windows (stage 1) for which the representation is computed and presented side-by-side (stages 2 \& 3). 

\paragraph*{Model-Based Drift Explanations}
\begin{figure}
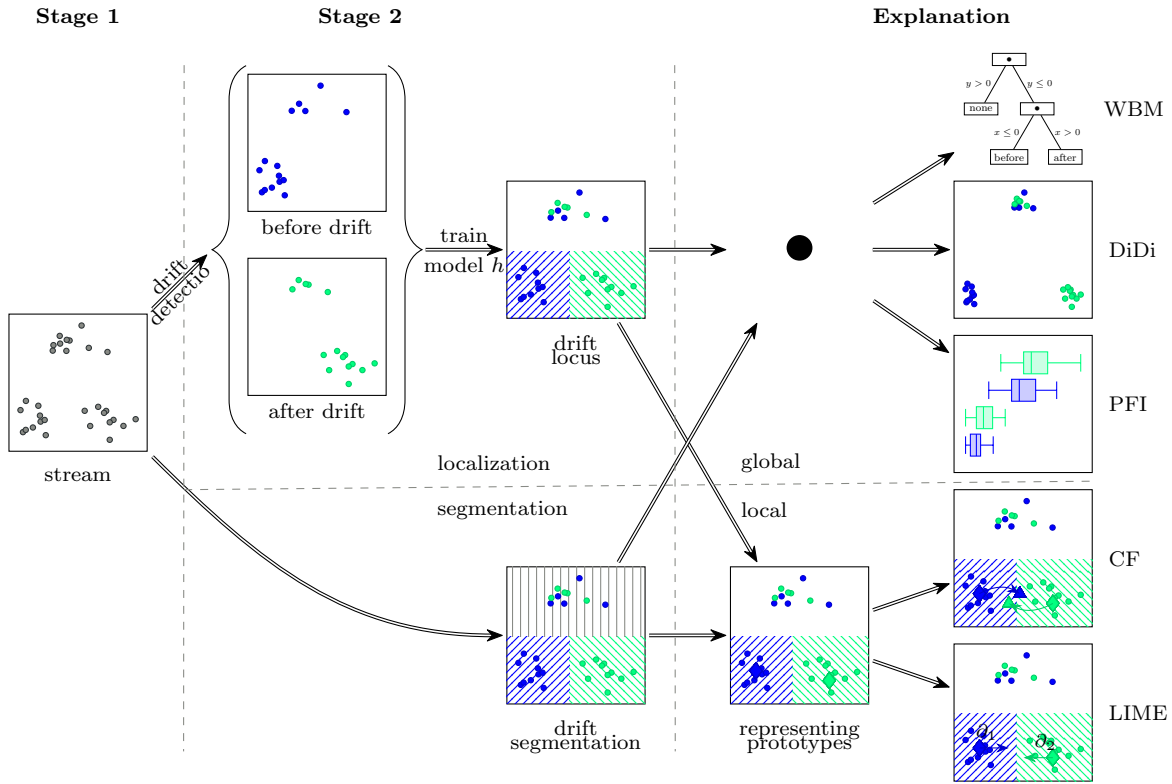

    \footnotesize
    \centering
    \tikzvoodoo
    \caption{Model-based drift explanation strategy: In the first step a model for drift localization or drift segmentation. Afterward, the obtained model is explained by means of an explanation method.}
    \label{fig:explanations}
\end{figure}
%
The notion of model-based drift explanations was coined in \citep{esann2022,neucomp}. Simply put, the fundamental idea is that drift explanations are supposed to tell us why a drift detector raised an alarm. As stated in section~\ref{sec:drift-detection} there are several approaches that explicitly make use of machine learning models as a descriptor to detect the drift. In these cases, explaining why the model used by the drift detector obtained its results also provides us with information on why the drift detector raised an alarm or not.

In order to obtain sufficiently informative explanations we however require certain kinds of models or more precisely training schemes. As discussed before, for drift detection frequently detecting a single drifting feature is sufficient. Since this might be not sufficient to provide a valuable explanation, we instead rely on drift localization or segmentation. \cite{neucomp} pointed out that these training schemes are suitable for explaining the change. Besides, they are applicable to a wide range of models.

Once a model is trained in an appropriate way we can obtain knowledge on the drift by analyzing it. This can be understood as follows: if the model contains all the information regarding the drift, we can analyze it as a proxy for the drift. This again fits into our staged scheme where the model serves as the descriptor: As visualized in \Figname~\ref{fig:explanations}, first, data is acquired and a model describing the data is trained (stage 1\&2). In the second step, the obtained model is explained by a suitable explanation approach.

There are several ways the explanation can be done. This is another strength of this approach: not only can we choose the model that fits our data best, but also the explanation that fits our problem best. In the original paper the authors considered the non-exhaustive list of the following explanation methods~\citep{neucomp}: 
\begin{itemize}
    \item Linear models or decision trees belong to the class of \emph{interpretable models} which have the nice properties of being naturally understandable by humans~\citep{molnar2019interpretable,du2019techniques}. Yet, they usually suffer problems due to low complexity.
    \item More complex explanations are provided by \emph{discriminative dimensionality reduction} which provides a global overview of the model behavior using model-enriched dimensionality reduction techniques~\citep{venna2010information,schulz2019deepview,yang2020diagnosing}. 
    \item \emph{Global feature importance and relevance} techniques like permutation feature importance, feature importance, feature relevance, and Shapley-values offer feature-wise explanations based on the model scheme~\citep{breiman2001random,nilsson2007consistent,shapley1951notes}. In contrast to other methods, those usually come with formal descriptions and guarantees on what they can and cannot do. Such have been shown to be useful for various setups with semantic features, in particular sensor networks~\citep{neucomp}.
    \item \emph{Local feature importance} techniques like Saliency Maps~\citep{simonyan2013deep} or Local Interpretable Model-agnostic Explanations (LIME)~\citep{Ribeiro2016WhySI} offer feature-wise analysis on a single instance basis. This can provide more information on the single instance and offer insight into the change of correlations, however, it also requires finding samples that are relevant enough to provide additional insight if analyzing. In \citep{neucomp} a technique for finding such samples was provided.
    \item \emph{Contrasting explanations and counterfactuals} offer explanations in terms of contrasting sample pairs~\citep{molnar2019interpretable,looveren2019interpretable,yang2021cade}. In contrast to local feature-wise explanations, those do not only show which features are affected but also how they are affected. Thus, the user is directly confronted with the effect of the drift in exemplary cases. Drawbacks of this approach are that it only works well with about drift, is computationally expensive, and there are usually no guarantees that valid explanations are found in practice. 
\end{itemize}

A further advantage of model-based explanations is that the connection of drift-related problems like drift detection and localization to explanation and analysis techniques can also be used to increase the performance of the other tasks. For example, in \citep{esann2023} this connection can be used not only to transfer ideas of feature relevance theory in order to obtain drift explanations but also to perform feature selection for drift detection which resulted in significant increases in accuracy. 

There did exist works prior to \cite{neucomp} that made use of a similar scheme. However, those approaches are hand-tailored for a specific setup rather than a general framework. In \cite{yang2021cade} a combination of an autoencoder and a distance-based outlier detection in the latent space is used for drift detection. Drift explanations are provided by counterfactuals of the outlier detector. In \cite{yang2020diagnosing} drift is detected using a Gaussian mixture model in a loss-based fashion. Then, the authors use a discriminative version of t-SNE to create an embedding. 

\subsection{Conclusion and Guidelines}
Focusing on drift explanations, we identified another research gap, as much of the work in this area is still very basic. Much more work is needed to provide user-friendly explanations across different domains and settings. Additionally, evaluations in the form of user studies will be required to evaluate future approaches. Regarding the discussed methodologies, model-based explanations seem the most promising as the framework is very flexible combining model-based localization and segmentation methods with a range of established explanation schemes. The latter can be chosen to fit the real-world scenario that needs to be targeted.

\section{Conclusion\label{sec:conclusion}}
In this work, we provided definitions and categorizations of drift detection and drift localization in an unsupervised setting. Furthermore, we categorized state-of-the-art approaches and analyzed them based on a four-staged general scheme we proposed. In addition, we briefly considered drift explanations and showcased some works targeting this task.

Next to providing an overview of existing work, we analyzed the different underlying strategies to contribute guidelines on how to choose methodologies based on the attributes of the setup and the expected drift mechanism. Finally, we found that more research is required, in particular focusing on the localization and explanation tasks.

\printbibliography
\end{document}

%% file: tikzvodoo.tex
\usepackage{tikz}
\usepackage{pgfplots}
\pgfplotsset{compat=1.8}
\usetikzlibrary{arrows.meta,patterns,decorations.text,fit,calc}
\usepgfplotslibrary{statistics}

\definecolor{tangored}{HTML}{CC0000}
\definecolor{tangoplum}{HTML}{75507B}
\definecolor{tangoblue}{HTML}{3465A4}
\definecolor{tangoorange}{HTML}{F57900}
\definecolor{tangogreen}{HTML}{73D216}
\definecolor{tangobutter}{HTML}{EDD400}

\definecolor{aluminium1}{HTML}{EEEEEC}
\definecolor{aluminium2}{HTML}{D3D7CF}
\definecolor{aluminium3}{HTML}{BABDB6}
\definecolor{aluminium4}{HTML}{888A85}
\definecolor{aluminium5}{HTML}{555753}
\definecolor{aluminium6}{HTML}{2E3436}

\definecolor{winter1}{HTML}{0000ff} 
\definecolor{winter2}{HTML}{00ff80} 

\colorlet{class0c}{aluminium4}
\colorlet{class1c}{winter1}
\colorlet{class2c}{winter2}
\colorlet{class3c}{tangoorange}
\colorlet{class4c}{tangoblue}

\tikzstyle{classinv}=[draw=none, mark size=0pt]
\tikzstyle{class0}=[fill=class0c, draw=aluminium6,mark size=1pt]
\tikzstyle{class1}=[fill=class1c, draw=class1c!80!black,mark size=1pt]
\tikzstyle{class2}=[fill=class2c, draw=class2c!80!black,mark size=1pt]

\pgfmathdeclarefunction{invgauss}{2}{%
	\pgfmathparse{sqrt(-2*ln(#1))*cos(deg(2*pi*#2))}%
}

\pgfdeclarelayer{top}
\pgfsetlayers{main,top}

\newcommand{\labelsize}{\scriptsize}

\pgfplotsset{ticks=none,
	yticklabels=\empty,
	xticklabels=\empty,
	width=0.2\textwidth,
	height=0.2\textwidth,
	label style={font=\labelsize},
}

\newcommand{\plotdataset}[4][1]{
	\pgfmathsetseed{42}
	\addplot [#2,only marks, samples=10] ({invgauss(rnd,rnd)},{invgauss(rnd,rnd)});
	\addplot [#2,only marks, samples=5] ({1.5*invgauss(rnd,rnd)+#4},{invgauss(rnd,rnd)+#1*\yshift});
	\addplot [#3, only marks, samples=10] ({invgauss(rnd,rnd)+2*#4},{invgauss(rnd,rnd)});
	\addplot [#3, only marks, samples=5] ({1.5*invgauss(rnd,rnd)+#4},{invgauss(rnd,rnd)+#1*\yshift});
}

\newcommand{\plotOnlyProtos}{
	\addplot[only marks,class1,mark=diamond*,mark size=3pt](0,\protoshift);			
	\addplot[only marks,class2,mark=diamond*,mark size=3pt](2*\xshift,-\protoshift);
}

\newcommand{\plotSegment}[1][]{
	\pattern[pattern=north east lines, pattern color=class1c] (axis cs:-5,-5)--(axis cs:-5,0.5*\yshift)--(axis cs:\xshift,0.5*\yshift)--(axis cs:\xshift,-5)--cycle;
	\pattern[pattern=north west lines, pattern color=class2c] (axis cs:\xshift,-5)--(axis cs:\xshift,0.5*\yshift)--(axis cs:4*\xshift,0.5*\yshift)--(axis cs:4*\xshift,-5)--cycle;
	\ifthenelse{\equal{#1}{}}{}{
		\pattern[pattern=vertical lines, pattern color=class0c] (axis cs:-5,0.5*\yshift)--(axis cs:-5,2*\yshift)--(axis cs:4*\xshift,2*\yshift)--(axis cs:4*\xshift,0.5*\yshift)--cycle;
	}
}

\def\xshift{3.5}
\def\yshift{8}
\def\protoshift{0.5}

\newcommand{\tikzvoodoo}{
	\begin{tikzpicture}[
		node distance=3em,
		from/.style={{Stealth[length=0.8em, round]}-,double,shorten >=-0.2em,shorten <=-0.2em},
		towards/.style={-{Stealth[length=0.8em, round]},double,shorten >=-0.2em,shorten <=-0.2em},
		]
			\node(beforedrift) {%
				\begin{tikzpicture}
					\begin{axis}[xlabel={\footnotesize before drift}]
						\plotdataset{class1}{classinv}{\xshift}
					\end{axis}
				\end{tikzpicture}
			};
				\node(afterdrift) [node distance=0em, below=of beforedrift] {%
					\begin{tikzpicture}
						\begin{axis}[xlabel={\footnotesize after drift}]
							\plotdataset{classinv}{class2}{\xshift}
						\end{axis}
					\end{tikzpicture}
				};
				\draw[decorate,decoration={brace,amplitude=1.5em,mirror,raise=-0.3em}] (beforedrift.north west) -- node[xshift=-1.2em](leftdrift){} (afterdrift.south west);
				\draw[decorate,decoration={brace,amplitude=1.5em,raise=-0.3em}] (beforedrift.north east) -- node[xshift=1.2em](rightdrift){} (afterdrift.south east);	
					\node(stream) [below left=of leftdrift] {%
						\begin{tikzpicture}
							\begin{axis}[xlabel={}]
								\plotdataset{class0}{class0}{\xshift}
							\end{axis}
						\end{tikzpicture}
					} edge[towards]
					[postaction={decoration={text along path, text align=center,text={|\footnotesize |drift ||},
							raise=0.2em,},decorate}]
					[postaction={decoration={text along path, text align=center,text={|\footnotesize |detection ||},
							raise=-0.7em,}, decorate}]
					(leftdrift);

					\node[node distance=0em,below=of stream] {stream};
					\node(driftingarea) [right=of rightdrift] {
						\begin{tikzpicture}
							\begin{axis}
								\plotdataset{class1}{class2}{\xshift}
								\plotSegment
							\end{axis}
						\end{tikzpicture}
					} edge [from] node[above]{\footnotesize train} 
					node[below]{\footnotesize model $h$} (rightdrift);
					\node[node distance=0em,below=of driftingarea,font=\footnotesize,align=center] {drift \\[-0.5em] locus};
					\node (nopreproc) [right=of driftingarea] {
						\begin{tikzpicture}
							\begin{axis}[axis line style={white,}]
								\plotdataset{classinv}{classinv}{\xshift}
							\end{axis}
						\end{tikzpicture}
					} edge [from] (driftingarea);
				    \node at (nopreproc) {\Huge $\bullet$};
					\node[label=east:DiDi] (didi) [right=of nopreproc] {
						\begin{tikzpicture}
							\begin{axis}
								\plotdataset[2]{class1}{class2}{4*\xshift}
							\end{axis}
						\end{tikzpicture}
					} edge [from] (nopreproc);
					\node[label=east:WBM] (wbm) [node distance=0em, above=of didi] {
						\resizebox{.1\textwidth}{!}{
							\begin{tikzpicture}
								\begin{scope}[X/.style={draw,minimum width=3em}]
									\node[X] {$\bullet$}
									child {node[X] {none} edge from parent node[left] {\labelsize $y>0$}}
									child {node[X] {$\bullet$}
										child {node[X] {before} edge from parent node[left] {\labelsize $x\leq0$}}
										child {node[X] {after}edge from parent node[right] {\labelsize $x>0$}}
										edge from parent node[right] {\labelsize $y\leq0$}};
								\end{scope}
							\end{tikzpicture}
						}
					} edge [from] (nopreproc);
					\node[label=east:PFI] (pfi) [node distance=0em,below=of didi] {
							\begin{tikzpicture}
								\begin{axis}[
									boxplot/draw direction=x,
									ymin=0,ymax=5,
									cycle list={{class1c},{class2c}},
									tick align=inside,
									ytick={1,2,3,4},
									yticklabels={$z_2$,$z_1$,$y$,$x$},
									]
									\addplot+[fill,fill opacity=0.2,
									boxplot prepared={
										median=2,
										upper quartile=2.35,
										lower quartile=1.5,
										upper whisker=3.4,
										lower whisker=1.1
									},
									] coordinates {};
									\addplot+[fill,fill opacity=0.2,
									boxplot prepared={
										median=2.59,
										upper quartile=3.35,
										lower quartile=2,
										upper whisker=4.4,
										lower whisker=1.1
									},
									] coordinates {};
									\addplot+[fill,fill opacity=0.2,
									boxplot prepared={
										median=5.57,
										upper quartile=6.93,
										lower quartile=4.93,
										upper whisker=8.68,
										lower whisker=3.03
									},
									] coordinates {};
									\addplot+[fill,fill opacity=0.2,
									boxplot prepared={
										median=6.57,
										upper quartile=7.93,
										lower quartile=5.93,
										upper whisker=10.68,
										lower whisker=4.03
									},
									] coordinates {};
								\end{axis}
							\end{tikzpicture}
					} edge [from] (nopreproc);
					\node[label=east:CF] (cf) [node distance=0em,below=of pfi] {
						\begin{tikzpicture}
							\begin{axis}
								\plotdataset{class1}{class2}{\xshift}
								\plotSegment
								\plotOnlyProtos
								\begin{pgfonlayer}{top}
									\draw[class1c!80!black,-{Stealth}] (axis cs:0,\protoshift) to [bend left] (axis cs:1.1*\xshift,\protoshift) node[above,midway,black] {};
									\draw[class2c!80!black,-{Stealth}] (axis cs:2*\xshift,-\protoshift) to [bend left] (axis cs:0.8*\xshift,-\protoshift) node[above,midway,black] {};
									\addplot[only marks,class1,mark=triangle*,mark size=2.5pt](1.1*\xshift,\protoshift);			
									\addplot[only marks,class2,mark=triangle*,mark size=2.5pt](0.8*\xshift,-\protoshift);
								\end{pgfonlayer}
							\end{axis}
						\end{tikzpicture}
					};
					\node[label=east:LIME] (lime) [node distance=0em, below=of cf] {
						\begin{tikzpicture}
							\begin{axis}
								\plotdataset{class1}{class2}{\xshift}
								\plotSegment
								\plotOnlyProtos
								\begin{pgfonlayer}{top}
									\draw[class1c!80!black,-{Stealth}] (axis cs:0,\protoshift) -- (axis cs:0.8*\xshift,\protoshift) node[above,near start,black] {\scriptsize $\partial_1$}; 
									\draw[class2c!80!black,-{Stealth}] (axis cs:2*\xshift,-\protoshift) -- (axis cs:1.2*\xshift,-\protoshift) node[above,near start,black] {\scriptsize $\partial_2$}; 
								\end{pgfonlayer}
							\end{axis}
						\end{tikzpicture}
					};
					\node[node distance=0em] (mid) at ($(cf.west)!0.5!(lime.west)$) {};
						\node (repproto) at (nopreproc |- mid) {
							\begin{tikzpicture}
								\begin{axis}
									\plotdataset{class1}{class2}{\xshift}
									\plotSegment
									\plotOnlyProtos
								\end{axis}
							\end{tikzpicture}
						} edge [from] (driftingarea);
						\node[node distance=0em,below=of repproto,font=\footnotesize,align=center] {representing \\[-0.5em] prototypes};
						\draw (repproto) edge[towards] (cf);
						\draw (repproto) edge[towards] (lime);
						\node (segmentation) [left =of repproto] {
							\begin{tikzpicture}
								\begin{axis}
									\plotdataset{class1}{class2}{\xshift}
									\plotSegment[1]
								\end{axis}
							\end{tikzpicture}
						} edge [towards] (nopreproc);
						\node[node distance=0em,below=of segmentation,font=\footnotesize,align=center] {drift \\[-0.5em] segmentation};
						\draw (segmentation) edge [towards] (repproto);
						\node[below right=of stream] {};
						\node (betweenPfiCf) at ($(pfi.east) !0.5! (cf.east)$) {};
						\node (betweenPfiCfShifted) at ($(betweenPfiCf) + (-20em,0)$) {};
						\draw(stream) edge [towards,out=315,in=180] (segmentation);
						\node[fit=(leftdrift) (beforedrift) (betweenPfiCfShifted) (driftingarea),inner ysep=0em,inner xsep=0.85em] (loc) {}; %
						\node[fit=(segmentation) (loc.south west) (loc.south east),inner ysep=1.7em,inner xsep=0em] (seg) {};
						\draw[dashed,class0c] (loc.north west) -- (seg.south west);
						\draw[dashed,class0c] (loc.south west) -- (betweenPfiCf);
						\draw[dashed,class0c] (loc.north east) -- (seg.south east);
						\node[node distance=-0.5em,below=of loc,anchor=south west] (textLoc) {localization};
						\node[node distance=0.125em,below=of loc,anchor=north west] (textSeg) {segmentation};
						\node[anchor=west] at ($(nopreproc.west |- textLoc) + (0.5em,0)$) {global}; 
						\node[anchor=west] at ($(nopreproc.west |- textSeg) + (0.5em,0)$) {local};
						
    \node (stage1) [above=12em of stream] {\textbf{Stage 1}};
    \node (stage2) [right=8em of stage1] {\textbf{Stage 2}};
    \node (stage3) [right=20em of stage2] {\textbf{Explanation}};
					\end{tikzpicture}
				}